\documentclass[10pt,twocolumn,twoside]{article}

\usepackage{amssymb,amsmath,amsthm,mathtools,graphicx} 
\usepackage{bm} 
\usepackage{diffcoeff} 
\usepackage{booktabs} 
\usepackage{array} 
\usepackage{tabularx} 
\usepackage{makecell} 
\usepackage{collcell} 
\usepackage[table]{xcolor} 
\usepackage{etoolbox,siunitx} 
\robustify\bfseries
\usepackage{float} 
\usepackage{color} 
\usepackage{marvosym, utfsym} 
\robustify\usym
\usepackage{adjustbox} 
\usepackage{comment} 
\usepackage{censor} 
\usepackage{import} 
\usepackage{refcount} 
\usepackage{algorithm}
\usepackage{algpseudocode}
\usepackage{xifthen} 
\usepackage{xparse} 
\usepackage{xstring} 
\usepackage[numbers]{natbib} 
\usepackage[colorlinks,urlcolor=blue,linkcolor=blue,citecolor=blue]{hyperref}
\usepackage{doi} 

\NewDocumentCommand\ProcStar{m}{
    \IfBooleanTF{#1}{%
        \def\ProcessedArgument{*}%
    }{%
        \def\ProcessedArgument{}%
    }%
}
\makeatletter
\NewDocumentCommand\subtoarg{m m}{
    \expandafter\def\csname #1@sub\endcsname_##1{\csname #2\endcsname[##1]}
    \expandafter\def\csname #1\endcsname{
        \@ifnextchar_{\csname #1@sub\endcsname}{\csname #2\endcsname[]}
    }%
}
\newlength{\boxwidth}
\newlength{\boxheight}
\newsavebox{\boxdim@box}
\NewDocumentCommand\boxdim{m}{
    \sbox{\boxdim@box}{#1}%
    \setlength{\boxwidth}{\wd\boxdim@box}%
    \setlength{\boxheight}{\dimexpr\ht\boxdim@box + \dp\boxdim@box\relax}%
}
\newsavebox{\autoparbox@box}
\NewDocumentCommand\autoparbox{O{} O{} O{} m}{%
    \sbox{\autoparbox@box}{#4}%
    \parbox[#1][#2][#3]{\wd\autoparbox@box}{\usebox{\autoparbox@box}}%
}
\makeatother
\providecommand\given{} 
\newcommand\GivenResize[1][]{
\renewcommand\given{\nonscript\:#1\vert\allowbreak\nonscript\:\mathopen{}}}
\DeclarePairedDelimiter\abs{\lvert}{\rvert} 
\DeclarePairedDelimiter\parens{(}{)} 
\DeclarePairedDelimiter\bracks{[}{]} 
\DeclarePairedDelimiter\tuple{(}{)} 
\DeclarePairedDelimiterX\set[1]{\lbrace}{\rbrace}{\GivenResize[\delimsize]#1} 
\DeclarePairedDelimiterX\group[2]{\langle}{\rangle}{\GivenResize[\delimsize]#1} 
\DeclarePairedDelimiterX\civ[2]{[}{]}{#1\,,\allowbreak#2} 
\DeclarePairedDelimiterX\oiv[2]{(}{)}{#1\,,\allowbreak#2} 
\DeclarePairedDelimiterX\coiv[2]{[}{)}{#1\,,\allowbreak#2} 
\DeclarePairedDelimiterX\ociv[2]{(}{]}{#1\,,\allowbreak#2} 
\DeclarePairedDelimiterXPP\prob[1]{P}{[}{]}{}{\GivenResize[\delimsize]#1} 
\makeatletter
\NewDocumentCommand\smashop{O{btlr} m}{
    \def\smashop@vargs{}
    \def\smashop@hargs{}
    \IfSubStr{#1}{t}{\g@addto@macro\smashop@vargs{t}}{}
    \IfSubStr{#1}{b}{\g@addto@macro\smashop@vargs{b}}{}%
    \IfSubStr{#1}{l}{\g@addto@macro\smashop@hargs{l}}{}%
    \IfSubStr{#1}{r}{\g@addto@macro\smashop@hargs{r}}{}%
    \smash[\smashop@vargs]{%
        \ifthenelse{\isempty{\smashop@hargs}}{%
            #2%
        }{%
            \smashoperator[\smashop@hargs]{#2}%
        }%
    }%
}
\newcommand\ExpectSymbol{\mathbb{E}}
\newcommand\ExpectMargin{1pt 0}
\DeclareMathOperator*{\ExpectOp}{\mathchoice
    {\vcenter{\hbox{\marginbox{\ExpectMargin}{\resizebox{10pt}{!}{$\ExpectSymbol$}}}}}
    {\ExpectSymbol}
    {\ExpectSymbol}
    {\ExpectSymbol}}
\DeclarePairedDelimiterX\ExpectDelim[1]{[}{]}{\GivenResize[\delimsize]#1}
\NewDocumentCommand\expect{o e{_} s o g}{
    \IfValueTF{#2}{
        \def\expect@size{#4}%
    }{
        \def\expect@size{#1}%
    }%
    \ExpandArgs{o}\IfValueTF{\expect@size}{
        \edef\expect@size{[\expandafter\noexpand\expect@size]}
    }{
        \def\expect@size{}%
    }%
    \IfBooleanT{#3}{
        \def\expect@size{*}%
    }%
    \IfValueTF{#2}{
        \IfValueTF{#1}{
            \smashop[#1]{\ExpectOp_{#2}}%
        }{
            \ExpectOp_{#2}%
        }%
    }{
        \ExpectOp%
    }%
    \IfValueT{#5}{
        \expandafter\ExpectDelim\expect@size{#5}
    }%
}
\DeclareMathOperator*{\entropyOp}{\mathrm{H}}
\DeclarePairedDelimiterXPP\entropyDelim[1]{\entropyOp}{(}{)}{}{#1}
\NewDocumentCommand\entropy{>{\ProcStar}s O{} m g}{\entropyDelim#1[#2]{#3\IfValueT{#4}{,#4}}}
\DeclareMathOperator*{\KLOp}{\mathrm{KL}}
\DeclarePairedDelimiterXPP\KL[2]{\KLOp}{[}{]}{}{\GivenResize[\delimsize\vert\delimsize]{#1 \given #2}}
\newcommand\union{
    \@ifnextchar_{%
        \bigcup%
    }{%
        \@ifnextchar^{%
            \bigcup%
        }{%
            \cup%
        }%
    }%
}
\robustify\union
\newcommand\intersec{
    \@ifnextchar_{%
        \bigcap%
    }{%
        \@ifnextchar^{%
            \bigcap%
        }{%
            \cap%
        }%
    }%
}
\robustify\intersec
\makeatother
\newcommand\mat[1]{\begin{bmatrix}#1\end{bmatrix}} 
\DeclareMathOperator*{\argmax}{arg\,max}

\def\mathscale{1}
\NewDocumentCommand\scalemath{O{\mathscale} m}{\scalebox{#1}{\mbox{\ensuremath{\displaystyle #2}}}}
\NewDocumentCommand\subalign{m}{{%
    \renewcommand\arraystretch{0.5}%
    \begin{array}{@{}>{\scriptstyle}r@{}>{\scriptstyle}l@{}}#1\end{array}%
}}
\makeatletter
\newcommand\msub[1]{
    \def\ms@sub{#1}
    \@ifnextchar_{\ms@merge}{_{\ms@sub}}
}
\def\ms@merge_#1{_{\ms@sub,#1}}
\robustify\msub 
\makeatother
\newcommand\vs[1]{\rule{0pt}{#1}}

\newcommand\lb[1]{#1\msub{\textsc{l}}} 
\newcommand\ub[1]{#1\msub{\textsc{u}}} 
\newcommand\ind{I} 

\makeatletter
\newcommand{\newreptheorem}[2]{%
    \newtheorem{rep@#1}{#2}%
    \NewDocumentEnvironment{rep#1}{m o}{%
        \lsuffix@set{##1}[##2]%
        \setcounterref{rep@#1}{\lsuffix@label}%
        \addtocounter{rep@#1}{-1}%
        \begin{rep@#1}%
    }{%
        \end{rep@#1}%
    }%
}
\makeatother

\newtheorem{theorem}{Theorem}
\newreptheorem{theorem}{Theorem}
\newtheorem{lemma}{Lemma}
\newreptheorem{lemma}{Lemma}

\newtheorem{proposition}{Proposition}
\newreptheorem{proposition}{Proposition}

\newtheorem*{remark}{Remark} 
\makeatletter
\let\oldsubequations\subequations
\let\endoldsubequations\endsubequations
\NewDocumentEnvironment{newsubequations}{o}{
    \IfValueT{#1}{\addtocounter{equation}{-1}}%
    \oldsubequations%
    \IfValueT{#1}{%
        \renewcommand{\theparentequation}{#1}%
        \def\@currentlabel{#1}%
    }%
    \RenewDocumentCommand\subtag{o m}{\tag{\theparentequation\IfValueT{##1}{##1}##2}}%
}{\endoldsubequations\ignorespacesafterend}
\RenewDocumentCommand\subequations{o}{\newsubequations[#1]}
\RenewDocumentCommand\endsubequations{}{\endnewsubequations}
\def\subtagsep{}
\NewDocumentCommand\subtag{o m}{\tag{\theequation\IfValueTF{#1}{#1}{\subtagsep}#2}}
\NewDocumentCommand\itemizedeqs@begin{s m}{%
    \RenewDocumentCommand\item{s o O{t} O{}}{
        & \autoparbox[##3][##4][t]{\IfBooleanTF{##1}{\hphantom{\textbullet}}{\textbullet}\hspace{\labelsep}%
        \IfValueT{##2}{\begin{tabular}[t]{@{}l@{\hspace{\labelsep}}}##2\end{tabular}}} &%
    }%
    \let\eq\item
    \IfBooleanTF{#1}{
        \csname alignat*\endcsname%
    }{
        \subequations\alignat%
    }
    {\numexpr#2+1\relax}
}
\NewDocumentCommand\itemizedeqs@end{s o}{%
    \IfBooleanTF{#1}{
        \csname endalignat*\endcsname%
    }{
        \endalignat%
        \IfValueT{#2}{#2[!]}%
        \endsubequations%
    }%
}
\NewDocumentEnvironment{itemizedeqs}{m o}{\itemizedeqs@begin{#1}}{\itemizedeqs@end[#2]}
\NewDocumentEnvironment{itemizedeqs*}{m}{\itemizedeqs@begin*{#1}}{\itemizedeqs@end*}

\NewDocumentCommand\optim@min{o}{\smashoperator[lr]{\optim@op@min_{#1}}}
\NewDocumentCommand\optim@max{o}{\smashoperator[lr]{\optim@op@max_{#1}}}
\def\optim@st{\text{s.t.}}
\let\optim@op@min\min 
\let\optim@op@max\max
\NewDocumentCommand\optim@op@body{s o m}{
    &%
    \IfBooleanTF{#1}{
        \optim@max[#2]%
    }{
        \optim@min[#2]%
    }%
    &\;&%
    \begin{aligned}[t] #3 \end{aligned}%
}
\NewDocumentCommand\optim@st@body{m}{%
    \\%
    & \; \optim@st &&%
    \begin{alignedat}[t]{\value{MaxMatrixCols}} #1 \end{alignedat}%
}
\NewDocumentCommand\optim@begin{O{c}}{
    \def\optim@min@disp{\optim@op@body}
    \def\optim@max@disp{\optim@op@body*}%
    \subtoarg{min}{optim@min@disp}
    \subtoarg{max}{optim@max@disp}
    \NewDocumentCommand\st{m}{%
        \RenewDocumentCommand\st{}{}
        \optim@st@body{##1}%
    }%
    \alignedat[#1]{2}%
}
\NewDocumentCommand\optim@end{s}{%
    \endalignedat
}
\NewDocumentEnvironment{optim}{O{c}}{\optim@begin[#1]}{\optim@end}
\makeatother
\NewDocumentCommand\setdisplayskips{m m}{%
    \setlength\abovedisplayskip{#1}%
    \setlength\belowdisplayskip{#2}%
    \setlength\abovedisplayshortskip{#1}%
    \setlength\belowdisplayshortskip{#2}%
}

\newcommand\mail[1]{\Letter\,\href{mailto:#1}{#1}}
\NewDocumentCommand\AuthorCite{o m}{%
    \IfNoValueTF{#1}{
        \begin{NoHyper}\citeauthor{#2}\end{NoHyper}~\cite{#2}%
    }{
        \begin{NoHyper}\citeauthor{#1}\end{NoHyper}~\cite{#1,#2}%
    }%
}

\newcommand\sq[1]{`#1'} 
\newcommand\dq[1]{``#1''} 

\AtBeginDocument{
  \let\oldlabel\label
  \let\oldref\ref
  \let\label\suffixedlabel
  \let\ref\suffixedref
}

\makeatletter
\def\lsuffix@sep{-}
\def\lsuffix@root{root}
\def\lsuffix@old{!}
\def\lsuffix{\lsuffix@root}
\NewDocumentCommand\lsuffix@set{m o}{%
  \IfNoValueTF{#2}{
    \def\lsuffix@suf{\lsuffix}%
  }{\ifthenelse{\isempty{#2}}{
    \def\lsuffix@suf{\lsuffix@root}%
  }{
    \def\lsuffix@suf{#2}%
  }}%
  \IfStrEq{\lsuffix@old}{\lsuffix@suf}{%
    \def\lsuffix@label{#1}
  }{%
    \def\lsuffix@label{#1\lsuffix@sep\lsuffix@suf}
  }%
}
\NewDocumentCommand\suffixedlabel{m o}{%
  \lsuffix@set{#1}[#2]%
  \expandafter\oldlabel\expandafter{\lsuffix@label}%
}
\NewDocumentCommand\suffixedref{m o}{%
  \lsuffix@set{#1}[#2]%
  \expandafter\oldref\expandafter{\lsuffix@label}%
}
\newcommand{\setlabelsuffix}[1]{\def\lsuffix{#1}}
\newcommand{\unsetlabelsuffix}{\def\lsuffix{\lsuffix@root}}
\makeatother

\NewDocumentCommand\chapref{m o}{Chapter~\ref{#1}[#2]}
\NewDocumentCommand\appref{m o}{Appendix~\ref{#1}[#2]}
\NewDocumentCommand\secref{m o}{Section~\ref{#1}[#2]}
\NewDocumentCommand\ssecref{m o}{Subsection~\ref{#1}[#2]}
\NewDocumentCommand\figref{m o}{\figurename~\ref{#1}[#2]}
\NewDocumentCommand\tabref{m o}{Table~\ref{#1}[#2]}
\NewDocumentCommand\thmref{m o}{Theorem~\ref{#1}[#2]}
\NewDocumentCommand\lemref{m o}{Lemma~\ref{#1}[#2]}
\NewDocumentCommand\corref{m o}{Corollary~\ref{#1}[#2]}
\NewDocumentCommand\propref{m o}{Proposition~\ref{#1}[#2]}
\let\oldalgref\algref
\RenewDocumentCommand\algref{m g o}{Algorithm~\IfValueTF{#2}{\oldalgref{#1}{#2}}{\ref{#1}[#3]}}
\makeatletter
\RenewDocumentCommand\eqref{m}{\textup{\tagform@{\oldref{#1}}}}
\makeatother

\newboolean{blinded}
\setboolean{blinded}{false}
\NewDocumentCommand\blind{m o}{%
    \ifthenelse{\boolean{blinded}}{
        \IfNoValueTF{#2}{
            \xblackout{#1}%
        }{
            \xblackout{#2}%
        }%
    }{
        #1%
    }%
}
\NewDocumentCommand\blindfootnotehref{m m}{%
    \ifthenelse{\boolean{blinded}}{
        #2\footnote{\blind{#1}}%
    }{
        \href{#1}{#2}\footnote{\url{#1}}%
    }%
}

\newcommand\timesub[1][]{%
    \ifx#1\empty%
    \else%
        _{t%
        \IfInteger{#1}{%
            \ifnum0<0#1
                +#1%
            \else\ifnum0=#1
            \else
                #1%
            \fi\fi%
        }{
            +#1%
        }%
        }%
    \fi%
}
\newcommand\s[1][]{
    \bm{s}\timesub[#1]%
}
\renewcommand\a[1][]{
    \bm{a}\timesub[#1]%
}

\renewcommand\S{\mathcal{S}}
\newcommand\A{\mathcal{A}}
\newcommand\p{\bm{\theta}}
\newcommand\exptup{\tuple{\s[0],\a[0],r_t^{},\s[1]}}
\newcommand\buf{\mathcal{B}}

\newcommand\rc{r\msub{\textsc{c}}}
\newcommand\rav{\bar{r}}

\setboolean{blinded}{false}

\usepackage[margin=1.6cm]{geometry}
\date{}
\newcommand\submitinfo{}
\newcommand\editorinfo{}
\let\oldmaketitle\maketitle
\renewcommand\maketitle{%
    \begingroup%
    \def\thefootnote{}%
    \def\footnotemark{}%
    \oldmaketitle%
    \endgroup%
    \let\oldtabular\tabular%
    \renewcommand\tabular{\footnotesize\oldtabular}%
}
\providecommand\initenvtitle{}
\newtheorem*{initenv}{\small\initenvtitle}
\RenewDocumentEnvironment{abstract}{}{\renewcommand\initenvtitle{Abstract}\initenv\small}{\endinitenv}
\NewDocumentEnvironment{IEEEImpStatement}{}{\renewcommand\initenvtitle{Impact Statement}\initenv\small}{\endinitenv}
\NewDocumentEnvironment{IEEEkeywords}{}{\renewcommand\initenvtitle{Keywords}\initenv\small}{\endinitenv}
\let\olditemize\itemize
\let\endolditemize\enditemize
\RenewDocumentEnvironment{itemize}{o}{\olditemize}{\endolditemize}
\usepackage{caption}
\providecommand\IEEEmembership[1]{}
\ProvideDocumentEnvironment{IEEEproof}{}{\begin{proof}}{\end{proof}}
\providecommand\IEEEPARstart[2]{#1#2}
\providecommand\appendices{%
    \gdef\thesection{\Alph{section}}%
    \setcounter{section}{0}%
    \setcounter{subsection}{0}%
    \setcounter{subsubsection}{0}%
    \setcounter{paragraph}{0}%
    \section*{Appendices}%
}

\begin{document}

\title{A Dual Perspective of Reinforcement Learning for \\ Imposing Policy Constraints}

\author{\blind{Bram De Cooman}\textsuperscript{(\Letter)}, and \blind{Johan Suykens\IEEEmembership{, Fellow, IEEE}}%
\submitinfo%
\thanks{\blind{The research leading to these results has received funding from the European Research Council under the European Union's Horizon 2020 research and innovation program / ERC Advanced Grant E-DUALITY (787960). This paper reflects only the authors' views and the Union is not liable for any use that may be made of the contained information.}}%
\thanks{\blind{This research received funding from the Flemish Government (AI Research Program). Johan Suykens and Bram De Cooman are also affiliated to Leuven.AI - KU Leuven institute for AI, B-3000, Leuven, Belgium.}}%
\thanks{\blind{B. De Cooman (\mail{bram.decooman@esat.kuleuven.be}) and J. Suykens (\mail{johan.suykens@esat.kuleuven.be})}[XXX]\blind{ are with STADIUS Center for Dynamical Systems, Signal Processing and Data Analytics, Department of Electrical Engineering (ESAT), KU Leuven, 3001 Leuven, Belgium.}}%
\editorinfo}

\maketitle

\begin{abstract}
Model-free reinforcement learning methods lack an inherent mechanism to impose behavioural constraints on the trained policies.
Although certain extensions exist, they remain limited to specific types of constraints, such as value constraints with additional reward signals or visitation density constraints.
In this work we unify these existing techniques and bridge the gap with classical optimization and control theory, using a generic primal-dual framework for value-based and actor-critic reinforcement learning methods.
The obtained dual formulations turn out to be especially useful for imposing additional constraints on the learned policy, as an intrinsic relationship between such dual constraints (or regularization terms) and reward modifications in the primal is revealed.
Furthermore, using this framework, we are able to introduce some novel types of constraints, allowing to impose bounds on the policy's action density or on costs associated with transitions between consecutive states and actions.
From the adjusted primal-dual optimization problems, a practical algorithm is derived that supports various combinations of policy constraints that are automatically handled throughout training using trainable reward modifications.
The proposed \texttt{DualCRL} method is examined in more detail and evaluated under different (combinations of) constraints on two interpretable environments.
The results highlight the efficacy of the method, which ultimately provides the designer of such systems with a versatile toolbox of possible policy constraints.
\end{abstract}

\begin{IEEEImpStatement}
For complex control tasks it is often hard to design a suitable, scalar reward signal that leads to desirable controller behaviour for all relevant states.
Manual tuning of the reward function to accommodate certain behaviour in a particular region of interest can be very time consuming and error-prone, as it can lead to unexpected or undesirable behaviour in other parts of the state-action space.
A more practical and natural approach would thus be to keep the reward function as simple as possible, while imposing some additional constraints on the policy's behaviour.
In this work we focus on dual formulations of popular reinforcement learning methods, where it is often easier to enforce such additional constraints.
The proposed \texttt{DualCRL} algorithms can therefore lower the burden on designers of such systems, allowing them to specify policy constraints that are automatically taken care of throughout training.
\end{IEEEImpStatement}


\section{Introduction}
\IEEEPARstart{T}{he} field of convex optimization has a rich body of work on duality theory. Depending on the application, optimization problems can be simplified or more easily solved in either their primal or dual formulations and it is often more straightforward to extend the problem --- by adding extra regularization terms to the objective function or extra constraints --- in certain representations.
Duality has also been explored in reinforcement learning literature, however it remains limited to applications within certain domains, such as constrained~\cite{Bhatnagar2012, Paternain2019, Tessler2018, Stooke2020}, offline~\cite{Nachum2019, Nachum2019a, Nachum2020, Polosky2022} or density constrained~\cite{Chen2019, Qin2021} reinforcement learning.

\begin{figure}
    \centerline{\includegraphics[width=0.77\linewidth]{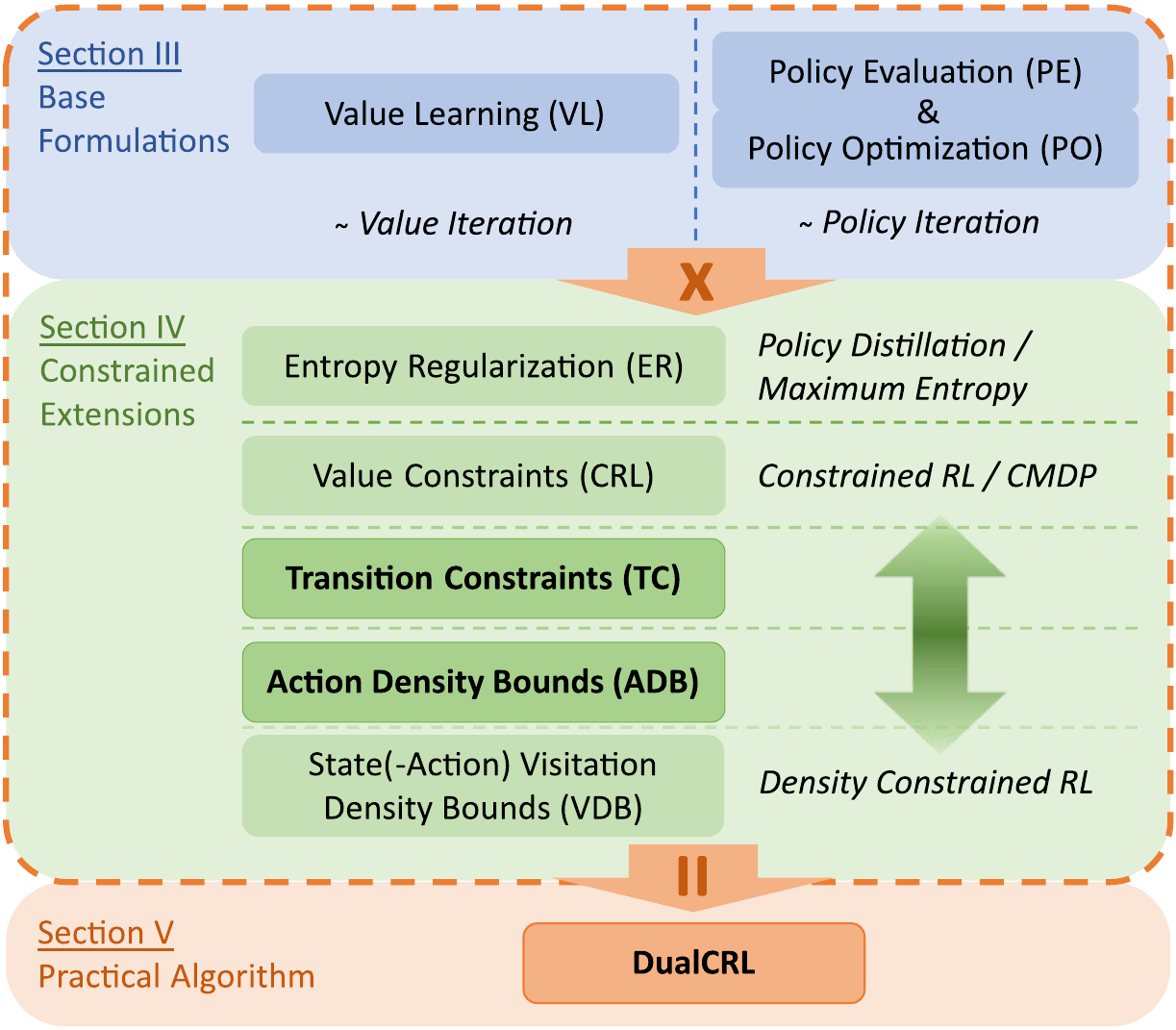}}
    \caption{Overview of the introduced framework's structure and the discussed optimization problems with their relation to different reinforcement learning domains. The base primal-dual formulations are combined with any constraint(s) using \texttt{DualCRL}. Novel contributions are highlighted in bold.}
	\label{fig:overview}
\end{figure}

\paragraph*{Research Objective}
In this paper, we aim at putting forward a more generic and complete primal-dual framework applicable to both value-based and actor-critic reinforcement learning methods. We show how the dual formulation is particularly useful to impose additional constraints on the learned policies and how this framework unifies seemingly different reinforcement learning fields. \figref{fig:overview} gives a general overview of the framework and its relation to diverse domains.

\paragraph*{Contributions}
Our main contributions can be summarized as follows. We provide a clear derivation of optimization problems for both value-based and actor-critic reinforcement learning methods under various policy constraints. Using this unifying primal-dual framework, we can provide novel insights into the intrinsic relationship between learnable reward modifications in the primal and policy constraints in the dual. Additionally, we extend the scala of policy constraints by introducing novel transition and action density constraints. This offers the designer a broad toolbox of policy constraints, acting on different scopes in the trajectory, bridging the gap between value and visitation constraints (\figref{fig:overview}). Our results closely align with some existing methods in the entropy regularized and constrained reinforcement learning settings. However, to the best of our knowledge, our \texttt{DualCRL} algorithm is the first to seamlessly combine different constraint types.

\paragraph*{Organization}
The necessary reinforcement learning essentials are briefly revised in \secref{sec:background}, after which the basic optimization problems that are used as the building blocks of our framework are introduced in \secref{sec:op}. These are then built upon and extended using various policy constraints in \secref{sec:ext}. This is followed by the introduction of the practical \texttt{DualCRL} algorithm in \secref{sec:practical}. Finally, in \secref{sec:experiments} the presented methods and constraints are investigated in more detail on two interpretable environments. The related work is discussed before the conclusion in \secref{sec:rel}. \appref{app:theory}[app] provides a proof for all Theorems, Lemmas and Propositions.

\section{Background}\label{sec:background}
In this section we briefly revise the basic concepts of Reinforcement Learning (RL) used throughout this work, for a more in-depth introduction the reader is referred to \AuthorCite{Sutton2018} and \AuthorCite{Altman1999}. \tabref{table:symbols} summarizes the most important symbols and notations used throughout this work.

\paragraph*{Reinforcement Learning}
We consider the Markov Decision Process (MDP)~\cite{Puterman1994} $\mathcal{M} = \tuple{\S, \A, \iota, \tau, r, \gamma}$ with state space $\S$, action space $\A$, initial-state distribution $\iota(\s_0)$, state-transition distribution $\tau(\s[1] | \s[0], \a[0])$, reward function $r(\s[0], \a[0], \s[1])$ and discount factor $\gamma \in \coiv{0}{1}$; and a policy $\pi(\a | \s)$ acting in this MDP. The expected discounted return of a policy $\pi(\a | \s)$ acting in this MDP is given by $\nu^\pi = \expect_{\iota, \pi, \tau}*{\sum_{t=0}^{\infty} \gamma^t r(\s[0], \a[0], \s[1])}$, where the expectation is taken over $\s_0 \sim \iota(\cdot)$, $\a[0] \sim \pi(\cdot | \s[0])$ and $\s[1] \sim \tau(\cdot | \s[0], \a[0])$ for all timesteps. The goal in reinforcement learning is then to find the optimal policy $\pi^*$, maximizing this expected discounted return, or (equivalently) the average reward $\rav^\pi = (1 - \gamma) \nu^\pi$,
\begin{equation}
    \pi^* = \argmax_\pi \expect_{\iota, \pi, \tau}*{\sum_{t=0}^{\infty} \gamma^t r(\s[0], \a[0], \s[1])}.
    \label{eq:rl_objective}
\end{equation}

\paragraph*{Constrained Reinforcement Learning}
Additional reward functions $r_k(\s[0], \a[0], \s[1])$, with $k$ an integer from the set $\bracks*{K}=\set{1, \dots, K}$, can be added to enforce $K$ additional constraints on the policy. This leads to a Constrained Markov Decision Process (CMDP)~\cite{Altman1999}, where the objective is to find an optimal policy $\pi^*$, maximizing the expected discounted return, subject to the $K$ extra constraints $\nu_k^\pi \ge V_k$, i.e.
\begin{equation}
    \begin{optim}
        \max_\pi{\expect_{\iota, \pi, \tau}*{\sum_{t=0}^{\infty} \gamma^t r(\s[0], \a[0], \s[1])}}
        \st{\expect_{\iota, \pi, \tau}*{\sum_{t=0}^{\infty} \gamma^t r_k(\s[0], \a[0], \s[1])} \ge V_k &\quad& \forall k}
    \end{optim}
    \label{eq:crl_objective}
\end{equation}

\paragraph*{Entropy Regularized Reinforcement Learning}
The average reward objective can be extended with an extra regularization term, such as the learned (student) policy's entropy or the cross-entropy with another (teacher) policy $\pi_{\textsc{t}}$. The goal is still to maximize the average reward, while either acting as randomly as possible for the former \emph{maximum entropy} methods~\cite{Haarnoja2018}, or extracting as much knowledge from the teacher as possible for the latter \emph{policy distillation} methods~\cite{Czarnecki2019}.
Both objectives can also be combined by minimizing the KL-divergence between teacher and student policies, as $\KL{p}{q} = \entropy{p}{q} - \entropy{p}$, yielding the entropy regularized objective
\begin{equation}\scalemath[0.96]{
    \mspace{-10mu}\max_\pi \expect_{\iota, \pi, \tau}*{\sum_{t=0}^{\infty} \gamma^t r(\s[0], \a[0], \s[1]) - \alpha \KL[\big]{\pi(\cdot | \s[0])}{\pi_{\textsc{t}}(\cdot | \s[0])}}\!.\mspace{-10mu}
    \label{eq:er_objective}
}\end{equation}

\paragraph*{Value Function}
The value of a policy $\pi$ acting in the MDP for a certain state or state-action pair is given by $v^\pi(\s) = \expect_{\pi, \tau}{\sum_{k=0}^{\infty} \gamma^k r(\s[k], \a[k], \s[k+1]) \given {\s[0]=\s}}$ and $q^\pi(\s, \a) = \expect_{\pi, \tau}{\sum_{k=0}^{\infty} \gamma^k r(\s[k], \a[k], \s[k+1]) \given {\s[0]=\s},\, {\a[0]=\a}}$. From these value functions, the expected discounted return can be easily obtained as $\nu^\pi = \expect_{\s_0 \sim \iota(\cdot)}{v^\pi(\s_0)} = \expect_{\s_0 \sim \iota(\cdot), \a_0 \sim \pi(\cdot | \s_0)}{q^\pi(\s_0, \a_0)}$. Both value functions satisfy recursive relationships, also known as the Bellman equations, $v^\pi(\s) = \expect_{\pi, \tau}{r(\s[0], \a[0], \s[1]) + \gamma v^\pi(\s[1]) \given {\s[0]=\s}}$ and $q^\pi(\s, \a) = \expect_{\pi, \tau}{r(\s[0], \a[0], \s[1]) + \gamma q^\pi(\s[1], \a[1]) \given {\s[0]=\s},\, {\a[0]=\a}}$.
The value functions for the optimal policy $\pi^*$ are denoted by $v^*$ and $q^*$ and satisfy the Bellman optimality equations $v^*(\s) = \max_{\a} \expect_{\tau}{r(\s[0], \a[0], \s[1]) + \gamma v^*(\s[1]) \given {\s[0]=\s},\, {\a[0]=\a}}$ and $q^*(\s, \a) = \expect_{\tau}{r(\s[0], \a[0], \s[1]) + \gamma \max_{\a'} q^*(\s[1], \a') \given {\s[0]=\s},\, {\a[0]=\a}}$.

\paragraph*{Visitation Density}
The visitation density (also referred to as the occupation or frequency) of a policy $\pi$ acting in the MDP for a certain state is defined as $d^\pi(\s) = (1 - \gamma) \sum_{t=0}^{\infty} \gamma^t \prob[\big]{\s_t = \s \given \s_0 \sim \iota(\cdot), \forall t: \a_t \sim \pi(\cdot | \s_t), \s_{t+1} \sim \tau(\cdot | \s_t, \a_t)}$, providing the discounted probability (density) of visiting state $\s$ at any possible timestep. Using this definition, the average reward can be expressed as $\rav^\pi = \expect_{\s \sim d^\pi(\cdot), \a \sim \pi(\cdot | \s), \s' \sim \tau(\cdot | \s, \a)}{r(\s, \a, \s')}$. The visitation density also satisfies a recursive relation $d^\pi(\s) = (1 - \gamma) \iota(\s) + \gamma \sum_{\s' \in \S}\sum_{\a' \in \A} d^\pi(\s') \pi(\a' | \s') \tau(\s | \s', \a')$, and the visitation density of a policy for state-action pairs can be written as $p^\pi(\s, \a) = d^\pi(\s) \pi(\a | \s)$.
To refer to the set of states visited by a policy $\pi$, we use the notation $\S^\pi = \set{\s \in \S \given d^\pi(\s) > 0}$, whereas the set of actions taken by a policy in a certain state $\s$ is denoted by $\A^\pi(\s) = \set{\a \in \A \given \pi(\a | \s) > 0}$.

\paragraph*{Model-Free Reinforcement Learning}
In the model-free RL setting, several parameters of the optimization problem \eqref{eq:rl_objective}, such as $\iota$, $\tau$ and $r$ are unknown, but we assume access to samples from these unknown distributions and functions.
More specifically, a replay buffer $\buf$ is available, containing experience samples $\set{\exptup}$, where $\s[0] \sim d^\beta(\cdot)$, $\a[0] \sim \beta(\cdot | \s[0])$, $\s[1] \sim \tau(\cdot | \s[0], \a[0])$ and $r_t = r(\s[0], \a[0], \s[1])$.
For on-policy methods, the behavioral policy $\beta$ is equal to the policy being learned $\pi$; for off-policy methods they might differ, for example to encourage sufficient exploration.

\begin{table*}[!t]
    \centering
	\caption{Overview of the most important symbols and notations.}
	\label{table:symbols}
    \hfill
	\begin{tabular}{@{\,} >{\raggedright}m{11em} >{$}l<{$} @{\,}}
		\toprule
		\textbf{Description} & \textbf{Notation} \\
		\midrule
        State space & \S \\
        Action space & \A \\
        State at timestep $t$ & \s[0] \in \S \\
        Action at timestep $t$ & \a[0] \in \A \\
        Reward function & r(\s[0], \a[0], \s[1]) \\
        Reward at timestep $t$ & r_t^{} \\
        Initial state distribution & \iota(\s_0) \\
        State transition distribution & \tau(\s[1] | \s[0], \a[0]) \\
        Experience tuple & \exptup \\
		\bottomrule
        & \\
	\end{tabular}
    \hfill
	\begin{tabular}{@{\,} >{\raggedright}m{13em} >{$}l<{$} @{\,}}
		\toprule
		\textbf{Description} & \textbf{Notation} \\
		\midrule
        Policy & \pi(\a[0] | \s[0]) \\
        Behavioural policy & \beta(\a[0] | \s[0]) \\
        Replay buffer & \buf \\
        Discount factor & \gamma \\
        Expected discounted return & \nu^\pi \\
        Average reward & \rav^\pi \\
        State value function & v^\pi(\s) \\
        State-action value function & q^\pi(\s, \a) \\
        State visitation density & d^\pi(\s) \\
        State-action visitation density & p^\pi(\s, \a) \\
		\bottomrule
	\end{tabular}
    \hfill
	\begin{tabular}{@{\,} >{\raggedright}m{8em} >{$}l<{$} @{\,}}
		\toprule
		\textbf{Description} & \textbf{Notation} \\
		\midrule
        KL-divergence & \KL{p}{q} \\
        Cross-entropy & \entropy{p}{q} \\
        Entropy & \entropy{p} \\
        Expectation & \expect_{x \sim p}{f(x)} \\
        Indicator function & \ind_X(x) \\
        Lagrangian & \mathcal{L}(x, \lambda, \mu) \\
        $\set{1, \dots, N} \subset \mathbb{N}$ & \bracks*{N} \\
        Parameter vector & \p_i \\
        Loss & L_i(\p_i) \\
        Learning rate & \eta_i \\
		\bottomrule
	\end{tabular}
    \hfill{}
\end{table*}

\section{Optimization Perspective}\label{sec:op}
In this work we analyze two optimization problems related to the various existing reinforcement learning algorithms to solve \eqref{eq:rl_objective}. More specifically, a Linear Programming (LP) formulation used for value learning and a minimax formulation used for policy learning are analyzed. The former formulation is related to the value iteration scheme and value-based RL methods, whereas the latter formulation is related to the policy iteration scheme and actor-critic RL methods.
To simplify the derivations, only discrete and finite-dimensional state and action spaces are considered in this work, i.e. $\S = \bracks{S}$ and $\A = \bracks{A}$. The experiments in \secref{sec:experiments} illustrate the application to continuous state and action spaces. \appref{app:theory}[app] provides a proof for all Theorems and Lemmas.

\subsection{Value Learning LP}
Value-based reinforcement learning methods solve \eqref{eq:rl_objective} indirectly by first finding the optimal value function $q^*(\s, \a)$ and then obtaining an optimal (deterministic) policy as $\pi(\a | \s) = \ind_{\A^*(\s)}(\a) / \abs{\A^*(\s)}$ with $\A^*(\s) = \argmax_{\a} q^*(\s, \a)$ the set of optimal actions for state $\s$. The popular DQN method~\cite{Mnih2015} (and its later extensions) follow such a strategy. This approach can be written as the following LP
\begin{subequations}[VL]
    \label{eq:vl}[!]
    \begin{align}
        &\begin{optim}
            \min_{\substack{v(\s) \\ \smash[b]{q(\s, \a)}}}{\expect[r]_{\;\s_0 \sim \iota(\cdot)}[\big]{(1 - \gamma) v(\s_0)}}
            \st{
                v(\s) &\ge q(\s, \a) &\;& \forall \s \, \forall \a \\
                q(\s, \a) &= \expect[lr]_{\qquad\s' \sim \tau(\cdot | \s, \a)}[\big]{r(\s, \a, \s') + \gamma v(\s')} && \forall \s \, \forall \a
            }
        \end{optim}
        \label{eq:vl_prim}\subtag[\textperiodcentered]{P} \\
        \intertext{and has as dual}
        &\begin{optim}
            \max_{\substack{d(\s) \\ p(\s, \a)}}{\expect[r]_{\subalign{(\s, \a) &\sim p(\cdot, \cdot) \\ \s' &\sim \tau(\cdot | \s, \a)}\mspace{-20mu}}[\big]{r(\s, \a, \s')}}
            \st{
                \sum_{\a \in \A} p(\s, \a) &= d(\s) &\;& \forall \s \\[-6pt]
                d(\s) &= (1 - \gamma) \iota(\s) \\
                &\mspace{40mu}+ \gamma \quad \expect[lr]_{(\s', \a') \sim p(\cdot, \cdot)\;}[\big]{\tau(\s | \s', \a')} && \forall \s \\[-6pt]
                p(\s, \a) &\ge 0 && \forall \s \, \forall \a
            }
        \end{optim}\mspace{-10mu}
        \label{eq:vl_dual}\subtag[\textperiodcentered]{D}
    \end{align}
\end{subequations}
This formulation is a slightly extended version of the primal and dual formulations commonly referred to as the $V$-LP~\cite{Puterman1994}. The extra $q(\s, \a)$ and $d(\s)$ primal and dual decision variables, introduced here, could easily be eliminated but will prove to be useful when altering the optimization problems in \secref{sec:ext}.
The following properties can be proven for this LP, using the optimality (KKT) conditions.
\begin{theorem}\label{thm:vl_dual}
    The policy that can be derived from the optimal dual variables, is an optimal policy for the considered MDP.
\end{theorem}
\begin{lemma}\label{lem:vl_prim}
    The optimal primal variables satisfy the Bellman optimality equation and are thus equivalent to the optimal value functions $v^*$ and $q^*$ of the considered MDP.
\end{lemma}
\appref{app:vl}[app] provides a proof and derivation of the Lagrangian objective $\mathcal{L}_{\mathrm{VL}}$.
The primal and dual objectives of \eqref{eq:vl} clearly show the two equivalent formulations for the average reward (or expected discounted return) of the optimal policy
\begin{equation*}
    \rav^* = (1 - \gamma) \nu^* = \expect[lr]_{\;\s \sim \iota(\cdot)}[\big]{(1 - \gamma) v^*(\s)} = \mspace{16mu}\expect[lr]_{\subalign{(\s, \a) &\sim p^*(\cdot, \cdot) \\ \s' &\sim \tau(\cdot | \s, \a)}\mspace{-20mu}}[\big]{r(\s, \a, \s')}.
\end{equation*}
Note that it can be written as an equality because strong duality holds for LPs (there is no duality gap).

\subsection{Policy Evaluation LP}
A closely related linear optimization problem can be defined for the task of policy evaluation, where the goal is to find the value function $q^\pi$ of a given (fixed) policy $\pi$. This can be done by solving the following LP
\begin{subequations}[PE]
    \label{eq:pe}[!]
    \begin{align}
        &\begin{optim}
            \min_{\substack{v(\s) \\ \smash[b]{q(\s, \a)}}}{\expect[r]_{\;\s_0 \sim \iota(\cdot)}[\big]{(1 - \gamma) v(\s_0)}}
            \st{
                v(\s) &\ge \expect[lr]_{\mspace{32mu}\a \sim \pi(\cdot | \s)}[\big]{q(\s, \a)} &\;& \forall \s \\
                q(\s, \a) &= \expect[lr]_{\qquad\s' \sim \tau(\cdot | \s, \a)}[\big]{r(\s, \a, \s') + \gamma v(\s')} && \forall \s \, \forall \a
            }
        \end{optim}
        \label{eq:pe_prim}\subtag[\textperiodcentered]{P} \\
        \intertext{with dual}
        &\begin{optim}
            \max_{\substack{d(\s) \\ p(\s, \a)}}{\expect[r]_{\subalign{(\s, \a) &\sim p(\cdot, \cdot) \\ \s' &\sim \tau(\cdot | \s, \a)}\mspace{-20mu}}[\big]{r(\s, \a, \s')}}
            \st{
                d(\s) &= (1 - \gamma) \iota(\s) \\
                &\mspace{40mu}+ \gamma \quad \expect[lr]_{(\s', \a') \sim p(\cdot, \cdot)\;}[\big]{\tau(\s | \s', \a')} \ge 0 &\;& \forall \s \\
                p(\s, \a) &= d(\s) \pi(\a | \s) && \forall \s \, \forall \a
            }
        \end{optim}
        \label{eq:pe_dual}\subtag[\textperiodcentered]{D}
    \end{align}
\end{subequations}
The following properties can be proven for this LP, using the optimality (KKT) conditions.
\begin{lemma}\label{lem:pe_dual}
    The optimal dual variables are the visitation densities of the given policy $\pi$ for the considered MDP.
\end{lemma}
\begin{lemma}\label{lem:pe_prim}
    The optimal primal variables satisfy the Bellman equation and are equivalent to the value functions $v^\pi$ and $q^\pi$ of the given policy $\pi$ for the considered MDP.
\end{lemma}
\appref{app:pe}[app] provides a proof and derivation of the Lagrangian objective $\mathcal{L}_{\mathrm{PE}}$.
Because of strong duality (no duality gap for LPs), the average reward (or expected discounted return) of policy $\pi$ can be written as the solution of the constrained primal or dual LPs
\begin{equation*}
    \rav^\pi = (1 - \gamma) \nu^\pi = \expect[lr]_{\s \sim \iota(\cdot)}[\big]{(1 - \gamma) v^\pi(\s)} = \mspace{16mu}\expect[lr]_{\subalign{(\s, \a) &\sim p^\pi(\cdot, \cdot) \\ \s' &\sim \tau(\cdot | \s, \a)}\mspace{-20mu}}{r(\s, \a, \s')},
\end{equation*}
or as the solution of the unconstrained Lagrangian optimization problem
\begin{equation*}
    \rav^\pi = (1 - \gamma) \nu^\pi = \max_{\substack{d(\s) \\ p(\s, \a)}} \min_{\substack{v(\s) \\ q(\s, \a)}} \mathcal{L}_{\mathrm{PE}} = \min_{\substack{v(\s) \\ q(\s, \a)}} \max_{\substack{d(\s) \\ p(\s, \a)}} \mathcal{L}_{\mathrm{PE}}.
\end{equation*}

\subsection{Policy Optimization}
The optimal policy $\pi^*$ for the considered MDP \eqref{eq:rl_objective} can also be found iteratively using a two-step scheme called \emph{generalized policy iteration} (GPI). In the first \emph{policy evaluation} step, the value of a current policy $\pi$ is determined, which is then used in the second \emph{policy improvement} step to find a better policy. Repeatedly iterating over the two steps, eventually results in an optimal policy. Some popular methods in this category are TD3~\cite{Fujimoto2018} and SAC~\cite{Haarnoja2019}. The GPI scheme can be written as a nested optimization problem, where the first policy evaluation step corresponds to an inner policy evaluation LP \eqref{eq:pe}, and the second policy improvement step corresponds to an outer policy optimization problem \eqref{eq:po}.
\begin{equation}
    \begin{optim}
        \max_{\pi(\a | \s)}{\rav^\pi}
        \st{
            \sum_{\a \in \A} \pi(\a | \s) &= 1 &\;& \forall \s \\[-6pt]
            \pi(\a | \s) &\ge 0 && \forall \s \, \forall \a
        }
    \end{optim}
    \label{eq:po}\tag{PO}
\end{equation}
\appref{app:po}[app] provides more details and derivations for this nested optimization problem. The most important result is summarized below.
\begin{theorem}\label{thm:po}
    The optimal solution of the nested Policy Optimization \eqref{eq:po} and Policy Evaluation \eqref{eq:pe} problems, is the optimal policy $\pi^*$ for the considered MDP.
\end{theorem}

\section{Extending the Primal-Dual Formulations}\label{sec:ext}
We are now ready to alter the dual optimization problems described in the previous section and reveal the effect on the primal formulations and optimal solutions. The dual formulations \eqref{eq:vl_dual} and \eqref{eq:pe_dual} are especially useful for imposing additional constraints on the policy. For example, the addition of an extra regularization term to the dual objective, reveals a link with the policy distillation and maximum entropy RL setting. Furthermore, extra dual constraints allow us to solve constrained RL problems (CMDPs) or to bound visitation and action densities to ensure that the policy does (not) visit certain states or actions. Alternatively, we can impose transition constraints to prevent or encourage the agent to transition between two consecutive states or actions.
Derivations and proofs for the various Theorems, Lemmas and Propositions introduced in this section are provided in Appendices \ref{app:er}[app] - \ref{app:tc}[app].

\subsection{Entropy Regularization}
To mimic a given teacher policy $\pi_{\textsc{t}}$ and encourage exploration, the entropy regularized objective \eqref{eq:er_objective} can be used. By expressing this objective in terms of the visitation densities $d^\pi$ and $p^\pi$, it can replace the policy evaluation dual objective of \eqref{eq:pe_dual}.
The resulting regularized dual LP is thus obtained as
\begin{subequations}[ER]
    \label{eq:er}[!]
    \begin{align}
        &\mspace{-10mu}\begin{optim}
            \max_{\substack{d(\s) \\ p(\s, \a)}}{\expect[r]_{\subalign{(\s, \a) &\sim p(\cdot, \cdot) \\ \s' &\sim \tau(\cdot | \s, \a)}\mspace{-20mu}}[\big]{r(\s, \a, \s')} - \alpha \expect[lr]_{\mspace{20mu}\s \sim d(\cdot)}[\big]{\KL[\big]{\pi(\cdot | \s)}{\pi_{\textsc{t}}(\cdot | \s)}}}
            \st{
                d(\s) &= (1 - \gamma) \iota(\s) \\
                &\mspace{40mu}+ \gamma \quad \expect[lr]_{(\s', \a') \sim p(\cdot, \cdot)\;}[\big]{\tau(\s | \s', \a')} \ge 0 &\;& \forall \s \\
                p(\s, \a) &= d(\s) \pi(\a | \s) && \forall \s \, \forall \a
            }
        \end{optim}\mspace{-20mu}
        \label{eq:er_dual}\subtag[\textperiodcentered]{D}
        \intertext{corresponding to the modified primal}
        &\mspace{-10mu}\begin{optim}
            \min_{\substack{v(\s) \\ \smash[b]{q(\s, \a)}}}{\expect[r]_{\;\s_0 \sim \iota(\cdot)}[\big]{(1 - \gamma) v(\s_0)}}
            \st{
                v(\s) &\ge \expect[l]_{\mspace{32mu}\a \sim \pi(\cdot | \s)}[\bigg]{q(\s, \a) - \alpha \log \frac{\pi(\a | \s)}{\pi_{\textsc{t}}(\a | \s)}} &\;& \forall \s \\
                q(\s, \a) &= \expect[lr]_{\qquad\s' \sim \tau(\cdot | \s, \a)}[\big]{r(\s, \a, \s') + \gamma v(\s')} && \forall \s \, \forall \a
            }
        \end{optim}\mspace{-20mu}
        \label{eq:er_prim}\subtag[\textperiodcentered]{P}
    \end{align}
\end{subequations}
\lemref{lem:pe_dual} still holds for this regularized Policy Evaluation problem, so the optimal dual variables again correspond to the visitation densities $d^\pi$ and $p^\pi$ of the considered policy.
Notice how the extra regularization term in the dual, leads to a modified reward
\begin{equation}
    r_{\mathrm{ER}}^{}(\s, \a, \s') = r(\s, \a, \s') - \alpha \log \frac{\pi(\a | \s)}{\pi_{\textsc{t}}(\a | \s)}
    \label{eq:er_mod}
\end{equation}
in the primal.
\lemref{lem:pe_prim} can be modified accordingly, taking the entropy regularization into account through adjusted value functions using this modified reward.
\begin{lemma}\label{lem:er_prim}
    The optimal primal variables satisfy the entropy regularized Bellman equation and are equivalent to the entropy regularized value functions $v_{\mathrm{ER}}^\pi$ and $q_{\mathrm{ER}}^\pi$ of the given policy $\pi$ for the considered MDP.
\end{lemma}

The optimal (Boltzmann) policy $\pi_{\mathrm{ER}}^*$ can then be found by plugging this regularized policy evaluation problem in the policy optimization problem \eqref{eq:po}, resulting in a \emph{Soft Policy Iteration} scheme~\cite{Haarnoja2018}.
\begin{theorem}\label{thm:er_pol}
    The optimal solution of the nested Policy Optimization \eqref{eq:po} and Entropy-Regularized Policy Evaluation \eqref{eq:er} problems, is a conditional Boltzmann policy $\pi_{\mathrm{ER}}^*$.
\end{theorem}

The resulting optimal policy maximizes the reward while staying close to the teacher policy. \AuthorCite{Schulman2018} showed the link between policy gradient methods and soft $Q$-learning in this entropy-regularized setting. Such regularized learning objectives for policy distillation were also investigated in the context of transfer and multitask learning~\cite{Parisotto2016, Schmitt2018, Czarnecki2019}.

For the special case of a uniform teaching policy $\pi_{\textsc{t}}(\a | \s) = 1 / \abs{\A}$, the dual's regularization term effectively simplifies to $+\alpha \expect_{\s \sim d(\cdot)}{\entropy{\pi(\cdot | \s)}}$ (up to a neglected constant $\alpha \log \abs{\A}$), which corresponds to the (plain) maximum entropy RL case~\cite{Haarnoja2018}. A popular deep RL algorithm in this field is \sq{Soft Actor-Critic} (SAC)~\cite{Haarnoja2019}.

The temperature parameter $\alpha$ determines how much influence the teacher policy has on the learned student policy. High temperatures lead to policies remaining closer to the teacher policy (or more random in the special maximum entropy case), whereas lower temperatures lead to policies moving closer to the optimal policy $\pi^*$ for the considered MDP. As $\alpha$ goes to $0$, the \dq{softmax} Boltzmann policy thus becomes a \dq{hardmax} greedy policy, and the primal and dual formulations become equivalent to the non-regularized Policy Evaluation LP \eqref{eq:pe}.
Typically the value of the temperature is decreased throughout training, ensuring that the policy gradually becomes more independent from the teacher and more deterministic.

\subsection{Constrained Reinforcement Learning}
The $K$ additional constraints of the CMDP \eqref{eq:crl_objective} can be expressed in terms of the visitation density $p^\pi(\s, \a)$ as $(1 - \gamma)^{-1} \expect_{(\s, \a) \sim p^\pi(\cdot, \cdot), \s' \sim \tau(\cdot | \s, \a)}{r_k(\s, \a, \s')} \ge V_k$. In this form, the constraints can be readily added to the dual formulations \eqref{eq:vl_dual} and \eqref{eq:pe_dual}. The constrained dual LP for value learning is then obtained as
\begin{subequations}[CRL]
    \label{eq:crl}[!]
    \begin{align}
        &\mspace{-8mu}\begin{optim}
            \max_{\substack{d(\s) \\ p(\s, \a)}}{\expect[r]_{\subalign{(\s, \a) &\sim p(\cdot, \cdot) \\ \s' &\sim \tau(\cdot | \s, \a)}\mspace{-20mu}}[\big]{r(\s, \a, \s')}}
            \st{
                \sum_{\a \in \A} p(\s, \a) &= d(\s) &\;& \forall \s \\[-6pt]
                d(\s) &= (1 - \gamma) \iota(\s) \\
                &\mspace{40mu}+ \gamma \quad \expect[lr]_{(\s', \a') \sim p(\cdot, \cdot)\;}[\big]{\tau(\s | \s', \a')} && \forall \s \\[-6pt]
                p(\s, \a) &\ge 0 && \forall \s \, \forall \a \\
                (1 - \gamma) V_k &\le \quad\expect[lr]_{\subalign{(\s, \a) &\sim p(\cdot, \cdot) \\ \s' &\sim \tau(\cdot | \s, \a)}\mspace{-20mu}}{r_k(\s, \a, \s')} && \forall k
            }
        \end{optim}\mspace{-10mu}
        \label{eq:crl_dual}\subtag[\textperiodcentered]{D}
        \intertext{and corresponds to the altered primal}
        &\mspace{-8mu}\begin{optim}[t]
            \min_{\substack{v(\s), w_k \\ \smash[b]{q(\s, \a)}}}{\;\expect[r]_{\;\s_0 \sim \iota(\cdot)}[\big]{(1 - \gamma) v(\s_0)} - \sum_{k = 0}^{K} w_k (1 - \gamma) V_k}
            \st{
                v(\s) &\ge q(\s, \a) &\;& \forall \s \, \forall \a \\
                q(\s, \a) &= \expect[blr]_{\vs{11pt}\qquad\s' \sim \tau(\cdot | \s, \a)}\bracks[\bigg]{r(\s, \a, \s') + \smash{\sum_{k = 0}^{K}} w_k r_k(\s, \a, \s') \\[-12pt]
                &\mspace{212mu} + \gamma v(\s')} && \forall \s \, \forall \a \\
                w_k &\ge 0 && \forall k
            }
        \end{optim}\mspace{-52mu}
        \label{eq:crl_prim}\subtag[\textperiodcentered]{P}
    \end{align}
\end{subequations}
In this primal formulation, the reward is modified to
\vspace*{-0.2\baselineskip}
\begin{equation}
    r_{\mathrm{CRL}}^{}(\s, \a, \s') = r(\s, \a, \s') + \sum\nolimits_{k = 0}^{K} w_k r_k(\s, \a, \s')
    \label{eq:crl_mod}
\end{equation}
with \emph{learnable} parameters $w_k \ge 0$.
If a feasible policy exists, satisfying all additional constraints, the following properties can be proven, extending \thmref{thm:vl_dual} and \lemref{lem:vl_prim}.
\begin{theorem}\label{thm:crl_dual}
    The policy that can be derived from the optimal dual variables is an optimal policy for the considered CMDP.
\end{theorem}
\begin{lemma}\label{lem:crl_prim}
    The optimal primal variables satisfy the Bellman optimality equation and are thus equivalent to the optimal adjusted value functions $v_{\mathrm{CRL}}^*$ and $q_{\mathrm{CRL}}^*$ for the considered MDP with optimal modified reward $r_{\mathrm{CRL}}^*$.
\end{lemma}
These optimal adjusted value functions and optimal reward modifications also satisfy following properties.
\begin{theorem}\label{thm:crl_greedy}
    The optimal policy is greedy with respect to the optimal adjusted value functions $v_{\mathrm{CRL}}^*$ and $q_{\mathrm{CRL}}^*$.
\end{theorem}
\begin{proposition}\label{prop:crl_mod}
    The optimal modified reward is only altered by the tight constraints $V_k = \nu_k^{\pi^*}$.
\end{proposition}
On the other hand, if there is no policy that can satisfy all additional $K$ constraints, the dual problem effectively becomes infeasible and the primal becomes unbounded, caused by the $w_k$ of the infeasible constraints approaching infinity.

Remark that the simplified version of this constrained value learning LP was also covered by \AuthorCite{Altman1999} in the context of CMDPs, where the primal and dual formulations are typically swapped. More recently, the Lagrangian method for solving constrained RL tasks has also been used to derive several actor-critic algorithms~\cite{Bhatnagar2012, Paternain2019, Stooke2020}, such as RCPO~\cite{Tessler2018}. Such approaches can be related to an equivalent constrained formulation for policy evaluation \eqref{eq:pe} to derive constrained GPI schemes. This is illustrated next for visitation density bounds.

\subsection{Visitation Density Bounds}
To ensure that the optimal policy does (not) visit certain states or state-action pairs with a certain (discounted) probability, visitation density bounds can be added to the dual formulations \eqref{eq:vl_dual} or \eqref{eq:pe_dual}. Such bounds can be imposed both on the state-action visitation density, $\lb{p}(\s, \a) \le p^\pi(\s, \a) \le \ub{p}(\s, \a)$, and on the (marginal) state-visitation density, $\lb{d}(\s) \le d^\pi(\s) \le \ub{d}(\s)$.
As $d^\pi(\s)$ and $p^\pi(\s, \a)$ are proper probability density functions, the lower and upper bounds should satisfy $\sum_{\s \in \S} \lb{d}(\s) \le 1$, $\sum_{\s \in \S} \ub{d}(\s) \ge 1$, $\sum_{\s \in \S} \sum_{\a \in \A} \lb{p}(\s, \a) \le 1$ and $\sum_{\s \in \S} \sum_{\a \in \A} \ub{p}(\s, \a) \ge 1$. These conditions on the bounds are however not sufficient to guarantee feasibility of the constrained optimization problems, as this depends on the dynamics of the considered MDP.

\appref{app:vdb}[app] provides primal and dual formulations of \eqref{eq:vl} with bounds on the marginal \emph{state} visitation density, whereas here we focus on the primal and dual formulations of \eqref{eq:pe} with \emph{state-action} visitation density bounds. Both types of constraints can however be applied to either optimization problem. The idea of imposing bounds on the state visitation density was also explored by \AuthorCite{Chen2019}, who related it to optimal control, and \AuthorCite{Qin2021}. The formulations in this and the following section, where bounds on the conditional action density are imposed, can be regarded as an extension of the (density) constrained RL field (\figref{fig:overview}).

The primal and dual formulations of \eqref{eq:pe} with state-action visitation density bounds are given by
\begin{subequations}[VDB]
    \label{eq:vdb_pe}[!]
    \begin{align}
        &\begin{optim}
            \min_{\substack{v(\s) \\ q(\s, \a) \\ \lb{r}(\s, \a) \\ \ub{r}(\s, \a)}}{\;\expect[r]_{\;\s_0 \sim \iota(\cdot)}[\big]{(1 - \gamma) v(\s_0)} - \sum_{\s \in \S} \sum_{\a \in \A} \lb{r}(\s, \a) \lb{p}(\s, \a) \\
            + \sum_{\s \in \S} \sum_{\a \in \A} \ub{r}(\s, \a) \ub{p}(\s, \a)}
            \st{
                v(\s) &\ge \expect[lr]_{\mspace{32mu}\a \sim \pi(\cdot | \s)}[\big]{q(\s, \a)} &\;& \forall \s \\
                q(\s, \a) &= \expect[blr]_{\qquad\s' \sim \tau(\cdot | \s, \a)}\bracks[\big]{r(\s, \a, \s') + \lb{r}(\s, \a) \\
                &\mspace{128mu} - \ub{r}(\s, \a) + \gamma v(\s')} && \forall \s \, \forall \a \\
                \lb{r}(\s, \a) &\ge 0 \land \ub{r}(\s, \a) \ge 0 && \forall \s \, \forall \a
            }
        \end{optim}
        \label{eq:vdb_pe_prim}\subtag[\textperiodcentered]{P}
        \intertext{and}
        &\begin{optim}
            \max_{\substack{d(\s) \\ p(\s, \a)}}{\expect[r]_{\subalign{(\s, \a) &\sim p(\cdot, \cdot) \\ \s' &\sim \tau(\cdot | \s, \a)}\mspace{-20mu}}[\big]{r(\s, \a, \s')}}
            \st{
                d(\s) &= (1 - \gamma) \iota(\s) \\
                &\mspace{40mu}+ \gamma \quad \expect[lr]_{(\s', \a') \sim p(\cdot, \cdot)\;}[\big]{\tau(\s | \s', \a')} \ge 0 &\;& \forall \s \\
                p(\s, \a) &= d(\s) \pi(\a | \s) && \forall \s \, \forall \a \\
                \lb{p}(\s, \a) &\le p(\s, \a) \le \ub{p}(\s, \a) && \forall \s \, \forall \a
            }
        \end{optim}\mspace{-20mu}
        \label{eq:vdb_pe_dual}\subtag[\textperiodcentered]{D}
    \end{align}
\end{subequations}
As before, the reward is modified in the primal and becomes
\begin{equation}
    r_{\mathrm{VDB}}^{}(\s, \a, \s') = r(\s, \a, \s') + \lb{r}(\s, \a) - \ub{r}(\s, \a).
    \label{eq:vdb_mod}
\end{equation}
Remark the \emph{automatic} adjustment of the reward signal, through the learnable decision variables $\lb{r}(\s, \a)$ and $\ub{r}(\s, \a)$. Intuitively, this modified reward can be understood as follows: to ensure a minimum visitation of a state-action pair, the reward can be increased by $\lb{r}(\s, \a)$; whereas to ensure a maximum visitation of a state-action pair, the reward can be decreased by $\ub{r}(\s, \a)$.

Let us first consider the case of a feasible policy, satisfying all imposed density constraints.
Then, \lemref{lem:pe_dual} still holds for this constrained Policy Evaluation problem, so the optimal dual variables again correspond to the visitation densities $d^\pi$ and $p^\pi$ of the considered policy.
\lemref{lem:pe_prim} can be modified to take the visitation density constraints into account through adjusted value functions using the modified reward.
\begin{lemma}\label{lem:vdb_prim}
    The optimal primal variables satisfy the Bellman equation and are equivalent to the adjusted value functions $v_{\mathrm{VDB}}^\pi$ and $q_{\mathrm{VDB}}^\pi$ of the given policy $\pi$ for the considered MDP with optimal modified reward $r_{\mathrm{VDB}}^*$.
\end{lemma}
These optimal reward modifications satisfy following property.
\begin{proposition}\label{prop:vdb_mod}
    The optimal modified reward is only altered by the tight bounds $\lb{p}(\s, \a) = p^{\pi}(\s, \a)$ and $p^{\pi}(\s, \a) = \ub{p}(\s, \a)$.
\end{proposition}

In the other case, if the policy does not satisfy all additional constraints, the dual problem effectively becomes infeasible and the primal becomes unbounded, caused by the $\lb{r}(\s, \a)$ and $\ub{r}(\s, \a)$ of the infeasible constraints approaching infinity.
It might thus seem that this altered Policy Evaluation LP is not really useful on its own. Its utility, however, becomes clear when combining it with the Policy Optimization problem to find optimal constrained policies.
\begin{theorem}\label{thm:vdb_pol}
    The optimal solution of the nested Policy Optimization \eqref{eq:po} and constrained Policy Evaluation \eqref{eq:vdb_pe} problems, is the optimal policy $\pi^*$ for the considered MDP with visitation density bounds.
\end{theorem}
The resulting nested optimization problem can be seen as a \emph{Constrained} GPI scheme, in which the policy, adjusted value functions and reward modifications are jointly learned.
Although the inner Policy Evaluation problem diverges for an infeasible policy, violating the visitation density bounds, the directions in which the learnable reward parameters $\lb{r}$ and $\ub{r}$ are updated, remain useful for the combined policy iteration scheme: the reward is effectively increased for \emph{undervisited} state-action pairs, while it is decreased for \emph{overvisited} state-action pairs. As a result, during subsequent policy optimization steps, there is an increased incentive for the policy to visit the undervisited states and actions, while staying away from the overvisited states and actions.
The reward is thus modified in such a way that constraint-violating policies are no longer optimal or greedy with respect to the adjusted value functions.

\subsection{Action Density Bounds}
With a similar trick, one can also impose bounds on the conditional action density of the learned policy. In this case, the introduced constraints are $d^\pi(\s) \lb{\pi}(\a | \s) \le p^\pi(\s, \a) \le d^\pi(\s) \ub{\pi}(\a | \s)$ and the restrictions on the bounds become $\sum_{\a \in \A} \lb{\pi}(\a | \s) \le 1$, $\sum_{\a \in \A} \ub{\pi}(\a | \s) \ge 1$. The altered dual formulation for the value learning problem \eqref{eq:vl} is then given by
\begin{subequations}[ADB]
    \label{eq:adb}[!]
    \begin{align}
        &\mspace{-6mu}\begin{optim}
            \max_{\substack{d(\s) \\ p(\s, \a)}}{\expect[r]_{\subalign{(\s, \a) &\sim p(\cdot, \cdot) \\ \s' &\sim \tau(\cdot | \s, \a)}\mspace{-20mu}}[\big]{r(\s, \a, \s')}}
            \st{
                \sum_{\a \in \A} p(\s, \a) &= d(\s) &\;& \forall \s \\[-6pt]
                d(\s) &= (1 - \gamma) \iota(\s) \\
                &\mspace{40mu}+ \gamma \quad \expect[lr]_{(\s', \a') \sim p(\cdot, \cdot)\;}[\big]{\tau(\s | \s', \a')} && \forall \s \\[-6pt]
                p(\s, \a) &\ge 0 && \forall \s \, \forall \a \\
                d(\s) \lb{\pi}(\a | \s) &\le p(\s, \a) \le d(\s) \ub{\pi}(\a | \s) && \forall \s \, \forall \a
            }
        \end{optim}\mspace{-16mu}
        \label{eq:adb_dual}\subtag[\textperiodcentered]{D}
        \intertext{and has the corresponding primal formulation}
        &\mspace{8mu}\begin{optim}[t]
            \min_{\substack{v(\s), q(\s, \a) \\ \smash[b]{\lb{r}(\s, \a), \ub{r}(\s, \a)}}}{\mspace{26mu}\expect[r]_{\;\s_0 \sim \iota(\cdot)}[\big]{(1 - \gamma) v(\s_0)}}
            \st{
                v(\s) &\ge q(\s, \a) &\;& \forall \s \, \forall \a \\
                \mspace{-16mu}q(\s, \a) &= \expect[blr]_{\vs{8pt}\qquad\s' \sim \tau(\cdot | \s, \a)}\bracks[\Big]{r(\s, \a, \s') + \lb{r}(\s, \a) - \ub{r}(\s, \a) \\[-4pt]
                &\mspace{128mu}- \sum_{\tilde{\a} \in \A} \lb{r}(\s, \tilde{\a}) \lb{\pi}(\tilde{\a} | \s) \\[-2pt]
                &\mspace{128mu}+ \smash[b]{\sum_{\tilde{\a} \in \A}} \ub{r}(\s, \tilde{\a}) \ub{\pi}(\tilde{\a} | \s) \\[-4pt]
                &\mspace{230mu}+ \gamma v(\s')} && \forall \s \, \forall \a \\
                \mspace{-16mu}\lb{r}(\s, \a) &\ge 0 \land \ub{r}(\s, \a) \ge 0 && \forall \s \, \forall \a
            }
        \end{optim}\mspace{-80mu}
        \label{eq:adb_prim}\subtag[\textperiodcentered]{P}
    \end{align}
\end{subequations}
The reward modifications in the primal are more complex in this case
\begin{equation}
    \mspace{-6mu}\begin{aligned}
        r_{\mathrm{ADB}}^{}(\s, \a, \s') = r(\s, \a, \s') &+ \lb{r}(\s, \a) - \sum_{\tilde{\a} \in \A} \lb{r}(\s, \tilde{\a}) \lb{\pi}(\tilde{\a} | \s) \\
        &- \ub{r}(\s, \a) + \sum_{\tilde{\a} \in \A} \ub{r}(\s, \tilde{\a}) \ub{\pi}(\tilde{\a} | \s), \\[-6pt]
    \end{aligned}
    \label{eq:adb_mod}
\end{equation}
and have learnable decision variables $\lb{r}(\s, \a)$ and $\ub{r}(\s, \a)$.
For feasible problems, following properties can be derived, extending \thmref{thm:vl_dual} and \lemref{lem:vl_prim}.
\begin{theorem}\label{thm:adb_dual}
    The policy that can be derived from the optimal dual variables is an optimal policy for the considered MDP with action density bounds.
\end{theorem}
\begin{lemma}\label{lem:adb_prim}
    The optimal primal variables satisfy the Bellman optimality equation and are thus equivalent to the optimal adjusted value functions $v_{\mathrm{VDB}}^*$ and $q_{\mathrm{VDB}}^*$ for the considered MDP with optimal modified reward $r_{\mathrm{VDB}}^*$.
\end{lemma}
Additionally, these optimal adjusted value functions and optimal reward modifications satisfy following properties.
\begin{theorem}\label{thm:adb_greedy}
    The optimal policy is greedy with respect to the optimal adjusted value functions $v_{\mathrm{ADB}}^*$ and $q_{\mathrm{ADB}}^*$.
\end{theorem}
\begin{proposition}\label{prop:adb_mod}
    The optimal modified reward is only altered by the tight bounds $\lb{\pi}(\a | \s) = \pi^\star(\a | \s)$ and $\pi^\star(\a | \s) = \ub{\pi}(\a | \s)$.
\end{proposition}
On the other hand, if there is no policy that can satisfy all additional constraints, the dual problem effectively becomes infeasible and the primal becomes unbounded, caused by the $\lb{r}(\s, \a)$ and $\ub{r}(\s, \a)$ of the infeasible constraints approaching infinity.

Similar derivations are possible for the policy evaluation problem \eqref{eq:pe}, leading to a Constrained GPI scheme that is comparable to the one for the visitation density bounds.

\subsection{Transition Constraints}
In control applications, one often wants to prevent abrupt changes in the applied controls and/or in the state. This can improve user comfort and lead to more smooth control. To enforce such constraints, typically a cost is associated with the variation of states and inputs, which can then be upper bounded.
The dual formulations \eqref{eq:vl_dual} and \eqref{eq:pe_dual} allow us to do a similar thing for reinforcement learning policies. We denote by $c(\s[0], \a[0], \s[1]): \S \times \A \times \S \rightarrow \mathbb{R}$ the cost of transitioning to state $\s[1]$ from state $\s[0]$ by taking action $\a[0]$. The expected transition cost of a certain state-action pair can then be written as $\bar{c}(\s, \a) = \expect_{\s' \sim \tau(\cdot | \s, \a)}[\big]{c(\s, \a, \s')}$. To ensure that this average transition cost remains upper bounded for states and actions that are visited by the policy, constraints of the form $p^\pi(\s, \a) \bar{c}(\s, \a) \le 0$ can then be added to the dual. Note that this could be easily extended to multiple such constraints with different cost functions and upper bounds, as was done for the constrained RL formulation \eqref{eq:crl}.
The altered value learning dual with average transition constraints then becomes
\begin{subequations}[ATC]
    \label{eq:atc}[!]
    \begin{align}
        &\begin{optim}
            \max_{\substack{d(\s) \\ p(\s, \a)}}{\expect[r]_{\subalign{(\s, \a) &\sim p(\cdot, \cdot) \\ \s' &\sim \tau(\cdot | \s, \a)}\mspace{-20mu}}[\big]{r(\s, \a, \s')}}
            \st{
                \sum_{\a \in \A} p(\s, \a) &= d(\s) &\;& \forall \s \\[-6pt]
                d(\s) &= (1 - \gamma) \iota(\s) \\
                &\mspace{40mu}+ \gamma \quad \expect[lr]_{(\s', \a') \sim p(\cdot, \cdot)\;}[\big]{\tau(\s | \s', \a')} && \forall \s \\[-6pt]
                p(\s, \a) &\ge 0 && \forall \s \, \forall \a \\
                0 &\ge p(\s, \a) \expect[lr]_{\qquad\s' \sim \tau(\cdot | \s, \a)}[\big]{c(\s, \a, \s')} && \forall \s \, \forall \a
            }
        \end{optim}\mspace{-20mu}
        \label{eq:atc_dual}\subtag[\textperiodcentered]{D}
        \intertext{with corresponding primal}
        &\begin{optim}[t]
            \min_{\substack{v(\s), q(\s, \a) \\ \smash[b]{\rc(\s, \a)}}}{\mspace{14mu}\expect[r]_{\;\s_0 \sim \iota(\cdot)}[\big]{(1 - \gamma) v(\s_0)}}
            \st{
                v(\s) &\ge q(\s, \a) &\;& \forall \s \, \forall \a \\
                q(\s, \a) &= \expect[blr]_{\qquad\s' \sim \tau(\cdot | \s, \a)}\bracks[\big]{r(\s, \a, \s') - \rc(\s, \a) c(\s, \a, \s') \\
                &\mspace{212mu}+ \gamma v(\s')} && \forall \s \, \forall \a \\
                \rc(\s, \a) &\ge 0 && \forall \s \, \forall \a
            }
        \end{optim}\mspace{-70mu}
        \label{eq:atc_prim}\subtag[\textperiodcentered]{P}
    \end{align}
\end{subequations}
The reward modifications in the primal have a slightly different form now, combining a learnable reward weight $\rc(\s, \a)$ with the given cost $c(\s, \a, \s')$ in what looks to be a mixture of the \eqref{eq:crl} and \eqref{eq:vdb_pe} formulations
\begin{equation}
    r_{\mathrm{ATC}}^{}(\s, \a, \s') = r(\s, \a, \s') - \rc(\s, \a) c(\s, \a, \s').
    \label{eq:atc_mod}
\end{equation}
For feasible problems, following properties hold, extending \thmref{thm:vl_dual} and \lemref{lem:vl_prim}.
\begin{theorem}\label{thm:tc_dual}
    The policy that can be derived from the optimal dual variables is an optimal policy for the considered MDP with average transition constraints.
\end{theorem}
\begin{lemma}\label{lem:tc_prim}
    The optimal primal variables satisfy the Bellman optimality equation and are thus equivalent to the optimal adjusted value functions $v_{\mathrm{ATC}}^*$ and $q_{\mathrm{ATC}}^*$ for the considered MDP with optimal modified reward $r_{\mathrm{ATC}}^*$.
\end{lemma}
Furthermore, following properties hold for these optimal adjusted value functions and optimal reward modifications.
\begin{theorem}\label{thm:tc_greedy}
    The optimal policy is greedy with respect to the optimal adjusted value functions $v_{\mathrm{ATC}}^*$ and $q_{\mathrm{ATC}}^*$.
\end{theorem}
\begin{proposition}\label{prop:tc_mod}
    The optimal modified reward is only altered by nonvisited state-action pairs or tight bounds $\bar{c}(\s, \a) = 0$.
\end{proposition}
On the other hand, if there is no policy that can satisfy all additional constraints, the dual problem effectively becomes infeasible and the primal becomes unbounded, caused by the $\rc(\s, \a)$ of the infeasible constraints approaching infinity.

Once again, a similar derivation is also possible for the policy evaluation problem \eqref{eq:pe}. Such a formulation has the added benefit that \emph{action} transition constraints can be enforced, as shown in \appref{app:tc}[app]. In this appendix an alternative \emph{immediate} transition constraint of the form $p^\pi(\s, \a) \tau(\s' | \s, \a) c(\s, \a, \s') \le 0$ is also examined, which is much stricter than the average transition constraints considered here.

\begin{table*}[t]
    \centering
	\caption{Overview of the altered primal problems. Novel constraint types are indicated with an asterisk ($*$).}
	\label{tab:alt}
	\begin{tabular}{@{\,} l l l l @{\,}}
		\toprule
		\textbf{Problem} & \textbf{Extra Decision Variables} & \textbf{Reward Modifications} & \textbf{Objective Modifications} \\
		\midrule
        \makecell[ml]{Policy Distillation \& \\ Maximum Entropy \eqref{eq:er} \cite{Schulman2018, Haarnoja2018}} & / & ${}-\alpha\log\frac{\pi(\a | \s)}{\pi_{\textsc{t}}(\a | \s)}$ & \;/ \\[2pt]
		Constrained RL \eqref{eq:crl} \cite{Altman1999} & $w_k \ge 0$ & ${}+\sum_{k=0}^K w_k r_k(\s, \a, \s')$ & ${}-\sum_{k=0}^K w_k V_k$ \\[2pt]
        Visitation Density Bounds \eqref{eq:vdb_pe} & $\lb{r}^{}(\s, \a) \ge 0$ & ${}+\lb{r}^{}(\s, \a)$ & $\textstyle {}-\sum_{\s \in \S} \sum_{\a \in \A} \lb{r}^{}(\s, \a) \lb{p}^{}(\s, \a)\,$ \\
        \cite{Chen2019, Qin2021} & $\ub{r}^{}(\s, \a) \ge 0$ & ${}- \ub{r}^{}(\s, \a)$ & $\textstyle {}+\sum_{\s \in \S} \sum_{\a \in \A} \ub{r}^{}(\s, \a) \ub{p}^{}(\s, \a)$ \\[2pt]
        Action Density Bounds \eqref{eq:adb} $*$ & $\lb{r}^{}(\s, \a) \ge 0$ & $\textstyle {}+\lb{r}^{}(\s, \a) - \sum_{\tilde{a} \in \A} \lb{r}^{}(\s, \tilde{\a}) \lb{\pi}^{}(\tilde{\a} | \s)$ & \;/ \\
        & $\ub{r}^{}(\s, \a) \ge 0$ & $\textstyle {}-\ub{r}^{}(\s, \a) + \sum_{\tilde{a} \in \A} \ub{r}^{}(\s, \tilde{\a}) \ub{\pi}^{}(\tilde{\a} | \s)$ \\[2pt]
		Transition Constraints \eqref{eq:atc} $*$ & $\rc^{}(\s, \a) \ge 0$ & ${}-\rc^{}(\s, \a) c(\s, \a, \s')$ & \;/ \\
		\bottomrule
	\end{tabular}
\end{table*}

\subsection{Overview and Discussion}
All altered optimization problems discussed in this section have in common that the effect of extra constraints or regularization terms in the dual leads to reward modifications \eqref{eq:er_mod}-\eqref{eq:atc_mod} and sometimes objective modifications in the primal. \tabref{tab:alt} gives an overview of the effect on the primal for each of the discussed extensions.

Except for the entropy regularized case, the reward modifications in the primal are automatically learned, as they depend on extra decision variables (Lagrange multipliers) associated with each of the introduced constraints in the dual.
These reward modifications also make sense intuitively: state-action pairs that need to be visited more frequently by the policy in order to satisfy the constraints get a (positive) reward bonus; whereas states that need to be visited less frequently get a (negative) reward penalty.

Propositions \ref{prop:crl_mod}, \ref{prop:vdb_mod}, \ref{prop:adb_mod}, \ref{prop:tc_mod} additionally show that the \dq{smallest possible} modifications are typically optimal. More precisely, the optimal reward modifications are zero for states and actions where the corresponding inequalities are not tight. In theory, this means that the optimal policy for the problem without these constraints would also satisfy the strict inequalities. In practice, the constraints are still useful as they can guide the policy towards the desirable regions in state-action space throughout learning.

If there is no policy that can satisfy all additional constraints, the dual problems effectively become infeasible, and the primal problems become unbounded, caused by the reward modification terms associated with the infeasible constraints approaching infinity.
\begin{remark}
    In summary, the reward is modified in such a way that constraint-violating policies can no longer be optimal or greedy with respect to the adjusted value functions, whereas policies that are greedy with respect to the optimal adjusted value functions are optimal solutions to the constrained RL problems (Theorems \ref{thm:crl_greedy}, \ref{thm:adb_greedy}, \ref{thm:tc_greedy}).
\end{remark}

Although the modified policy evaluation LPs can thus become infeasible for certain policies, the optimal feasible policy can still be found when it is used as the inner optimization problem in a GPI scheme. The reward modifications (and hence also the adjusted value functions and objective) never reach $\pm\infty$, as the policy learns to avoid the unwanted regions in state-action space and to move towards the wanted regions in state-action space.

To conclude, the various problems summarized in \tabref{tab:alt} provide the designer with a versatile toolbox of policy constraints, acting on different scopes in the trajectory. In particular, there are constraints acting on individual visited states and actions \eqref{eq:vdb_pe}, \eqref{eq:adb}, on transitions between consecutively visited states \eqref{eq:atc}, and on whole trajectories of visited states \eqref{eq:crl}.
While presented as individual optimization problems, the different constraints and corresponding reward modifications can be straightforwardly combined, as is also shown in the experiments (\secref{sec:experiments}).

\section{Practical Implementations}\label{sec:practical}
The previously introduced optimization problems cannot be solved directly, as the distributions $\iota$ and $\tau$ and the reward function $r$ are unknown in the general model-free reinforcement learning setup. Using parameterized models and experience samples from the replay buffer, they can be approximately solved though, leading to practical value-based and actor-critic algorithms, referred to as \sq{Dual Constrained Reinforcement Learning} (DualCRL) methods.
The introduced parameterizations are denoted as $q(\s, \a; \p_q)$ for the state-action value function (or critic), $\pi(\a | \s; \p_\pi)$ for the policy (or actor), $d(\s; \p_d)$ for the state visitation density and $\tilde{r}(\s, \a, \s'; \p_r)$ for the reward modifications. For small and discrete state and action spaces, tabular models can be used, whereas for larger dimensions neural networks are more suitable. Generic pseudocode for training each of the models required for solving the various constrained RL tasks is given in \algref{alg:generic}.
\begin{algorithm}
    \caption{DualCRL with Learnable Reward Modifications}\label{alg:generic}
    \begin{algorithmic}
        \State \textbf{Initialize} $\p_q, \p_\pi, \p_d, \p_r$ and collect experience in $\buf$
        \For{each gradient step}
            \State Sample batch $\mathcal{E} = \set{\exptup_b}_{b \in \bracks{B}}$ from $\buf$
            \State $\p_q \gets \p_q - \eta_q \nabla_{\p_q} \widehat{L}_q(\p_q; \mathcal{E})$ \Comment{Subsections \ref{ssec:vb}, \ref{ssec:ac}}
            \State $\p_\pi \gets \p_\pi - \eta_\pi \nabla_{\p_\pi} \widehat{L}_\pi(\p_\pi; \mathcal{E})$ \Comment{\ssecref{ssec:ac}}
            \State $\p_d \gets \p_d - \eta_d \nabla_{\p_d} \widehat{L}_d(\p_d; \mathcal{E})$ \Comment{\ssecref{ssec:vde}}
            \State $\p_r \gets \p_r - \eta_r \nabla_{\p_r} \widehat{L}_r(\p_r; \mathcal{E})$ \Comment{\ssecref{ssec:arm}}
        \EndFor
    \end{algorithmic}
\end{algorithm}

The stochastic estimate of the true gradient $\nabla_{\p} L(\p)$ using a batch of experience samples $\mathcal{E}$ is denoted by $\nabla_{\p} \widehat{L}(\p; \mathcal{E})$.
Following state-of-the-art deep RL algorithms, additional independent parameterizations and target networks (denoted with a prime, e.g. $\p_q'$) can be introduced as well to further improve stability and performance~\cite{VanHasselt2016, Fujimoto2018}.

Tuning the various learning rates $\eta$ (and strides) for each model is an important, yet difficult and problem dependent, task. Generally, the visitation density models should be updated on a faster timescale than the actor and reward modification models, such that the used density estimates properly reflect the learned policy's behaviour. Otherwise, a high bias in the visitation density estimates could lead to erroneous updates of the reward modifications and eventually to suboptimal policies.
For the reward modifications there is a certain trade-off. If $\eta_r$ is too low (or the stride too high), policies are only slowly forced to comply with constraints, leading to many constraint violating iterations. On the other hand, if $\eta_r$ is too high (or the stride too low), reward modifications might oscillate or overshoot, before being properly reflected in the value estimates, destabilizing the learning process.

\subsection{Value-Based}\label{ssec:vb}
The Value Learning LP \eqref{eq:vl} and its extensions discussed in the previous section can be turned into value-based algorithms, resembling $Q$-learning.
Because the optimal $q^*$ should satisfy the Bellman optimality equation, it can be learned by minimizing the squared Bellman residual (or temporal difference error) $e_t = q(\s[0], \a[0]) - \expect_{\s[1] \sim \tau(\cdot | \s[0], \a[0])}{r(\s[0], \a[0], \s[1]) + \tilde{r}(\s[0], \a[0], \s[1]) + \gamma \max_{\a'} q(\s[1], \a')}$ for all $\s[0]$ and $\a[0]$. This can be approximated using samples from the replay buffer, leading to the critic loss $L_q(\p_q) = \expect_{\exptup \sim \buf}*{\parens*{q(\s[0], \a[0]; \p_q) - y_t}^2}$ with $y_t = r_t + \tilde{r}(\s[0], \a[0], \s[1]; \p_r) + \gamma \max_{\a'} q(\s[1], \a'; \p_q')$. There is no explicit policy model in this algorithm, but the policy can be implicitly derived as $\pi(\a | \s) = \ind_{\A^*(\s)}(\a) / \abs{\A^*(\s)}$ with $\A^*(\s) = \argmax_{\a} q(\s, \a; \p_q)$. The optimal state value function can similarly be obtained as $v^*(\s) = \max_{\a} q(\s, \a; \p_q^*)$.

Note that the value-based algorithm always learns a deterministic policy, which is generally not applicable to constrained RL problems, for which the optimal policy might be stochastic. Hence, if it is unknown beforehand whether the optimal policy can be deterministic for a certain problem, the actor-critic algorithm is typically the better choice, as it is capable of learning stochastic policies.

\subsection{Actor-Critic}\label{ssec:ac}
The GPI scheme, based on the outer Policy Optimization problem \eqref{eq:po} and inner Policy Evaluation LP \eqref{eq:pe}, and its alterations can be turned into actor-critic algorithms.
The optimal $q^*$ should satisfy the Bellman equation, and can thus be learned by minimizing the squared Bellman residual (or temporal difference error), leading to the critic loss $L_q(\p_q) = \expect_{\exptup \sim \buf}*{\parens*{q(\s[0], \a[0]; \p_q) - y_t}^2}$ with $y_t = r_t + \tilde{r}(\s[0], \a[0], \s[1]; \p_r) + \gamma \expect_{\a' \sim \pi(\cdot | \s[1]; \p_\pi)}{q(\s[1], \a';\p_q')}$. During Policy Evaluation steps, this loss is minimized, which moves $q$ towards the value function of the current policy $\pi$. The state value function can be approximated as $v(\s) = \expect_{\a \sim \pi(\cdot | \s; \p_\pi)}{q(\s, \a; \p_q)}$.

The optimal $\pi^*$ should maximize the average reward, and can thus be learned by minimizing the actor loss (derived from the Lagrangian objective) $L_\pi(\p_\pi) = -\expect_{\s[0] \sim \buf, \a \sim \pi(\cdot | \s[0]; \p_\pi)}{q(\s[0], \a; \p_q)}$.
During Policy Improvement steps, this loss is minimized, which moves $\pi$ towards the greedy policy with respect to the current value function $q$.
Note that for certain extensions, this objective contains some additional terms (originating from the Lagrangian objective). For example for the entropy regularized case the extra term $\alpha \log \frac{\pi(\a | \s[0]; \p_\pi)}{\pi_{\textsc{t}}(\a | \s[0])}$ appears.
Depending on the chosen parametric distribution for $\pi$, calculating the gradient of $L_\pi$ typically involves using the reparametrization trick or implicit reparametrization gradients~\cite{Kingma2015, Figurnov2018}.

\subsection{Policy-Based}
The actor loss expression $L_\pi$ is consistent with the policy gradient theorem~\cite{Sutton2018}. The same Policy Optimization problem \eqref{eq:po} could thus also be used to derive \emph{policy-based} methods, in which a Monte Carlo estimate is used for determining the value $q^\pi(\s, \a)$ instead of a separate critic model. This setup is however not further investigated in this work.

\subsection{Visitation Density Estimation}\label{ssec:vde}
Samples from the visitation densities can be easily obtained from the replay buffer, as is done in the evaluation of the actor and critic losses. However, for certain updates of the reward modification models, actual density probabilities need to be estimated.
Having access to accurate visitation density estimates is crucial for correctly learning such reward modifications and ultimately for obtaining optimal policies.

For discrete, finite-dimensional state spaces, (discounted) counting of the visited states in tabular density models is possible. For larger (or continuous) state spaces, other tractable density estimation techniques --- such as (Gaussian) mixture models or normalizing flows~\cite{Rezende2015} --- can be leveraged to train the parameterized visitation density model $d(\s; \p_d)$ using samples from the replay buffer.
Such density models are trained by maximizing the average log likelihood, leading to the density loss $L_d(\p_d) = -\expect_{\s \sim \buf}{\log d(\s; \p_d)}$. Alternatively, for smaller dimensional state spaces, non-parametric density estimation methods, such as Kernel Density Estimation (KDE)~\cite{Parzen1962}, could be applied on the most recently visited states in the replay buffer.

The state-action visitation density can either be estimated in a similar way or calculated as $p(\s, \a; \p_p) = d(\s; \p_d) \pi(\a | \s; \p_\pi)$.

\subsection{Learnable Reward Modifications}\label{ssec:arm}
The reward modification models $\tilde{r}$ (see also third column of \tabref{tab:alt}) are updated by minimizing the reward loss $L_r$, which is derived from the Lagrangian objective and hence depends on the specific constraints that are enforced. For tabular parameterizations, a projected gradient descent step is applied to ensure that the reward modifications (or weights $w_k$ for the constrained RL case) remain nonnegative. For neural parameterizations, this property can be enforced on the network architecture itself, for example by using a softplus activation function ($\log[1 + \exp(x)]$) in the output layer.

For the value constraints of \eqref{eq:crl}, we have $L_r(\p_r) = \sum_{k=0}^K w_k \expect_{(\s, \a, \s') \sim \buf}{r_k(\s, \a, \s')-(1 - \gamma) V_k}$ with $\p_r = \mat{w_1; \dots; w_K}$. The visitation density bounds of \eqref{eq:vdb_pe} lead to $L_r(\p_r) = \expect_{(\s, \a) \sim \buf}[\Big]{\lb{r}(\s, \a; \p_{\lb{r}}) \parens[\Big]{1 - \frac{\lb{p}(\s, \a)}{p(\s, \a; \p_p)}} + \ub{r}(\s, \a; \p_{\ub{r}}) \parens[\Big]{\frac{\ub{p}(\s, \a)}{p(\s, \a; \p_p)} - 1}}$ with $\p_r = \mat{\p_{\lb{r}}; \p_{\ub{r}}}$. This is very similar to the reward loss for action density bounds \eqref{eq:adb}, given by $L_r(\p_r) = \expect_{(\s, \a) \sim \buf}[\Big]{\lb{r}(\s, \a; \p_{\lb{r}}) \parens[\Big]{1 - \frac{d(\s; \p_d) \lb{\pi}(\a | \s)}{p(\s, \a; \p_p)}} + \ub{r}(\s, \a; \p_{\ub{r}}) \parens[\Big]{\frac{d(\s; \p_d) \ub{\pi}(\a | \s)}{p(\s, \a; \p_p)} - 1}}$. Finally, for the average transition costs of \eqref{eq:atc}, we obtain $L_r(\p_r) = -\expect_{(\s, \a, \s') \sim \buf}{\rc(\s, \a; \p_r) c(\s, \a, \s')}$.

Remark that for the specific case of individual policy constraints, the derived algorithm is practically equivalent to some existing state-of-the-art actor-critic methods, such as SAC~\cite{Haarnoja2018} in the entropy regularized setting, RCPO~\cite{Tessler2018} in the CMDP setting and DCRL~\cite{Qin2021} in the visitation density constrained setting.
Nevertheless, our unifying framework allows to seamlessly combine multiple kinds of constraints, by simply adding the reward modifications for each type of constraint (and likewise for the corresponding losses).

\section{Experiments}\label{sec:experiments}
The presented \texttt{DualCRL} algorithm is applied to two environments to demonstrate its utility and the effect of different types of constraints discussed in this work.
Both the value-based and actor-critic implementations are first evaluated on the \texttt{CliffWalking} environment with discrete state and action spaces, using tabular models. The actor-critic implementation is afterwards used in the \texttt{Pendulum} environment with continuous state and action spaces, using neural networks.
The choice for these relatively simple environments is twofold. First of all, the small dimensions of their state and action spaces, allow to visualize the effect of the various constraints and reward modifications on the learned policy's behaviour. Secondly, their simple reward structure leads to well-known controller behaviour, which can be easily extended using the various policy constraints introduced here to obtain more complex behaviour.
We specifically focus on a combination of different types of policy constraints in these experiments, as individual types of constraints have been extensively studied already in the entropy regularized and constrained RL domains~\cite{Haarnoja2018, Tessler2018, Qin2021} --- including also more challenging environments with higher dimensions.

An implementation of the \texttt{DualCRL} method in PyTorch is made available \blindfootnotehref{https://github.com/dcbr/dualcrl}{here}.
Hyperparameter values were empirically determined and are summarized in \appref{app:hyper}[app].

\begin{figure}
    \centerline{\includegraphics[width=\linewidth]{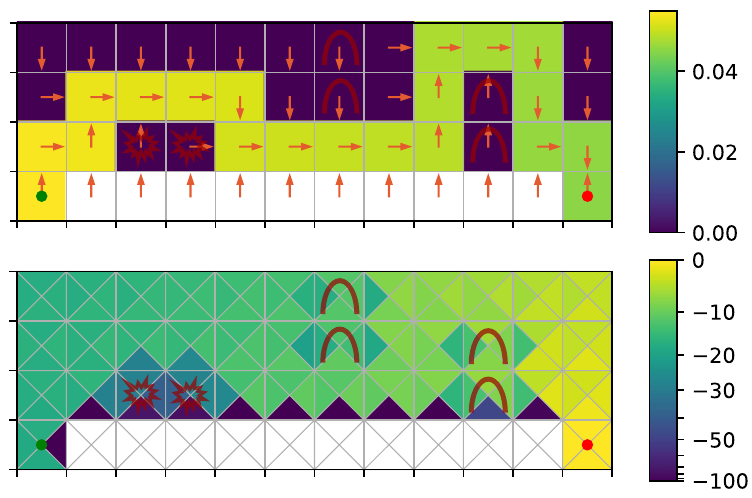}}
    \caption{CliffWalking with state visitation density bounds and transition constraints. The red labels \usym{1F4A5} and $\cap$ indicate the unstable squares and ridges respectively. Top: resulting policy and estimated visitation density after \num{20000} timesteps. Bottom: learned adjusted value function after \num{50000} timesteps.}
	\label{fig:cliff_db_tc}
\end{figure}

\subsection{Cliff Walking}
In the \texttt{CliffWalking} environment~\cite{Sutton2018}, the goal is to move from the lower left corner of a rectangular grid to the lower right corner, while avoiding the dangerous squares on the bottom row (the cliff). The state space $\S = \bracks{S}$ enumerates all squares of the grid and the action space $\A = \set{\uparrow, \rightarrow, \downarrow, \leftarrow}$ provides the possible movements to neighbouring squares. The reward is $-1$ for each state, except for the cliff states where it is $-100$.

As a first extension to this problem, we examine the effect of state-visitation density bounds and transition constraints using the value-based method. Suppose we know that two squares near the cliff edge are unstable. Then we can avoid these by imposing $d(\s) \le \epsilon$ with $\epsilon$ a small, strictly positive number. Additionally, if there are two ridges making horizontal movements more difficult, we can assign a strictly positive transition cost $c(\s, \a, \s') > 0$ for all states $\s$ and $\s'$ on the ridge and horizontal movements $\a \in \set{\rightarrow, \leftarrow}$. After some initial exploration and training, \figref{fig:cliff_db_tc} shows the learned adjusted value function, the resulting policy and estimated visitation density.
It is immediately clear that the policy has learned to avoid the two unstable squares and ridges by going around them. This behaviour is also clearly reflected in the learned adjusted value function, which assigns more negative values to the state-action pairs that need to be avoided. 
Remark that while the imposed visitation bounds lead to a reward penalty for the unwanted \emph{states}, this also affects the value for \emph{state-action pairs} of neighbouring states moving towards these (unstable) states. On the other hand, the imposed transition constraints only penalize the costly state-action pairs associated with horizontal movements across the ridge, while leaving the vertical movements unaffected (for this particular setup, a similar result could also be obtained using state-action visitation density bounds).

\begin{figure}
    \centerline{\includegraphics[width=\linewidth]{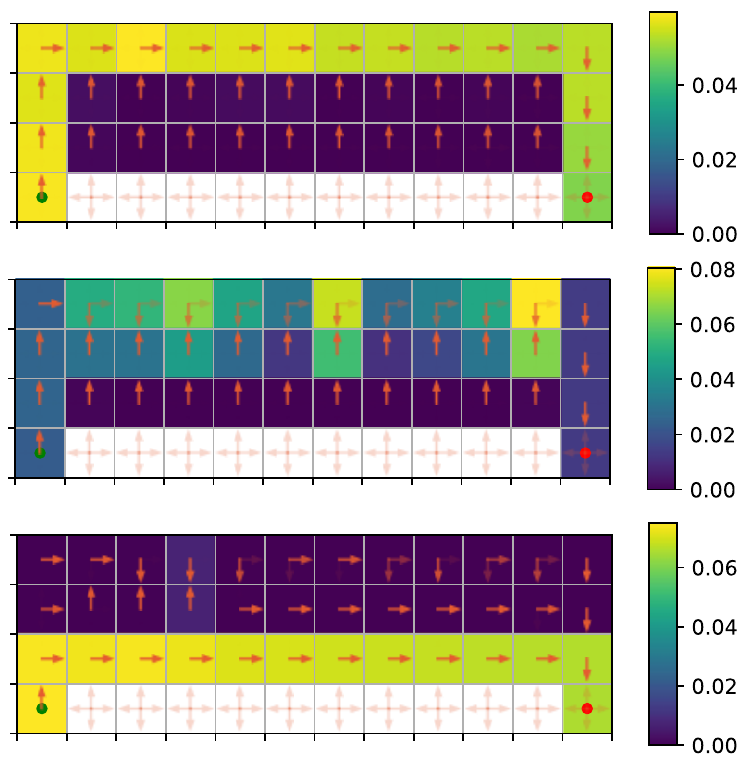}}
    \caption{CliffWalking with entropy regularization and action density bounds. The learned policy and estimated visitation density are shown after \num{4000}, \num{8000} and \num{50000} timesteps (from top to bottom).}
	\label{fig:cliff_er_ab}
\end{figure}

In a second experiment, the entropy regularization and action density bounds are combined. As the teacher policy, we choose a safe policy, moving as far away as possible from the cliff. The action density bounds are chosen to \dq{oppose} the teacher policy, by imposing $\pi(\downarrow | \s) \ge 0.5$ for all states in the top row (except the first and last column). The temperature parameter $\alpha$ is exponentially decayed throughout training. \figref{fig:cliff_er_ab} summarizes the results of this experiment.
At the start of training the learned policy clearly mimics the teacher policy in all states. As training goes on, and the temperature $\alpha$ is lowered, the policy gains more freedom to deviate from the teacher's behaviour, while the importance of the action density bounds also increases through updated reward modifiers. These bounds push the policy away from the top row, allowing it to further explore the state-action space and find more efficient paths towards the goal. Finally, at the end of training, the policy has found the more efficient path towards the goal, which remains closer to the cliff.
This example illustrates how the learned policy behaviour can be influenced by the temperature parameter $\alpha$, controlling the relative importance of the (conflicting) entropy regularization and average reward maximization objective with policy constraints.

\subsection{Pendulum}
In the \texttt{Pendulum} environment, the goal is to swing up and stabilize a pendulum around the upward position. The continuous state consists of the angle $\theta$ and angular velocity $\dot{\theta}$ of the pendulum and the agent receives this information through observations $\mat{x, y, \dot{\theta}}$ with $x=\cos(\theta)$ and $y=\sin(\theta)$. The continuous action is the torque applied to the pendulum. The reward penalizes deviations from the upward position and, to a lesser degree, high angular velocity or applied torque.

The experiments conducted in this environment use the actor-critic algorithm with entropy regularization to encourage exploration (similar to SAC~\cite{Haarnoja2019}) with an exponentially decreasing temperature. The state visitation density is estimated using Gaussian Kernel Density Estimation (KDE)~\cite{Parzen1962} using the last $5000$ visited states of the behavioural policy.

\begin{figure}
    \centerline{\includegraphics[width=\linewidth]{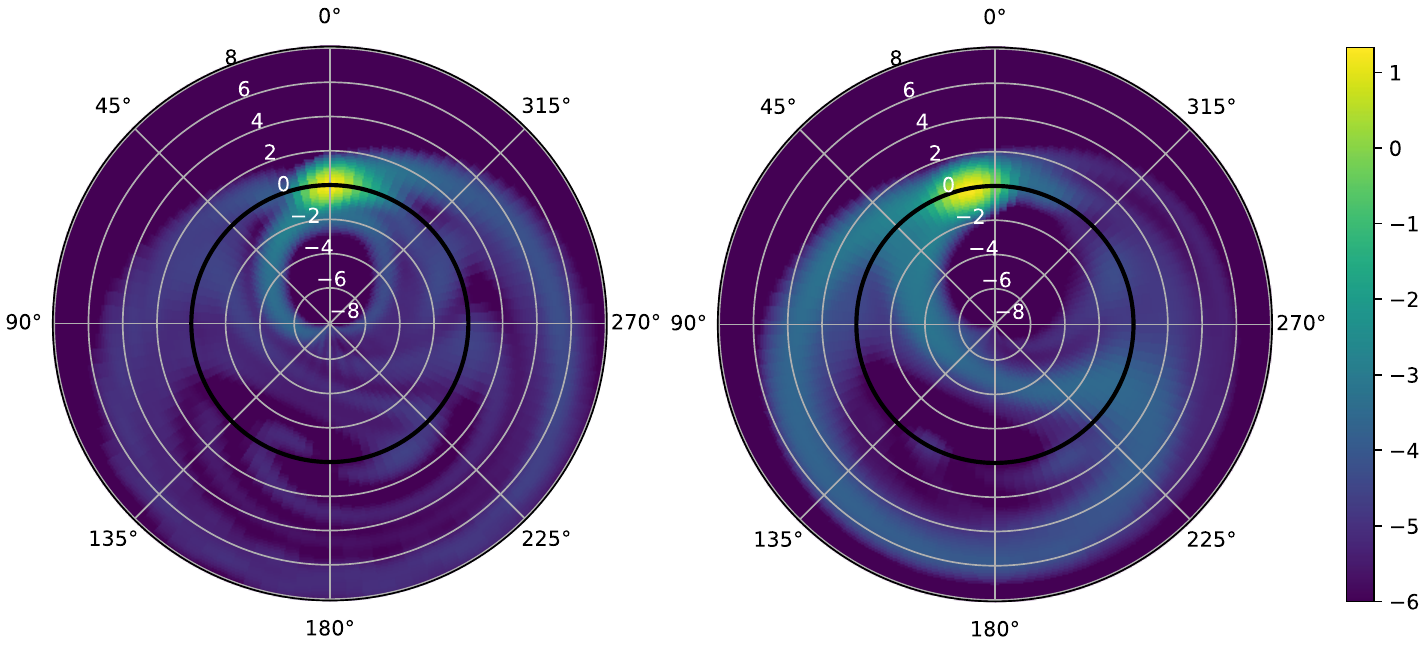}}
    \caption{Visualization of the estimated state visitation density on a (natural) logarithmic scale for the Pendulum experiments after \num{50000} timesteps. The angle $\theta$ and angular velocity $\dot{\theta}$ are shown on the angular and radial axis respectively. Left: Entropy regularization. Right: Entropy regularization, state-visitation density bounds and value constraints.}
	\label{fig:pendulum_state}
\end{figure}

The problem is extended by imposing state-visitation density bounds of the form $d(\s) \le \epsilon$ for all states in the upper right quadrant, and adding a value constraint with additional reward $r_\mathrm{V}^{}(\s, \a, \s') = -\dot{\theta}^2$ (for all states). The aim of these additional constraints is to avoid swing-ups along the right side and avoid high angular velocities. \figref{fig:pendulum_state} visualizes the estimated visitation density of the learned policy and compares it with a baseline policy trained without such constraints. The shown visitation densities are estimated using KDE on states collected from $10$ independently trained models in $4$ evaluation episodes with $200$ timesteps each.
From those visitation plots, we can clearly see the constrained policy chooses the left side for swingups and stabilization, while avoiding the highest angular velocities. The right side is only used for gaining momentum (lower right quadrant) and stabilizing around the upward position (upper right quadrant). Note that some visitations of the upper right quadrant are still visible, as random initializations can still put the pendulum in such a state.

\section{Related Work}\label{sec:rel}
Linear Programming (LP) formulations of Markov Decision Processes (MDP) or Reinforcement Learning (RL) have been analyzed by \AuthorCite{Puterman1994} and \AuthorCite{Altman1999}.
When the environment dynamics and reward function are known, the value learning LP can be used to approximately solve MDPs~\cite{deFarias2003, Cogill2015}. However, this is not easy in practice, when dealing with unknown dynamics and large state and action spaces. In this work, the LPs form the basis of our framework to derive primal-dual formulations of various constrained RL tasks, from which practical algorithms can then be deduced.

Various primal-dual algorithms for solving different RL tasks have already been proposed. Some of those analyze the \dq{pure} setting, without additional policy constraints~\cite{Yang2009, Wang2016, Dai2018}, whereas others investigate one specific type of constraint. For example, there is a significant amount of research related to value constraints (CMDPs)~\cite{Bhatnagar2012, Achiam2017, Tessler2018, Paternain2019, Stooke2020, Ding2020, Chen2021, Paternain2023, Li2023}. Other work covers visitation density constraints~\cite{Chen2019, Qin2021} and entropy regularization~\cite{Neu2017}.
The intent of this paper is to provide a unifying framework for these seemingly different subdomains and derive a practical algorithm that can handle and combine different kinds of policy constraints.
Moreover, by separately treating the value iteration and generalized policy iteration schemes, the impact of additional policy-dependent regularization terms or constraints can be straightforwardly analyzed in our framework. This is in contrast to other existing derivations for practical primal-dual algorithms, which introduce the policy in the dual or Lagrangian representation of \eqref{eq:vl}. As this results in a bilinear dual problem, without clear primal representation, such analyses are no longer possible.

\AuthorCite{Nachum2020} also used primal and dual formulations of the Value Learning LP to derive \emph{off-policy} and \emph{offline} RL algorithms, such as DualDICE~\cite{Nachum2019} and AlgaeDICE~\cite{Nachum2019a}.
By replacing the dual objective with an $f$-divergence, they derived suitable offline formulations. Instead of replacing the objective, we add a KL-divergence regularization term to the dual objective, which then leads to the maximum entropy and policy distillation subdomains.
For the theoretical derivations in this paper, we assume on-policy experience samples are available. An extension of our proposed framework to offline RL would be an interesting future path of research, as \AuthorCite{Polosky2022} have shown that LP formulations for offline policy evaluation can be combined with value constraints.

Our work can be related to \emph{convex} RL literature~\cite{Zhang2020a, Zahavy2021, Geist2022}, where the dual reward maximization objective is generalized to any concave utility function of the policy visitation.
All of the discussed optimization problems in this work also fit in this convex RL framework. For example, our entropy regularization formulations can be seen as a hybrid combination of the standard RL objective with the imitation learning and pure exploration objectives. This is similar to the objective of \AuthorCite{Zhang2021} with a convex regularizer for finding optimal caution-sensitive policies.
We specifically focus on a subset of convex RL problems, for which primal and dual formulations can be easily derived and modified, with a focus on constraining the policy. For these problems, we show an intrinsic relationship between dual constraints and primal reward modifications. Our work could be further extended to other convex RL problems, or leverage some of the proposed game-theoretic solution strategies~\cite{Miryoosefi2019, Zahavy2021, Geist2022}.

Finally, there are various approaches that try to learn or finetune the reward in the field of \emph{inverse} and \emph{automated} RL. This is however not the focus of this work, where the learned reward modifications in the primal have a particular structure that is directly related to certain policy constraints in the dual.

\section{Conclusion}
In this paper we outlined a generic primal-dual framework for value-based and actor-critic reinforcement learning methods. The dual formulations allowed us to easily plug in and impose additional constraints on the learned policy, unifying several reinforcement learning subdomains. Furthermore, the derivations revealed that these dual policy constraints correspond to learnable reward modifications in the primal. A practical \texttt{DualCRL} method was derived and investigated in more detail with different combinations of constraints on two interpretable environments.
Ultimately the presented algorithm aims to lower the burden on reinforcement learning designers, allowing them to naturally divide the problem at hand in a primary objective, through the reward function, and additional policy constraints. With a carefully selected reward and set of constraints, the design remains simple and easy to understand, while retaining the ability to learn complex behaviour.

For future work, it would be interesting to apply the presented methods in more complex environments with higher dimensional state and action spaces. As this was already done for individual constraint types \cite{Haarnoja2018, Tessler2018, Qin2021}, we expect no major difficulties when scaling up the \texttt{DualCRL} algorithm using similar techniques. Furthermore, the effect of the learning rates for the various models could be further investigated, as they can have a significant impact on training speed and performance, especially in more challenging setups. Finally, a more thorough analysis of the introduced approximation errors for the practical algorithm (due to the introduction of parameterized models) would be useful, similar to the work of \AuthorCite[Paternain2019]{Paternain2023} for the constrained RL case. Alternatively, function approximators that are convex in their parameters, such as kernel methods (with feature map representations $w^\top \phi(x)$) can be used without introducing a duality gap.

\bibliographystyle{IEEEtranN}
\bibliography{IEEEabrv,refs}

\clearpage
\appendices
\setlabelsuffix{app}

\section{}\label{app:theory}
\def\mathscale{0.84}
In this appendix, the derivations of all discussed primal and dual optimization problems are given, together with optimality conditions. Our derivations align with the standard practices in the convex optimization field, thoroughly described by \AuthorCite{Boyd2004} and \AuthorCite{Bertsekas2009}. Furthermore, it is assumed the initial state distribution $\iota(\s_0)$ is strictly nonzero for all states, as commonly done in MDP literature. This means every state gets visited, i.e. $\S^\pi = \S$ for any policy $\pi$. If this is not the case, sufficient exploration is needed to ensure that every state can be visited.
Remark that optimal solutions to the given optimization problems are denoted by a star $x^\star$ to distinguish them from optimal solutions to the considered reinforcement learning task, which are denoted by an asterisk $x^*$.

\subsection{Value Learning}\label{app:vl}
Starting from the primal formulation of the Value Learning problem \eqref{eq:vl_prim}
\begin{equation*}\scalemath{
    \begin{optim}
        \min_{\substack{v(\s) \\ \smash[b]{q(\s, \a)}}}{\expect[r]_{\;\s_0 \sim \iota(\cdot)}[\big]{(1 - \gamma) v(\s_0)}}
        \st{
            v(\s) &\ge q(\s, \a) &\;& \forall \s \, \forall \a \\
            q(\s, \a) &= \expect[lr]_{\qquad\s' \sim \tau(\cdot | \s, \a)}[\big]{r(\s, \a, \s') + \gamma v(\s')} && \forall \s \, \forall \a
        }
    \end{optim}
}\end{equation*}
the Lagrangian can be obtained as
\begin{equation*}\scalemath{\begin{aligned}
    \MoveEqLeft[2] \mathcal{L}_{\mathrm{VL}}(v(\s), q(\s, \a), \mu(\s, \a), \lambda(\s, \a)) \\
    = &\sum_{\s \in \S} \iota(\s) (1 - \gamma) v(\s) - \sum_{\s \in \S} \sum_{\a \in \A} \mu(\s, \a) \bracks*{v(\s) - q(\s, \a)} \\
    &-\sum_{\s \in \S} \sum_{\a \in \A} \lambda(\s, \a) \bracks[\Bigg]{q(\s, \a) - \sum_{\s' \in \S} \tau(\s' | \s, \a) \parens[\big]{r(\s, \a, \s') + \gamma v(\s')}} \\
    = &\sum_{\s \in \S}\sum_{\a \in \A} \lambda(\s, \a) \sum_{\s' \in \S} \tau(\s' | \s, \a) r(\s, \a, \s') \\
    &-\sum_{\s \in \S} v(\s) \bracks[\Bigg]{\sum_{\a \in \A} \mu(\s, \a) - (1 - \gamma)\iota(\s) \\[-16pt]
    &\mspace{182mu} - \gamma \sum_{\s' \in \S} \sum_{\a' \in \A} \lambda(\s', \a') \tau(\s | \s', \a')} \\
    &- \sum_{\s \in \S} \sum_{\a \in \A} q(\s, \a) \bracks[\big]{\lambda(\s, \a) - \mu(\s, \a)}.
\end{aligned}}\end{equation*}
The optimality (KKT) conditions are given by
\begin{itemize}[\itemindent=-7pt\labelsep=4pt]
    \setdisplayskips{2pt}{2pt}
    \item Stationarity of respectively $q$ and $v$, introducing auxiliary $d$
    \begin{equation*}\scalemath{\begin{alignedat}{2}
        p^\star(\s, \a) &= \lambda^\star(\s, \a) = \mu^\star(\s, \a) &\;& \forall \s \, \forall \a \\
        d^\star(\s) &= \sum_{\a \in \A} p^\star(\s, \a) \\
        &= (1 - \gamma) \iota(\s) + \gamma \sum_{\s' \in \S} \sum_{\a' \in \A} p^\star(\s', \a') \tau(\s | \s', \a') && \forall \s
    \end{alignedat}}\end{equation*}
    \item Primal feasibility
    \begin{equation*}\scalemath{\begin{alignedat}{2}
        v^\star(\s) &\ge q^\star(\s, \a) = \sum_{\s' \in \S} \tau(\s' | \s, \a) \parens[\big]{r(\s, \a, \s') + \gamma v^\star(\s')} &\;& \forall \s \, \forall \a
    \end{alignedat}}\end{equation*}
    \item Dual feasibility \qquad $\scalemath{p^\star(\s, \a) \ge 0 \; \forall \s \, \forall \a}$
    \item Complementary slackness
    \begin{equation*}\scalemath{\begin{alignedat}{2}
        v^\star(\s) = q^\star(\s, \a) &= \smash[b]{\sum_{\s' \in \S}} \tau(\s' | \s, \a) \parens[\big]{r(\s, \a, \s') + \gamma v^\star(\s')} \\
        & \mspace{182mu} \lor p^\star(\s, \a) = 0 &\;& \forall \s \, \forall \a
    \end{alignedat}}\end{equation*}
\end{itemize}
and the dual can be derived as
\begin{equation*}\scalemath{
    \begin{optim}
        \max_{\substack{d(\s) \\ p(\s, \a)}}{\expect[r]_{\subalign{(\s, \a) &\sim p(\cdot, \cdot) \\ \s' &\sim \tau(\cdot | \s, \a)}\mspace{-20mu}}[\big]{r(\s, \a, \s')}}
        \st{
            \sum_{\a \in \A} p(\s, \a) &= d(\s) &\;& \forall \s \\[-6pt]
            d(\s) &= (1 - \gamma) \iota(\s) + \gamma \quad \expect[lr]_{(\s', \a') \sim p(\cdot, \cdot)\;}[\big]{\tau(\s | \s', \a')} && \forall \s \\[-6pt]
            p(\s, \a) &\ge 0 && \forall \s \, \forall \a
        }
    \end{optim}
}\end{equation*}

From the optimality conditions we can prove the following properties of the feasible or optimal primal and dual solutions of the Value Learning LP.
\begin{proposition}\label{prop:vl_prob}
    All feasible dual variables $d$ and $p$ are probability distributions over $\S$ and $\S \times \A$ respectively. Furthermore, they are the visitation densities of the policy $\pi(\a | \s) = \frac{p(\s, \a)}{d(\s)}$ for the considered MDP.
\end{proposition}
\begin{IEEEproof}
    From the equality constraints (or stationarity conditions), we can derive that $\sum_{\s \in \S} d(\s) = \sum_{\s \in \S} \sum_{\a \in \A} p(\s, \a) = 1$. Combined with the inequality constraints (or dual feasibility conditions), which imply $p(\s, \a) \ge 0$ and $d(\s) \ge 0$ for all $\s$ and $\a$, this means $p(\s, \a)$ is a probability distribution over $\S \times \A$, with marginal distribution $d(\s)$.
    This allows to define a policy $\pi$ as the conditional distribution $\pi(\a | \s) = p(\s, \a) / d(\s)$.
    The equality constraints (or stationarity conditions) can then be rewritten as $d(\s) = (1 - \gamma) \iota(\s) + \gamma \sum_{\s' \in \S} \sum_{\a' \in \A} d(\s') \pi(\a' | \s') \tau(\s | \s', \a')$, which is equivalent to the definition of the visitation density $d^\pi$.
    From any feasible solution, we can thus extract a policy $\pi$ with corresponding visitation densities $d(\s) = d^\pi(\s)$ and $p(\s, \a) = p^\pi(\s, \a)$.
\end{IEEEproof}

\begin{reptheorem}{thm:vl_dual}[]\label{thm:vl_dual}
    The policy $\pi^\star$ that can be derived from the optimal dual variables $d^\star$ and $p^\star$ is an optimal policy for the considered MDP.
\end{reptheorem}
\begin{IEEEproof}
    From \propref{prop:vl_prob}, we know that the optimal dual variables correspond to the visitation densities of a policy $\pi^\star$ with $p^\star(\s, \a) = d^{\pi^\star}(\s) \pi^\star(\a | \s)$.
    Furthermore, the optimal dual variables maximize the dual objective, which can be rewritten in terms of the policy $\pi^\star$ as $\expect_{\s \sim d^{\pi^\star}(\cdot), \a \sim \pi^\star(\cdot | \s), \s' \sim \tau(\cdot | \s, \a)}{r(\s, \a, \s')} = \rav^{\pi^\star}$.
    The optimal dual variables $d^\star$ and $p^\star$ are thus the visitation densities of the optimal policy $\pi^*$ for the considered MDP, maximizing the average reward (or expected discounted return).
\end{IEEEproof}

\begin{replemma}{lem:vl_prim}[]\label{lem:vl_prim}
    The optimal primal variables $v^\star$ and $q^\star$ satisfy the Bellman optimality equation and are thus equivalent to the optimal value functions $v^*$ and $q^*$ of the considered MDP.
\end{replemma}
\begin{IEEEproof}
    The complementary slackness conditions imply that for each visited state $\s \in \S^{\pi^\star}$, the primal feasibility constraints become tight for all actions $\a \in \A^{\pi^\star}(\s)$ taken by the policy, as $p^\star(\s, \a) > 0$ for such states and actions.
    Combined with the primal feasibility conditions, we thus have $v^\star(\s) = \max_{\a} \sum_{\s' \in \S} \tau(\s' | \s, \a) \parens[\big]{r(\s, \a, \s') + \gamma v^\star(\s')}$ or $\A^{\pi^\star}(\s) = \A^*(\s) = \argmax_{\a} \sum_{\s' \in \S} \tau(\s' | \s, \a) \parens[\big]{r(\s, \a, \s') + \gamma v^\star(\s')}$.
    Satisfying the Bellman optimality equation, $v^\star$ is thus equivalent to the optimal value function $v^*$ for the considered MDP.
    From the primal feasibility conditions $q^\star(\s, \a) = \sum_{\s' \in \S} \tau(\s' | \s, \a) \parens[\big]{r(\s, \a, \s') + \gamma v^\star(\s')}$, we can then conclude that $q^\star$ is also equivalent to the optimal value function $q^*$ for the considered MDP.
\end{IEEEproof}

\subsection{Policy Evaluation}\label{app:pe}
The primal formulation of the Policy Evaluation problem \eqref{eq:pe_prim} is given by
\begin{equation*}\scalemath{
    \begin{optim}
        \min_{\substack{v(\s) \\ \smash[b]{q(\s, \a)}}}{\expect[r]_{\;\s_0 \sim \iota(\cdot)}[\big]{(1 - \gamma) v(\s_0)}}
        \st{
            v(\s) &\ge \expect[lr]_{\mspace{32mu}\a \sim \pi(\cdot | \s)}[\big]{q(\s, \a)} &\;& \forall \s \\
            q(\s, \a) &= \expect[lr]_{\qquad\s' \sim \tau(\cdot | \s, \a)}[\big]{r(\s, \a, \s') + \gamma v(\s')} && \forall \s \, \forall \a
        }
    \end{optim}
}\end{equation*}
from which the Lagrangian can be written as
\begin{equation*}\scalemath{\begin{aligned}
    \MoveEqLeft[2] \mathcal{L}_{\mathrm{PE}}(v(\s), q(\s, \a), d(\s), p(\s, \a)) \\
    = &\sum_{\s \in \S} \iota(\s) (1 - \gamma) v(\s) - \sum_{\s \in \S} d(\s) \bracks*{v(\s) - \sum_{\a \in \A} \pi(\a | \s) q(\s, \a)} \\
    &-\sum_{\s \in \S} \sum_{\a \in \A} p(\s, \a) \bracks*{q(\s, \a) - \sum_{\s' \in \S} \tau(\s' | \s, \a) \parens[\big]{r(\s, \a, \s') + \gamma v(\s')}} \\
    = &\sum_{\s \in \S}\sum_{\a \in \A} p(\s, \a) \sum_{\s' \in \S} \tau(\s' | \s, \a) r(\s, \a, \s') \\
    &-\sum_{\s \in \S} v(\s) \bracks*{d(\s) - (1 - \gamma)\iota(\s) - \gamma \sum_{\s' \in \S} \sum_{\a' \in \A} p(\s', \a') \tau(\s | \s', \a')} \\
    &- \sum_{\s \in \S} \sum_{\a \in \A} q(\s, \a) \bracks[\big]{p(\s, \a) - d(\s) \pi(\a | \s)}.
\end{aligned}}\end{equation*}
The optimality (KKT) conditions can be derived as
\begin{itemize}
    \setdisplayskips{2pt}{2pt}
    \item Stationarity of respectively $v$ and $q$
    \begin{equation*}\scalemath{\begin{alignedat}{2}
        d^\star(\s) &= (1 - \gamma) \iota(\s) + \gamma \sum_{\s' \in \S} \sum_{\a' \in \A} p^\star(\s', \a') \tau(\s | \s', \a') &\;& \forall \s \\
        p^\star(\s, \a) &= d^\star(\s) \pi(\a | \s) && \forall \s \, \forall \a \\
    \end{alignedat}}\end{equation*}
    \item Primal feasibility
    \begin{equation*}\scalemath{\begin{alignedat}{2}
        v^\star(\s) &\ge \sum_{\a \in \A} \pi(\a | \s) q^\star(\s, \a) &\;& \forall \s \\
        q^\star(\s, \a) &= \sum_{\s' \in \S} \tau(\s' | \s, \a) \parens[\big]{r(\s, \a, \s') + \gamma v^\star(\s')} && \forall \s \, \forall \a
    \end{alignedat}}\end{equation*}
    \item Dual feasibility \qquad $\scalemath{d^\star(\s) \ge 0 \; \forall \s}$
    \item Complementary slackness
    \begin{equation*}\scalemath{\begin{alignedat}{2}
        v^\star(\s) &= \sum_{\a \in \A} \pi(\a | \s) q^\star(\s, \a) \lor d^\star(\s) = 0 &\;& \forall \s
    \end{alignedat}}\end{equation*}
\end{itemize}
and the dual is given by
\begin{equation*}\scalemath{
    \begin{optim}
        \max_{\substack{d(\s) \\ p(\s, \a)}}{\expect[r]_{\subalign{(\s, \a) &\sim p(\cdot, \cdot) \\ \s' &\sim \tau(\cdot | \s, \a)}\mspace{-20mu}}[\big]{r(\s, \a, \s')}}
        \st{
            d(\s) &= (1 - \gamma) \iota(\s) + \gamma \quad \expect[lr]_{(\s', \a') \sim p(\cdot, \cdot)\;}[\big]{\tau(\s | \s', \a')} \ge 0 &\;& \forall \s \\
            p(\s, \a) &= d(\s) \pi(\a | \s) && \forall \s \, \forall \a
        }
    \end{optim}
}\end{equation*}
From the optimality conditions, the following properties of the optimal primal and dual solutions of the Policy Evaluation LP can be proven.
\begin{replemma}{lem:pe_dual}[]\label{lem:pe_dual}
    The optimal dual variables $d^\star$ and $p^\star$ are probability distributions over $\S$ and $\S \times \A$ respectively. Furthermore, they are the visitation densities of the given policy $\pi$ for the considered MDP.
\end{replemma}
\begin{IEEEproof}
    From the stationarity conditions, we can derive that $\sum_{\s \in \S} \sum_{\a \in \A} p^\star(\s, \a) = \sum_{\s \in \S} d^\star(\s) = 1$. Combined with the dual feasibility conditions, which imply $d^\star(\s) \ge 0$ and $p^\star(\s, \a) \ge 0$ for all $\s$ and $\a$, this means the optimal $p^\star(\s, \a)$ is a probability distribution over $\S \times \A$, with marginal distribution $d^\star(\s)$.
    Substituting $p^\star$, the stationarity conditions can be rewritten as $d^\star(\s) = (1 - \gamma) \iota(\s) + \gamma \sum_{\s' \in \S} \sum_{\a' \in \A} d^\star(\s') \pi(\a' | \s') \tau(\s | \s', \a')$, which is equivalent to the definition of the visitation density $d^\pi$. It follows that $p^\star(\s, \a) = d^\star(\s) \pi(\a | \s)$ is also equivalent to the visitation density $p^\pi$.
\end{IEEEproof}

\begin{replemma}{lem:pe_prim}[]\label{lem:pe_prim}
    The optimal primal variables $v^\star$ and $q^\star$ satisfy the Bellman equation and are equivalent to the value functions of the given policy $\pi$ for the considered MDP.
\end{replemma}
\begin{IEEEproof}
    The complementary slackness conditions imply that for each visited state $\s \in \S^{\pi^\star}$, the primal feasibility constraints become tight, as $d^\star(\s) > 0$ for such states.
    Using the primal feasibility conditions to substitute $q^\star$, we then have $v^\star(\s) = \sum_{\a \in \A} \pi(\a | \s) \sum_{\s' \in \S} \tau(\s' | \s, \a) \parens[\big]{r(\s, \a, \s') + \gamma v^\star(\s')}$.
    Satisfying the Bellman equation, $v^\star$ is thus equivalent to the value function $v^\pi$ of the given policy for the considered MDP.
    From the primal feasibility conditions $q^\star(\s, \a) = \sum_{\s' \in \S} \tau(\s' | \s, \a) \parens[\big]{r(\s, \a, \s') + \gamma v^\star(\s')}$, we can then conclude that $q^\star$ is also equivalent to the value function $q^\pi$ of the given policy for the considered MDP.
\end{IEEEproof}

\subsection{Policy Optimization}\label{app:po}
The outer Policy Optimization problem \eqref{eq:po} of the Generalized Policy Iteration (GPI) scheme can be written as
\begin{equation*}\scalemath{
    \begin{optim}
        \max_{\pi(\a | \s)}{\rav^\pi}
        \st{
            \sum_{\a \in \A} \pi(\a | \s) &= 1 &\;& \forall \s \\[-6pt]
            \pi(\a | \s) &\ge 0 && \forall \s \; \forall \a
        }
    \end{optim}
}\end{equation*}
and has as Lagrangian
\begin{equation*}\scalemath{\begin{aligned}
    \MoveEqLeft[2] \mathcal{L}_{\mathrm{PO}}(\pi(\a | \s), \lambda(\s), \mu(\s, \a)) \\
    &= -\rav^\pi - \sum_{\s \in \S} \lambda(\s) \bracks*{1 - \sum_{\a \in \A} \pi(\a | \s)} - \sum_{\s \in \S} \sum_{\a \in \A} \mu(\s, \a) \pi(\a | \s) \\
    &= -\sum_{\s \in \S} \lambda(\s) - \rav^\pi - \sum_{\s \in \S} \sum_{\a \in \A} \pi(\a | \s)  \bracks[\big]{\mu(\s, \a) - \lambda(\s)}.
\end{aligned}}\end{equation*}
The optimality (KKT) conditions are given by
\begin{itemize}
    \setdisplayskips{2pt}{2pt}
    \item Stationarity of $\pi$
    \begin{equation*}\scalemath{\begin{alignedat}{2}
        \lambda^\star(\s) &= \mu^\star(\s, \a) + \diffp{\rav^\pi}{\pi(\a | \s)}[\pi^\star] &\;& \forall \s \, \forall \a \\
    \end{alignedat}}\end{equation*}
    \item Primal feasibility
    \begin{equation*}\scalemath{\begin{alignedat}{2}
        \sum_{\a \in \A} \pi^\star(\a | \s) &= 1 \land \pi^\star(\a | \s) \ge 0 &\;& \forall \s \, \forall \a
    \end{alignedat}}\end{equation*}
    \item Dual feasibility \qquad $\scalemath{\mu^\star(\s, \a) \ge 0 \; \forall \s \, \forall \a}$
    \item Complementary slackness
    \begin{equation*}\scalemath{\begin{alignedat}{2}
        \mu^\star(\s, \a) &= 0 \lor \pi^\star(\a | \s) = 0 &\;& \forall \s \, \forall \a
    \end{alignedat}}\end{equation*}
\end{itemize}
From these optimality conditions, we can derive following property of the optimal solution to the Policy Optimization problem.
\begin{proposition}\label{prop:po_cond}
    The optimal solution $\pi^\star$ is a conditional probability distribution over $\A$ given $\S$, and actions taken by this policy maximize the partial derivative of the average reward $\rav^\pi$ with respect to the action sampling probability.
\end{proposition}
\begin{IEEEproof}
    From the primal feasibility conditions, it immediately follows that $\pi^\star$ is a conditional probability distribution over $\A$ given $\S$.
    Combining the dual feasibility and stationarity conditions, we can derive that $\lambda^\star(\s) \ge \diffp{\rav^\pi}{\pi(\a | \s)}[\pi^\star]$ for all $\s$ and $\a$. The complementary slackness conditions then imply that this inequality becomes tight for all actions $\a \in \A^{\pi^\star}(\s)$ taken by the policy, i.e. $\A^{\pi^\star}(\s) = \argmax_{\a} \diffp{\rav^\pi}{\pi(\a | \s)}[\pi^\star]$.
\end{IEEEproof}
More interesting properties can be derived when integrating an inner Policy Evaluation problem, resulting in a GPI scheme. The partial derivative of the average reward $\rav^\pi$ can then typically be evaluated using Danskin's theorem.
\begin{reptheorem}{thm:po}[]
    The optimal solution $\pi^\star$ of the nested Policy Optimization \eqref{eq:po} and Policy Evaluation \eqref{eq:pe} problems, is the optimal policy $\pi^*$ for the considered MDP.
\end{reptheorem}
\begin{IEEEproof}
    The partial derivative of the expected discounted return $\rav^\pi$ can be evaluated using Danskin's theorem as $\diffp{\rav^\pi}{\pi(\a | \s)}[\pi^\star] = \diffp{\mathcal{L}_{\mathrm{PE}}}{\pi(\a | \s)}[\pi^\star,d^\star,q^\star] = d^{\pi^\star}(\s) q^{\pi^\star}(\s, \a)$.
    From the optimality conditions and \propref{prop:po_cond}, we then have $\lambda^\star(\s) \ge d^{\pi^\star}(\s) q^{\pi^\star}(\s, \a)$ for all $\s$ and $\a$, and $\A^{\pi^\star}(\s) = \argmax_{\a} d^{\pi^\star}(\s) q^{\pi^\star}(\s, \a)$.
    Hence, $\pi^\star$ is a greedy policy with respect to its own value function $q^{\pi^\star}$. This means $q^{\pi^\star}$ satisfies the Bellman optimality equation and $\pi^\star$ is an optimal policy for the considered MDP.
\end{IEEEproof}

\subsection{Entropy Regularization}\label{app:er}
The altered policy evaluation dual with added KL-divergence regularization term is given by
\begin{equation*}\scalemath{
    \begin{optim}
        \max_{\substack{d(\s) \\ p(\s, \a)}}{\expect[r]_{\subalign{(\s, \a) &\sim p(\cdot, \cdot) \\ \s' &\sim \tau(\cdot | \s, \a)}\mspace{-20mu}}[\big]{r(\s, \a, \s')} - \alpha \expect[lr]_{\mspace{20mu}\s \sim d(\cdot)}[\big]{\KL[\big]{\pi(\cdot | \s)}{\pi_{\textsc{t}}(\cdot | \s)}}}
        \st{
            d(\s) &= (1 - \gamma) \iota(\s) + \gamma \quad \expect[lr]_{(\s', \a') \sim p(\cdot, \cdot)\;}[\big]{\tau(\s | \s', \a')} \ge 0 &\;& \forall \s \\
            p(\s, \a) &= d(\s) \pi(\a | \s) && \forall \s \, \forall \a
        }
    \end{optim}
}\end{equation*}
The Lagrangian of this regularized LP is given by
\begin{equation*}\scalemath{\begin{aligned}
    \MoveEqLeft[2] \mathcal{L}_{\mathrm{ER}}(d(\s), p(\s, \a), v(\s), q(\s, \a), \mu(\s)) \\
    = &-\sum_{\s \in \S}\sum_{\a \in \A} p(\s, \a) \sum_{\s' \in \S} \tau(\s' | \s, \a) r(\s, \a, \s') \\
    &+ \sum_{\s \in \S} d(\s) \sum_{\a \in \A} \pi(\a | \s) \alpha \log \frac{\pi(\a | \s)}{\pi_{\textsc{t}}(\a | \s)} \\
    &+ \sum_{\s \in \S} v(\s) \bracks*{d(\s) - (1 - \gamma)\iota(\s) - \gamma \sum_{\s' \in \S} \sum_{\a' \in \A} p(\s', \a') \tau(\s | \s', \a')} \\
    &- \sum_{\s \in \S} \mu(\s) d(\s) + \sum_{\s \in \S} \sum_{\a \in \A} q(\s, \a) \bracks[\big]{p(\s, \a) - d(\s) \pi(\a | \s)} \\
    = &-\sum_{\s \in \S} \iota(\s) (1 - \gamma) v(\s) \\
    &+ \sum_{\s \in \S} d(\s) \bracks*{v(\s) - \mu(\s) - \sum_{\a \in \A} \pi(\a | \s) \parens*{q(\s, \a) - \alpha \log \frac{\pi(\a | \s)}{\pi_{\textsc{t}}(\a | \s)}}} \\
    &+\sum_{\s \in \S} \sum_{\a \in \A} p(\s, \a) \bracks[\Bigg]{q(\s, \a) - \sum_{\s' \in \S} \tau(\s' | \s, \a) \parens[\big]{r(\s, \a, \s') + \gamma v(\s')}}.
\end{aligned}}\end{equation*}
The optimality (KKT) conditions can then be written as
\begin{itemize}
    \setdisplayskips{2pt}{2pt}
    \item Stationarity of respectively $d$ and $p$
    \begin{equation*}\scalemath{\begin{alignedat}{2}
        \mu^\star(\s) &= v^\star(\s) - \sum_{\a \in \A} \pi(\a | \s) \parens*{q^\star(\s, \a) - \alpha \log \frac{\pi(\a | \s)}{\pi_{\textsc{t}}(\a | \s)}} &\;& \forall \s \\
        q^\star(\s, \a) &= \sum_{\s' \in \S} \tau(\s' | \s, \a) \parens[\big]{r(\s, \a, \s') + \gamma v^\star(\s')} && \forall \s \, \forall \a
    \end{alignedat}}\end{equation*}
    \item Dual feasibility
    \begin{equation*}\scalemath{\begin{alignedat}{2}
        d^\star(\s) &= (1 - \gamma) \iota(\s) \\
        &\mspace{40mu}+ \gamma \sum_{\s' \in \S} \sum_{\a' \in \A} p^\star(\s', \a') \tau(\s | \s', \a') \ge 0 &\;& \forall \s \\
        p^\star(\s, \a) &= d^\star(\s) \pi(\a | \s) && \forall \s \, \forall \a
    \end{alignedat}}\end{equation*}
    \item Primal feasibility
    \begin{equation*}\scalemath{\begin{alignedat}{2}
        v^\star(\s) &\ge \sum_{\a \in \A} \pi(\a | \s) \parens*{q^\star(\s, \a) - \alpha \log \frac{\pi(\a | \s)}{\pi_{\textsc{t}}(\a | \s)}} &\;& \forall \s
    \end{alignedat}}\end{equation*}
    \item Complementary slackness
    \begin{equation*}\scalemath{\begin{alignedat}{2}
        v^\star(\s) &= \smash[b]{\sum_{\a \in \A}} \pi(\a | \s) \parens*{q^\star(\s, \a) - \alpha \log \frac{\pi(\a | \s)}{\pi_{\textsc{t}}(\a | \s)}} \\
            & \mspace{220mu} \lor d^\star(\s) = 0 &\;& \forall \s
    \end{alignedat}}\end{equation*}
\end{itemize}
and the primal formulation is given by
\begin{equation*}\scalemath{
    \begin{optim}
        \min_{\substack{v(\s) \\ \smash[b]{q(\s, \a)}}}{\expect[r]_{\;\s_0 \sim \iota(\cdot)}[\big]{(1 - \gamma) v(\s_0)}}
        \st{
            v(\s) &\ge \expect[l]_{\mspace{32mu}\a \sim \pi(\cdot | \s)}[\bigg]{q(\s, \a) - \alpha \log \frac{\pi(\a | \s)}{\pi_{\textsc{t}}(\a | \s)}} &\;& \forall \s \\
            q(\s, \a) &= \expect[lr]_{\qquad\s' \sim \tau(\cdot | \s, \a)}[\big]{r(\s, \a, \s') + \gamma v(\s')} && \forall \s \, \forall \a
        }
    \end{optim}
}\end{equation*}
\lemref{lem:pe_dual} can be proven for this regularized Policy Evaluation problem without major changes. \lemref{lem:pe_prim} also holds, albeit with a modified definition of the reward \eqref{eq:er_mod} and value functions, taking the entropy regularization into account.
\begin{replemma}{lem:er_prim}[]
    The optimal primal variables $v^\star$ and $q^\star$ satisfy the entropy regularized Bellman equation and are equivalent to the entropy regularized value functions of the given policy $\pi$ for the considered MDP.
\end{replemma}
\begin{IEEEproof}
    Following similar arguments as in \lemref{lem:pe_prim}, we obtain $v^\star(\s) = \sum_{\a \in \A} \pi(\a | \s) \sum_{\s' \in \S} \tau(\s' | \s, \a) \parens[\big]{r(\s, \a, \s') - \alpha \log \frac{\pi(\a | \s)}{\pi_{\textsc{t}}(\a | \s)} + \gamma v^\star(\s')}$.
    Denoting by $r_\mathrm{ER}^{}(\s, \a, \s') = r(\s, \a, \s') - \alpha \log \frac{\pi(\a | \s)}{\pi_{\textsc{t}}(\a | \s)}$ the entropy-regularized reward, we see that $v^\star$ satisfies the Bellman equation using this modified reward. We can thus conclude that $v^\star$ (and $q^\star$) are equivalent to the entropy-regularized value function $v_\mathrm{ER}^\pi$ (and $q_\mathrm{ER}^\pi$) of the given policy for the considered MDP.
\end{IEEEproof}

When plugging this entropy regularized policy evaluation LP in the Policy Optimization problem, we can prove the following property.
\begin{reptheorem}{thm:er_pol}[]
    The optimal solution $\pi^\star$ of the nested Policy Optimization \eqref{eq:po} and Entropy-Regularized Policy Evaluation \eqref{eq:er} problems, is a conditional Boltzmann policy with as energy function the negation of the sum of the entropy-regularized value function $q_\mathrm{ER}^\star$ and the teacher policy's log probability.
\end{reptheorem}
\begin{IEEEproof}
    The partial derivative of the average reward $\rav^\pi$ can be evaluated using Danskin's theorem as $\diffp{\rav^\pi}{\pi(\a | \s)}[\pi^\star] = -\diffp{\mathcal{L}_{\mathrm{ER}}}{\pi(\a | \s)}[\pi^\star,d^\star,q^\star] = d^{\pi^\star}(\s) \bracks[\Big]{q_{\mathrm{ER}}^{\pi^\star}(\s, \a) - \alpha - \alpha \log \frac{\pi^\star(\a | \s)}{\pi_{\textsc{t}}(\a | \s)}}$.
    From the optimality conditions and \propref{prop:po_cond}, we then have $\lambda^\star(\s) \ge d^{\pi^\star}(\s) \bracks[\Big]{q_{\mathrm{ER}}^{\pi^\star}(\s, \a) - \alpha - \alpha \log \frac{\pi^\star(\a | \s)}{\pi_{\textsc{t}}(\a | \s)}}$ for all $\s$ and $\a$, and the equality holds for all actions taken by the policy $\A^{\pi^\star}(\s) = \argmax_{\a} \bracks[\Big]{q_{\mathrm{ER}}^{\pi^\star}(\s, \a) - \alpha \log \frac{\pi^\star(\a | \s)}{\pi_{\textsc{t}}(\a | \s)}}$.
    This means that $\pi^\star(\a | \s) > 0$ for all actions; as otherwise the maximum would become unbounded and be reached for the actions that are not taken. Hence $\alpha \log \frac{\pi^\star(\a | \s)}{\pi_{\textsc{t}}(\a | \s)} = q_{\mathrm{ER}}^{\pi^\star}(\s, \a) - \frac{\lambda^\star(\s)}{d^{\pi^\star}(\s)} - \alpha$ holds for all states and actions, implying that $\pi^\star(\a | \s) \propto \pi_{\textsc{t}}(\a | \s) \exp\parens*{\frac{q_{\mathrm{ER}}^{\pi^\star}(\s, \a)}{\alpha}}$.
    The optimal policy $\pi^\star$ is thus a conditional Boltzmann distribution with energy function $- q_{\mathrm{ER}}^{\pi^\star}(\s, \a) - \alpha \log \pi_{\textsc{t}}(\a | \s)$ and temperature $\alpha$.
\end{IEEEproof}
The resulting nested optimization problem can be seen as a \emph{Soft Policy Iteration} scheme, for which convergence proofs can be derived, e.g. by \AuthorCite{Haarnoja2018} for the case of maximum entropy RL (or a uniform teacher policy).

\subsection{Constrained Reinforcement Learning}\label{app:crl}
The constrained dual value learning problem can be written as
\begin{equation*}\scalemath{
    \begin{optim}
        \max_{\substack{d(\s) \\ p(\s, \a)}}{\expect[r]_{\subalign{(\s, \a) &\sim p(\cdot, \cdot) \\ \s' &\sim \tau(\cdot | \s, \a)}\mspace{-20mu}}[\big]{r(\s, \a, \s')}}
        \st{
            \sum_{\a \in \A} p(\s, \a) &= d(\s) &\;& \forall \s \\[-6pt]
            d(\s) &= (1 - \gamma) \iota(\s) + \gamma \quad \expect[lr]_{(\s', \a') \sim p(\cdot, \cdot)\;}[\big]{\tau(\s | \s', \a')} && \forall \s \\[-6pt]
            p(\s, \a) &\ge 0 && \forall \s \, \forall \a \\
            (1 - \gamma) V_k &\le \quad\expect[lr]_{\subalign{(\s, \a) &\sim p(\cdot, \cdot) \\ \s' &\sim \tau(\cdot | \s, \a)}\mspace{-20mu}}{r_k(\s, \a, \s')} && \forall k
        }
    \end{optim}
}\end{equation*}
The Lagrangian of this altered dual problem can be obtained as
\begin{equation*}\scalemath{\begin{aligned}
    \MoveEqLeft[2] \mathcal{L}_{\mathrm{CRL}}(d(\s), p(\s, \a), \lambda(\s), v(\s), \mu(\s, \a), w_k) \\
    = &-\sum_{\s \in \S}\sum_{\a \in \A} p(\s, \a) \sum_{\s' \in \S} \tau(\s' | \s, \a) r(\s, \a, \s') \\
    &+ \sum_{\s \in \S} \lambda(\s) \bracks*{\sum_{\a \in \A} p(\s, \a) - d(\s)} \\
    &+ \sum_{\s \in \S} v(\s) \bracks*{d(\s) - (1 - \gamma)\iota(\s) - \gamma \sum_{\s' \in \S} \sum_{\a' \in \A} p(\s', \a') \tau(\s | \s', \a')} \\
    &- \sum_{\s \in \S} \sum_{\a \in \A} \mu(\s, \a) p(\s, \a) \\
    &- \sum_{k = 0}^{K} w_k \bracks*{\sum_{\s \in \S}\sum_{\a \in \A} p(\s, \a) \sum_{\s' \in \S} \tau(\s' | \s, \a) r_k(\s, \a, \s') - (1 - \gamma) V_k} \\
    = &-\sum_{\s \in \S} \iota(\s) (1 - \gamma) v(\s) + \sum_{k = 0}^{K} w_k (1 - \gamma) V_k \\
    &+ \sum_{\s \in \S} d(\s) \bracks*{v(\s) - \lambda(\s)} \\[-6pt]
    &+ \sum_{\s \in \S} \sum_{\a \in \A} p(\s, \a) \bracks[\Bigg]{\lambda(\s) - \mu(\s, \a) \\[-14pt]
    &\mspace{182mu} - \sum_{\s' \in \S} \tau(\s' | \s, \a) \parens[\bigg]{r(\s, \a, \s') + \gamma v(\s') \\[-12pt]
    &\mspace{326mu} + \sum_{k = 0}^{K} w_k r_k(\s, \a, \s')}},
\end{aligned}}\end{equation*}
resulting in following optimality (KKT) conditions
\begin{itemize}
    \setdisplayskips{2pt}{2pt}
    \item Stationarity of respectively $d$ and $p$
    \begin{equation*}\scalemath{\begin{alignedat}{2}
        \lambda^\star(\s) &= v^\star(\s) &\;& \forall \s \\
        \mu^\star(\s, \a) &= v^\star(\s) - \sum_{\s' \in \S} \tau(\s' | \s, \a) \bracks[\bigg]{r(\s, \a, \s') + \gamma v^\star(\s') \\[-12pt]
        &\mspace{214mu} + \sum_{k = 0}^{K} w_k^\star r_k(\s, \a, \s')} && \forall \s \, \forall \a
    \end{alignedat}}\end{equation*}
    \item Dual feasibility
    \begin{equation*}\scalemath{\mspace{-10mu}\begin{alignedat}{2}
        \sum_{\a \in \A} p^\star(\s, \a) &= d^\star(\s) &\;& \forall \s \\[-6pt]
        d^\star(\s) &= (1 - \gamma) \iota(\s) + \gamma \sum_{\s' \in \S} \sum_{\a' \in \A} p^\star(\s', \a') \tau(\s | \s', \a') && \forall \s \\[-6pt]
        p^\star(\s, \a) &\ge 0 && \forall \s \, \forall \a \\
        (1 - \gamma) V_k &\le \sum_{\s \in \S}\sum_{\a \in \A} p^\star(\s, \a) \sum_{\s' \in \S} \tau(\s' | \s, \a) r_k(\s, \a, \s') && \forall k
    \end{alignedat}}\end{equation*}
    \item Primal feasibility, introducing auxiliary $q$
    \begin{equation*}\scalemath{\begin{alignedat}{2}
        v^\star(\s) &\ge q^\star(\s, \a) &\;& \forall \s \, \forall \a \\[-8pt]
        q^\star(\s, \a) &= \sum_{\s' \in \S} \tau(\s' | \s, \a) \bracks[\bigg]{r(\s, \a, \s') + \sum_{k = 0}^{K} w_k^\star r_k(\s, \a, \s') \\[-14pt]
        &\mspace{292mu} + \gamma v^\star(\s')} && \forall \s \, \forall \a \\
        w_k^\star &\ge 0 && \forall k
    \end{alignedat}}\end{equation*}
    \item Complementary slackness
    \begin{equation*}\scalemath{\begin{alignedat}{2}
        v^\star(\s) &= q^\star(\s, \a) 
        \lor p^\star(\s, \a) = 0 &\;& \forall \s \, \forall \a \\
        (1 - \gamma) V_k &= \sum_{\s \in \S}\sum_{\a \in \A} p^\star(\s, \a) \sum_{\s' \in \S} \tau(\s' | \s, \a) r_k(\s, \a, \s') \\[-8pt]
        & \mspace{266mu} \lor w_k^\star = 0 && \forall k
    \end{alignedat}}\end{equation*}
\end{itemize}
The corresponding primal formulation is thus derived as
\begin{equation*}\scalemath{
    \begin{optim}
        \min_{\substack{v(\s), w_k \\ \smash[b]{q(\s, \a)}}}{\;\expect[r]_{\;\s_0 \sim \iota(\cdot)}[\big]{(1 - \gamma) v(\s_0)} - \sum_{k = 0}^{K} w_k (1 - \gamma) V_k}
        \st{
            v(\s) &\ge q(\s, \a) &\;& \forall \s \, \forall \a \\
            q(\s, \a) &= \expect[l]_{\qquad\s' \sim \tau(\cdot | \s, \a)}\bracks[\bigg]{r(\s, \a, \s') + \sum_{k = 0}^{K} w_k r_k(\s, \a, \s') + \gamma v(\s')} && \forall \s \, \forall \a \\
            w_k &\ge 0 && \forall k
        }
    \end{optim}
}\end{equation*}
From the optimality conditions, it is immediately clear \propref{prop:vl_prob} still holds for this constrained variant and it can be extended as follows.
\begin{proposition}
    The policy $\pi(\a | \s)$ that can be derived from any feasible dual variables $d$ and $p$ is a feasible policy for the considered CMDP.
\end{proposition}
\begin{IEEEproof}
    From \propref{prop:vl_prob}, we know that the dual variables correspond to the visitation densities of a policy $\pi$ with $p(\s, \a) = d^\pi(\s) \pi(\a | \s)$. The $K$ additional dual constraints can hence be rewritten in terms of the policy $\pi$ as $V_k \le (1 - \gamma)^{-1} \expect_{\s \sim d^\pi(\cdot), \a \sim \pi(\cdot | \s), \s' \sim \tau(\cdot | \s, \a)}{r_k(\s, \a, \s')} = \nu_k^\pi$. The dual variables $d$ and $p$ are thus the visitation densities of a policy satisfying the CMDP constraints.
\end{IEEEproof}
As a result, \thmref{thm:vl_dual} can also be straightforwardly adapted as follows.
\begin{reptheorem}{thm:crl_dual}[]
    The policy $\pi^\star$ that can be derived from the optimal dual variables $d^\star$ and $p^\star$ is an optimal policy for the considered CMDP.
\end{reptheorem}
\begin{IEEEproof}
    Combine the results from the previous Proposition with the proof of \thmref{thm:vl_dual}.
\end{IEEEproof}
\lemref{lem:vl_prim} also holds, albeit for the modified reward function $r_{\mathrm{CRL}}^{}$ \eqref{eq:crl_mod}, with learnable parameters $w_k$, and corresponding value functions.
\begin{replemma}{lem:crl_prim}[]
    The optimal primal variables $v^\star$ and $q^\star$ satisfy the Bellman optimality equation and are thus equivalent to the optimal adjusted value functions $v_{\mathrm{CRL}}^*$ and $q_{\mathrm{CRL}}^*$ for the considered MDP with optimal modified reward $r_{\mathrm{CRL}}^\star(\s, \a, \s') = r(\s, \a, \s') + \sum_{k = 0}^{K} w_k^\star r_k(\s, \a, \s')$.
\end{replemma}
\begin{IEEEproof}
    Following the same arguments as in \lemref{lem:vl_prim}, we obtain $v^\star(\s) = \max_{\a} \sum_{\s' \in \S} \tau(\s' | \s, \a) \parens[\big]{r_{\mathrm{CRL}}^\star(\s, \a, \s') + \gamma v^\star(\s')}$.
    Satisfying the Bellman optimality equation, $v^\star$ is thus equivalent to the optimal value function $v_{\mathrm{CRL}}^*$ for the considered MDP with altered reward $r_{\mathrm{CRL}}^\star$.
    From the primal feasibility conditions $q^\star(\s, \a) = \sum_{\s' \in \S} \tau(\s' | \s, \a) \parens[\big]{r_{\mathrm{CRL}}^\star(\s, \a, \s') + \gamma v^\star(\s')}$, we then also obtain the equivalency between $q^\star$ and $q_{\mathrm{CRL}}^*$.
\end{IEEEproof}
Finally, these optimal adjusted value functions and optimal reward modifications satisfy following properties.
\begin{reptheorem}{thm:crl_greedy}[]
    The optimal policy $\pi^\star$ is greedy with respect to the optimal adjusted value functions $v_{\mathrm{CRL}}^*$ and $q_{\mathrm{CRL}}^*$.
\end{reptheorem}
\begin{IEEEproof}
    Following the same arguments as in the proof for \lemref{lem:vl_prim}, we obtain $\A^{\pi^\star}(\s) = \A^*(\s) = \argmax_{\a} \sum_{\s' \in \S} \tau(\s' | \s, \a) \parens[\big]{r_{\mathrm{CRL}}^\star(\s, \a, \s') + \gamma v^\star(\s')}$.
    From the previous Lemma, it then follows that $\A^{\pi^\star}(\s) = \argmax_{\a} q_{\mathrm{CRL}}^*(\s, \a)$, showing the greediness of $\pi^\star$ with respect to the optimal adjusted value function.
\end{IEEEproof}
\begin{repproposition}{prop:crl_mod}[]
    The optimal modified reward $r_{\mathrm{CRL}}^\star$ is only altered by the tight constraints $V_k = \nu_k^{\pi^\star}$.
\end{repproposition}
\begin{IEEEproof}
    This follows immediately from the complementary slackness conditions, as $w_k^\star = 0$ (and hence the corresponding $r_k$ does not influence the modified reward $r_{\mathrm{CRL}}$) for the strictly satisfied constraints $V_k < \nu_k^{\pi^\star}$.
\end{IEEEproof}

\subsection{Visitation Density Bounds}\label{app:vdb}
\subsubsection{Value Learning}
The altered dual value learning problem with state-visitation density bounds is given by
\begin{equation*}\scalemath{
    \begin{optim}
        \max_{\substack{d(\s) \\ p(\s, \a)}}{\expect[r]_{\subalign{(\s, \a) &\sim p(\cdot, \cdot) \\ \s' &\sim \tau(\cdot | \s, \a)}\mspace{-20mu}}[\big]{r(\s, \a, \s')}}
        \st{
            \sum_{\a \in \A} p(\s, \a) &= d(\s) &\;& \forall \s \\[-6pt]
            d(\s) &= (1 - \gamma) \iota(\s) + \gamma \quad \expect[lr]_{(\s', \a') \sim p(\cdot, \cdot)\;}[\big]{\tau(\s | \s', \a')} && \forall \s \\[-6pt]
            p(\s, \a) &\ge 0 && \forall \s \, \forall \a \\
            \lb{d}(\s) &\le d(\s) \le \ub{d}(\s) && \forall \s
        }
    \end{optim}
}\end{equation*}
for which the Lagrangian can be derived as
\begin{equation*}\scalemath{\begin{aligned}
    \MoveEqLeft[2] \mathcal{L}_{VDB}(d(\s), p(\s, \a), \lambda(\s), v(\s), \mu(\s, \a), \lb{r}(\s), \ub{r}(\s)) \\
    = &-\sum_{\s \in \S}\sum_{\a \in \A} p(\s, \a) \sum_{\s' \in \S} \tau(\s' | \s, \a) r(\s, \a, \s') \\
    &+ \sum_{\s \in \S} \lambda(\s) \bracks*{\sum_{\a \in \A} p(\s, \a) - d(\s)} \\
    &+ \sum_{\s \in \S} v(\s) \bracks*{d(\s) - (1 - \gamma)\iota(\s) - \gamma \sum_{\s' \in \S} \sum_{\a' \in \A} p(\s', \a') \tau(\s | \s', \a')} \\
    &- \sum_{\s \in \S} \sum_{\a \in \A} \mu(\s, \a) p(\s, \a) \\
    &- \sum_{\s \in \S} \lb{r}(\s) \bracks*{d(\s) - \lb{d}(\s)} - \sum_{\s \in \S} \ub{r}(\s) \bracks*{\ub{d}(\s) - d(\s)} \\
    = &-\sum_{\s \in \S} \iota(\s) (1 - \gamma) v(\s) + \sum_{\s \in \S} \lb{r}(\s) \lb{d}(\s) - \sum_{\s \in \S} \ub{r}(\s) \ub{d}(\s) \\
    &+ \sum_{\s \in \S} d(\s) \bracks*{v(\s) - \lambda(\s) - \lb{r}(\s) + \ub{r}(\s)} \\
    &+ \sum_{\s \in \S} \sum_{\a \in \A} p(\s, \a) \bracks[\Bigg]{\lambda(\s) - \mu(\s, \a) \\[-16pt]
    &\mspace{182mu}- \sum_{\s' \in \S} \tau(\s' | \s, \a) \parens[\big]{r(\s, \a, \s') + \gamma v(\s')}},
\end{aligned}}\end{equation*}
resulting in following optimality (KKT) conditions
\begin{itemize}
    \setdisplayskips{2pt}{2pt}
    \item Stationarity of respectively $d$ and $p$
    \begin{equation*}\scalemath{\begin{alignedat}{2}
        \lambda^\star(\s) &= v^\star(\s) - \lb{r}^\star(\s) + \ub{r}^\star(\s) &\;& \forall \s \\
        \mu^\star(\s, \a) &= v^\star(\s) - \lb{r}^\star(\s) + \ub{r}^\star(\s) \\
        &\mspace{66mu} - \sum_{\s' \in \S} \tau(\s' | \s, \a) \parens[\big]{r(\s, \a, \s') + \gamma v^\star(\s')} && \forall \s \, \forall \a
    \end{alignedat}}\end{equation*}
    \item Dual feasibility
    \begin{equation*}\scalemath{\mspace{-10mu}\begin{alignedat}{2}
        \sum_{\a \in \A} p^\star(\s, \a) &= d^\star(\s) &\;& \forall \s \\
        d^\star(\s) &= (1 - \gamma) \iota(\s) + \gamma \sum_{\s' \in \S} \sum_{\a' \in \A} p^\star(\s', \a') \tau(\s | \s', \a') && \forall \s \\
        p^\star(\s, \a) &\ge 0 && \forall \s \, \forall \a \\
        \lb{d}(\s) &\le d^\star(\s) \le \ub{d}(\s) && \forall \s
    \end{alignedat}}\end{equation*}
    \item Primal feasibility, introducing auxiliary $q$
    \begin{equation*}\scalemath{\begin{alignedat}{2}
        v^\star(\s) &\ge q^\star(\s, \a) &\;& \forall \s \, \forall \a \\
        q^\star(\s, \a) &= \smash[b]{\sum_{\s' \in \S}} \tau(\s' | \s, \a) \parens[\big]{r(\s, \a, \s') + \lb{r}^\star(\s) - \ub{r}^\star(\s) \\
        &\mspace{274mu} + \gamma v^\star(\s')} && \forall \s \, \forall \a \\
        \lb{r}^\star(\s) &\ge 0 \land \ub{r}^\star(\s) \ge 0 && \forall \s
    \end{alignedat}}\end{equation*}
    \item Complementary slackness
    \begin{equation*}\scalemath{\begin{alignedat}{2}
        v^\star(\s) &= q^\star(\s, \a) 
        \lor p^\star(\s, \a) = 0 &\;& \forall \s \, \forall \a \\
        d^\star(\s) &= \lb{d}(\s) \lor \lb{r}^\star(\s) = 0 && \forall \s \\
        d^\star(\s) &= \ub{d}(\s) \lor \ub{r}^\star(\s) = 0 && \forall \s
    \end{alignedat}}\end{equation*}
\end{itemize}
and primal formulation
\begin{equation*}\scalemath{
    \begin{optim}
        \min_{\substack{v(\s), q(\s, \a) \\ \smash[b]{\lb{r}(\s), \ub{r}(\s)}}}{\quad\expect[r]_{\;\s_0 \sim \iota(\cdot)}[\big]{(1 - \gamma) v(\s_0)} - \sum_{\s \in \S} \lb{r}(\s) \lb{d}(\s) + \sum_{\s \in \S} \ub{r}(\s) \ub{d}(\s)}
        \st{
            v(\s) &\ge q(\s, \a) &\;& \forall \s \, \forall \a \\
            q(\s, \a) &= \expect[lr]_{\qquad\s' \sim \tau(\cdot | \s, \a)}[\big]{r(\s, \a, \s') + \lb{r}(\s) - \ub{r}(\s) + \gamma v(\s')} && \forall \s \, \forall \a \\
            \lb{r}(\s) &\ge 0 \land \ub{r}(\s) \ge 0 && \forall \s
        }
    \end{optim}
}\end{equation*}
From the optimality conditions, it is immediately clear \propref{prop:vl_prob} still holds for this constrained variant and it can be extended as follows.
\begin{proposition}
    The policy $\pi(\a | \s)$ that can be derived from any feasible dual variables $d$ and $p$ is a feasible policy for the considered MDP with visitation density bounds.
\end{proposition}
\begin{IEEEproof}
    From \propref{prop:vl_prob}, we know that the dual variables correspond to the visitation densities of a policy $\pi$. The additional constraints can thus be rewritten as $\lb{d}(\s) \le d^\pi(\s) \le \ub{d}(\s)$. The dual variables $d$ and $p$ are thus the visitation densities of a policy satisfying those visitation density bounds.
\end{IEEEproof}
Consequently, \thmref{thm:vl_dual} can also be straightforwardly adapted as follows.
\begin{lemma}\label{lem:vdb_opt}
    The policy $\pi^\star$ that can be derived from the optimal dual variables $d^\star$ and $p^\star$ is an optimal policy for the MDP with visitation density bounds.
\end{lemma}
\begin{IEEEproof}
    Combine the results from the previous Proposition with the proof of \thmref{thm:vl_dual}.
\end{IEEEproof}
\lemref{lem:vl_prim} also holds, albeit for the modified reward function $r_{\mathrm{VDB}}^{}(\s, \a, \s') = r(\s, \a, \s') + \lb{r}(\s) - \ub{r}(\s)$, with learnable parameters $\lb{r}$ and $\ub{r}$, and corresponding value functions.
\begin{lemma}
    The optimal primal variables $v^\star$ and $q^\star$ satisfy the Bellman optimality equation and are thus equivalent to the optimal adjusted value functions $v_{\mathrm{VDB}}^*$ and $q_{\mathrm{VDB}}^*$ for the considered MDP with optimal modified reward $r_{\mathrm{VDB}}^\star(\s, \a, \s') = r(\s, \a, \s') + \lb{r}^\star(\s) - \ub{r}^\star(\s)$.
\end{lemma}
\begin{IEEEproof}
    Following the same arguments as in \lemref{lem:vl_prim}, we obtain $v^\star(\s) = \max_{\a} \sum_{\s' \in \S} \tau(\s' | \s, \a) \parens[\big]{r_{\mathrm{VDB}}^\star(\s, \a, \s') + \gamma v^\star(\s')}$.
    Satisfying the Bellman optimality equation, $v^\star$ is thus equivalent to the optimal value function $v_{\mathrm{VDB}}^*$ for the considered MDP with altered reward $r_{\mathrm{VDB}}^\star$.
    From the primal feasibility conditions $q^\star(\s, \a) = \sum_{\s' \in \S} \tau(\s' | \s, \a) \parens[\big]{r_{\mathrm{VDB}}^\star(\s, \a, \s') + \gamma v^\star(\s')}$, we then also obtain the equivalency between $q^\star$ and $q_{\mathrm{VDB}}^*$.
\end{IEEEproof}
Finally, these optimal adjusted value functions and optimal reward modifications satisfy following properties.
\begin{lemma}\label{lem:vdb_greedy}
    The optimal policy $\pi^\star$ is greedy with respect to the optimal adjusted value functions $v_{\mathrm{VDB}}^*$ and $q_{\mathrm{VDB}}^*$.
\end{lemma}
\begin{IEEEproof}
    Following the same arguments as in the proof for \lemref{lem:vl_prim}, we obtain $\A^{\pi^\star}(\s) = \A^*(\s) = \argmax_{\a} \sum_{\s' \in \S} \tau(\s' | \s, \a) \parens[\big]{r_{\mathrm{VDB}}^\star(\s, \a, \s') + \gamma v^\star(\s')}$.
    From the previous Lemma, it then follows that $\A^{\pi^\star}(\s) = \argmax_{\a} q_{\mathrm{VDB}}^*(\s, \a)$, showing the greediness of $\pi^\star$ with respect to the optimal adjusted value functions.
\end{IEEEproof}
\begin{lemma}
    The optimal modified reward $r_{\mathrm{VDB}}^\star$ is only altered by the tight bounds $\lb{d}(\s) = d^{\pi^\star}(\s)$ and $d^{\pi^\star}(\s) = \ub{d}(\s)$.
\end{lemma}
\begin{IEEEproof}
    This follows immediately from the complementary slackness conditions, as $\lb{r}^\star(\s) = 0$ for the strictly satisfied constraints $\lb{d}(\s) < d^{\pi^\star}(\s)$, while $\ub{r}^\star(\s) = 0$ for the strictly satisfied constraints $d^{\pi^\star}(\s) < \ub{d}(\s)$.
\end{IEEEproof}
In case there is no policy that satisfies all additional constraints, the dual problem effectively becomes infeasible and the primal becomes unbounded, caused by the $\lb{r}(\s, \a)$ and $\ub{r}(\s, \a)$ of the infeasible constraints approaching infinity.

\subsubsection{Policy Iteration}
The altered dual policy evaluation problem with state-action-visitation density bounds can be written as
\begin{equation*}\scalemath{
    \begin{optim}
        \max_{\substack{d(\s) \\ p(\s, \a)}}{\expect[r]_{\subalign{(\s, \a) &\sim p(\cdot, \cdot) \\ \s' &\sim \tau(\cdot | \s, \a)}\mspace{-20mu}}[\big]{r(\s, \a, \s')}}
        \st{
            d(\s) &= (1 - \gamma) \iota(\s) + \gamma \quad \expect[lr]_{(\s', \a') \sim p(\cdot, \cdot)\;}[\big]{\tau(\s | \s', \a')} \ge 0 &\;& \forall \s \\
            p(\s, \a) &= d(\s) \pi(\a | \s) && \forall \s \, \forall \a \\
            \lb{p}(\s, \a) &\le p(\s, \a) \le \ub{p}(\s, \a) && \forall \s \, \forall \a
        }
    \end{optim}
}\end{equation*}
The Lagrangian of this altered LP is given by
\begin{equation*}\scalemath{\begin{aligned}
    \MoveEqLeft[2] \mathcal{L}_{\mathrm{VDB}}(d(\s), p(\s, \a), v(\s), q(\s, \a), \mu(\s), \lb{r}(\s, \a), \ub{r}(\s, \a)) \\
    = &-\sum_{\s \in \S}\sum_{\a \in \A} p(\s, \a) \sum_{\s' \in \S} \tau(\s' | \s, \a) r(\s, \a, \s') \\
    &+ \sum_{\s \in \S} v(\s) \bracks*{d(\s) - (1 - \gamma)\iota(\s) - \gamma \sum_{\s' \in \S} \sum_{\a' \in \A} p(\s', \a') \tau(\s | \s', \a')} \\
    &- \sum_{\s \in \S} \mu(\s) d(\s) + \sum_{\s \in \S} \sum_{\a \in \A} q(\s, \a) \bracks[\big]{p(\s, \a) - d(\s) \pi(\a | \s)} \\
    &- \sum_{\s \in \S} \sum_{\a \in \A} \lb{r}(\s, \a) \bracks*{p(\s, \a) - \lb{p}(\s, \a)} \\
    &- \sum_{\s \in \S} \sum_{\a \in \A} \ub{r}(\s, \a) \bracks*{\ub{p}(\s, \a) - p(\s, \a)} \\
    = &-\sum_{\s \in \S} \iota(\s) (1 - \gamma) v(\s) \\
    &+ \sum_{\s \in \S} \sum_{\a \in \A} \lb{r}(\s, \a) \lb{p}(\s, \a) - \sum_{\s \in \S} \sum_{\a \in \A} \ub{r}(\s, \a) \ub{p}(\s, \a) \\
    &+ \sum_{\s \in \S} d(\s) \bracks*{v(\s) - \mu(\s) - \sum_{\a \in \A} \pi(\a | \s) q(\s, \a)} \\
    &+ \sum_{\s \in \S} \sum_{\a \in \A} p(\s, \a) \bracks[\Bigg]{q(\s, \a) - \lb{r}(\s, \a) + \ub{r}(\s, \a) \\[-16pt]
    &\mspace{180mu} - \sum_{\s' \in \S} \tau(\s' | \s, \a) \parens[\big]{r(\s, \a, \s') + \gamma v(\s')}},
\end{aligned}}\end{equation*}
from which following optimality (KKT) conditions can be derived
\begin{itemize}
    \setdisplayskips{2pt}{2pt}
    \item Stationarity of respectively $d$ and $p$
    \begin{equation*}\scalemath{\begin{alignedat}{2}
        \mu^\star(\s) &= v^\star(\s) - \sum_{\a \in \A} \pi(\a | \s) q^\star(\s, \a) &\;& \forall \s \\
        q^\star(\s, \a) &= \lb{r}^\star(\s, \a) - \ub{r}^\star(\s, \a) \\
        &\mspace{84mu} + \sum_{\s' \in \S} \tau(\s' | \s, \a) \parens[\big]{r(\s, \a, \s') + \gamma v^\star(\s')} && \forall \s \, \forall \a
    \end{alignedat}}\end{equation*}
    \item Dual feasibility
    \begin{equation*}\scalemath{\begin{alignedat}{2}
        d^\star(\s) &= (1 - \gamma) \iota(\s) + \gamma \sum_{\s' \in \S} \sum_{\a' \in \A} p(\s', \a') \tau(\s | \s', \a') \ge 0 &\;& \forall \s \\
        p^\star(\s, \a) &= d^\star(\s) \pi(\a | \s) && \forall \s \, \forall \a \\
        \lb{p}(\s, \a) &\le p^\star(\s, \a) \le \ub{p}(\s, \a) && \forall \s \, \forall \a
    \end{alignedat}}\end{equation*}
    \item Primal feasibility
    \begin{equation*}\scalemath{\begin{alignedat}{2}
        v^\star(\s) &\ge \sum_{\a \in \A} \pi(\a | \s) q^\star(\s, \a) &\;& \forall \s \\
        \lb{r}^\star(\s, \a) &\ge 0 \land \ub{r}^\star(\s, \a) \ge 0 && \forall \s \, \forall \a
    \end{alignedat}}\end{equation*}
    \item Complementary slackness
    \begin{equation*}\scalemath{\begin{alignedat}{2}
        v^\star(\s) &= \sum_{\a \in \A} \pi(\a | \s) q^\star(\s, \a) \lor d^\star(\s) = 0 &\;& \forall \s \\
        p^\star(\s, \a) &= \lb{p}(\s, \a) \lor \lb{r}^\star(\s, \a) = 0 && \forall \s \, \forall \a \\
        p^\star(\s, \a) &= \ub{p}(\s, \a) \lor \ub{r}^\star(\s, \a) = 0 && \forall \s \, \forall \a
    \end{alignedat}}\end{equation*}
\end{itemize}
The corresponding primal formulation can be written as
\begin{equation*}\scalemath{
    \begin{optim}
        \mspace{10mu}\min_{\substack{v(\s), q(\s, \a) \\ \lb{r}(\s, \a) \\ \smash[b]{\ub{r}(\s, \a)}}}{\quad\expect[r]_{\;\s_0 \sim \iota(\cdot)}[\big]{(1 - \gamma) v(\s_0)} - \sum_{\s \in \S} \sum_{\a \in \A} \lb{r}(\s, \a) \lb{p}(\s, \a) \\
        + \sum_{\s \in \S} \sum_{\a \in \A} \ub{r}(\s, \a) \ub{p}(\s, \a)}
        \st{
            v(\s) &\ge \expect[lr]_{\mspace{32mu}\a \sim \pi(\cdot | \s)}[\big]{q(\s, \a)} &\;& \forall \s \\
            q(\s, \a) &= \expect[lr]_{\qquad\s' \sim \tau(\cdot | \s, \a)}\bracks[\big]{r(\s, \a, \s') + \lb{r}(\s, \a) - \ub{r}(\s, \a) + \gamma v(\s')} && \forall \s \, \forall \a \\
            \lb{r}(\s, \a) &\ge 0 \land \ub{r}(\s, \a) \ge 0 && \forall \s \, \forall \a
        }
    \end{optim}
}\end{equation*}
\lemref{lem:pe_dual} can be proven for this constrained Policy Evaluation problem without major changes. \lemref{lem:pe_prim} also holds, albeit for the modified reward function $r_{\mathrm{VDB}}^{}$ \eqref{eq:vdb_mod}, with learnable parameters $\lb{r}$ and $\ub{r}$, and corresponding value functions.
\begin{replemma}{lem:vdb_prim}[]
    The optimal primal variables $v^\star$ and $q^\star$ satisfy the Bellman equation and are equivalent to the adjusted value functions of the given policy $\pi$ for the considered MDP with optimal modified reward $r_{\mathrm{VDB}}^\star(\s, \a, \s') = r(\s, \a, \s') + \lb{r}^\star(\s, \a) - \ub{r}^\star(\s, \a)$.
\end{replemma}
\begin{IEEEproof}
    Following similar arguments as in \lemref{lem:pe_prim}, we obtain $v^\star(\s) = \sum_{\a \in \A} \pi(\a | \s) \sum_{\s' \in \S} \tau(\s' | \s, \a) \parens[\big]{r_{\mathrm{VDB}}^\star(\s, \a, \s') + \gamma v^\star(\s')}$ and $q^\star(\s, \a) = \sum_{\s' \in \S} \tau(\s' | \s, \a) \parens[\big]{r_{\mathrm{VDB}}^\star(\s, \a, \s') + \gamma v^\star(\s')}$.
    From the former equality it is clear that $v^\star$ satisfies the Bellman equation and is thus equivalent to the adjusted value function $v_\mathrm{VDB}^\pi$ of the given policy for the considered MDP with altered reward $r_{\mathrm{VDB}}^\star$. From the latter equality we can then obtain the equivalency between $q^\star$ and $q_{\mathrm{VDB}}^\pi$.
\end{IEEEproof}
Furthermore, these optimal reward modifications satisfy following property.
\begin{repproposition}{prop:vdb_mod}[]
    The optimal modified reward $r_{\mathrm{VDB}}^\star$ is only altered by the tight bounds $\lb{p}(\s, \a) = p^{\pi}(\s, \a)$ and $p^{\pi}(\s, \a) = \ub{p}(\s, \a)$.
\end{repproposition}
\begin{IEEEproof}
    This follows immediately from the complementary slackness conditions, as $\lb{r}^\star(\s, \a) = 0$ for the strictly satisfied constraints $\lb{p}(\s, \a) < p^{\pi}(\s, \a)$, while $\ub{r}^\star(\s, \a) = 0$ for the strictly satisfied constraints $p^{\pi}(\s, \a) < \ub{p}(\s, \a)$.
\end{IEEEproof}
By plugging this altered policy evaluation LP in the Policy Optimization problem, we can prove the following property.
\begin{reptheorem}{thm:vdb_pol}[]
    The optimal solution $\pi^\star$ of the nested Policy Optimization \eqref{eq:po} and constrained Policy Evaluation \eqref{eq:vdb_pe} problems, is the optimal policy $\pi^*$ for the considered MDP with visitation density bounds.
\end{reptheorem}
\begin{IEEEproof}
    The partial derivative of the average reward $\rav^\pi$ can be evaluated using Danskin's theorem as $\diffp{\rav^\pi}{\pi(\a | \s)}[\pi^\star] = -\diffp{\mathcal{L}_{\mathrm{VDB}}}{\pi(\a | \s)}[\pi^\star,d^\star,q^\star] = d^{\pi^\star}(\s) q_{\mathrm{VDB}}^{\pi^\star}(\s, \a)$.
    From the optimality conditions and \propref{prop:po_cond}, we then have $\lambda^\star(\s) \ge d^{\pi^\star}(\s) q_{\mathrm{VDB}}^{\pi^\star}(\s, \a)$ for all $\s$ and $\a$, and the equality holds for all actions taken by the policy $\A^{\pi^\star}(\s) = \argmax_{\a} q_{\mathrm{VDB}}^{\pi^\star}(\s, \a)$.
    Hence, $\pi^\star$ is a greedy policy with respect to its own adjusted value function $q_{\mathrm{VDB}}^{\pi^\star}$. This means $v_{\mathrm{VDB}}^{\pi^\star}$ and $q_{\mathrm{VDB}}^{\pi^\star}$ satisfy the Bellman optimality equation and are equivalent to the optimal adjusted value functions $v_{\mathrm{VDB}}^*$ and $q_{\mathrm{VDB}}^*$. Combining the results from \lemref{lem:vdb_opt} and  \lemref{lem:vdb_greedy}, we can then conclude that $\pi^\star$ is an optimal policy for the considered MDP with visitation density bounds.
\end{IEEEproof}

\subsection{Action Density Bounds}\label{app:adb}
The altered dual value learning problem with action density bounds can be written as
\begin{equation*}\scalemath{
    \begin{optim}
        \max_{\substack{d(\s) \\ p(\s, \a)}}{\expect[r]_{\subalign{(\s, \a) &\sim p(\cdot, \cdot) \\ \s' &\sim \tau(\cdot | \s, \a)}\mspace{-20mu}}[\big]{r(\s, \a, \s')}}
        \st{
            \sum_{\a \in \A} p(\s, \a) &= d(\s) &\;& \forall \s \\[-6pt]
            d(\s) &= (1 - \gamma) \iota(\s) + \gamma \quad \expect[lr]_{(\s', \a') \sim p(\cdot, \cdot)\;}[\big]{\tau(\s | \s', \a')} && \forall \s \\[-6pt]
            p(\s, \a) &\ge 0 && \forall \s \, \forall \a \\
            d(\s) \lb{\pi}(\a | \s) &\le p(\s, \a) \le d(\s) \ub{\pi}(\a | \s) && \forall \s \, \forall \a
        }
    \end{optim}
}\end{equation*}
The Lagrangian of this altered dual problem is given by
\begin{equation*}\scalemath{\begin{aligned}
    \MoveEqLeft[2] \mathcal{L}_{ADB}(d(\s), p(\s, \a), \lambda(\s), v(\s), \mu(\s, \a), \lb{r}(\s, \a), \ub{r}(\s, \a)) \\
    = &-\sum_{\s \in \S}\sum_{\a \in \A} p(\s, \a) \sum_{\s' \in \S} \tau(\s' | \s, \a) r(\s, \a, \s') \\
    &+ \sum_{\s \in \S} \lambda(\s) \bracks*{\sum_{\a \in \A} p(\s, \a) - d(\s)} \\
    &+ \sum_{\s \in \S} v(\s) \bracks*{d(\s) - (1 - \gamma)\iota(\s) - \gamma \sum_{\s' \in \S} \sum_{\a' \in \A} p(\s', \a') \tau(\s | \s', \a')} \\
    &- \sum_{\s \in \S} \sum_{\a \in \A} \mu(\s, \a) p(\s, \a) \\
    &- \sum_{\s \in \S} \sum_{\a \in \A} \lb{r}(\s, \a) \bracks*{p(\s, \a) - d(\s) \lb{\pi}(\a | \s)} \\
    &- \sum_{\s \in \S} \sum_{\a \in \A} \ub{r}(\s, \a) \bracks*{d(\s) \ub{\pi}(\a | \s) - p(\s, \a)} \\
    = &-\sum_{\s \in \S} \iota(\s) (1 - \gamma) v(\s) \\
    &+ \sum_{\s \in \S} d(\s) \bracks[\Bigg]{v(\s) - \lambda(\s) + \sum_{\a \in \A} \parens[\big]{\lb{r}(\s, \a) \lb{\pi}(\a | \s) \\[-18pt]
    &\mspace{280mu} -  \ub{r}(\s, \a) \ub{\pi}(\a | \s)}} \\
    &+ \sum_{\s \in \S} \sum_{\a \in \A} p(\s, \a) \bracks[\Bigg]{\lambda(\s) - \mu(\s, \a) - \lb{r}(\s, \a) + \ub{r}(\s, \a) \\[-16pt]
    &\mspace{180mu} - \sum_{\s' \in \S} \tau(\s' | \s, \a) \parens[\big]{r(\s, \a, \s') + \gamma v(\s')}},
\end{aligned}}\end{equation*}
resulting in following optimality (KKT) conditions
\begin{itemize}
    \setdisplayskips{2pt}{2pt}
    \item Stationarity of respectively $d$ and $p$
    \begin{equation*}\scalemath{\mspace{-4mu}\begin{alignedat}{2}
        \lambda^\star(\s) &= v^\star(\s) + \sum_{\a \in \A} \parens[\big]{\lb{r}^\star(\s, \a) \lb{\pi}(\a | \s) - \ub{r}^\star(\s, \a) \ub{\pi}(\a | \s)} &\;& \forall \s \\
        \mu^\star(\s, \a) &= v^\star(\s) + \sum_{\tilde{\a} \in \A} \parens[\big]{\lb{r}^\star(\s, \tilde{\a}) \lb{\pi}(\tilde{\a} | \s) - \ub{r}^\star(\s, \tilde{\a}) \ub{\pi}(\tilde{\a} | \s)} \\
        &\mspace{64mu} - \lb{r}^\star(\s, \a) + \ub{r}^\star(\s, \a) \\
        &\mspace{64mu} - \sum_{\s' \in \S} \tau(\s' | \s, \a) \parens[\big]{r(\s, \a, \s') + \gamma v^\star(\s')} && \forall \s \, \forall \a
    \end{alignedat}}\end{equation*}
    \item Dual feasibility
    \begin{equation*}\scalemath{\mspace{-10mu}\begin{alignedat}{2}
        \sum_{\a \in \A} p^\star(\s, \a) &= d^\star(\s) &\;& \forall \s \\
        d^\star(\s) &= (1 - \gamma) \iota(\s) + \gamma \sum_{\s' \in \S} \sum_{\a' \in \A} p^\star(\s', \a') \tau(\s | \s', \a') && \forall \s \\
        p^\star(\s, \a) &\ge 0 && \forall \s \, \forall \a \\
        d(\s) \lb{\pi}(\a | \s) &\le p^\star(\s, \a) \le d(\s) \ub{\pi}(\a | \s) && \forall \s
    \end{alignedat}}\end{equation*}
    \item Primal feasibility, introducing auxiliary $q$
    \begin{equation*}\scalemath{\begin{alignedat}{2}
        v^\star(\s) &\ge q^\star(\s, \a) \\[-2pt]
        q^\star(\s, \a) &= \sum_{\s' \in \S} \tau(\s' | \s, \a) \bracks[\Big]{r(\s, \a, \s') + \lb{r}^\star(\s, \a) - \ub{r}^\star(\s, \a) \\[-10pt]
        &\mspace{204mu} - \sum_{\tilde{\a} \in \A} \lb{r}^\star(\s, \tilde{\a}) \lb{\pi}(\tilde{\a} | \s) \\[-2pt]
        &\mspace{204mu} + \sum_{\tilde{\a} \in \A} \ub{r}^\star(\s, \tilde{\a}) \ub{\pi}(\tilde{\a} | \s) \\[-12pt]
        &\mspace{300mu} + \gamma v^\star(\s')} &\;& \forall \s \, \forall \a \\
        \lb{r}^\star(\s, \a) &\ge 0 \land \ub{r}^\star(\s, \a) \ge 0 && \forall \s \, \forall \a
    \end{alignedat}}\end{equation*}
    \item Complementary slackness
    \begin{equation*}\scalemath{\begin{alignedat}{2}
        v^\star(\s) &= q^\star(\s, \a) 
        \lor p^\star(\s, \a) = 0 &\;& \forall \s \, \forall \a \\
        p^\star(\s, \a) &= d^\star(\s) \lb{\pi}(\a | \s) \lor \lb{r}^\star(\s, \a) = 0 && \forall \s \, \forall \a \\
        p^\star(\s, \a) &= d^\star(\s) \ub{\pi}(\a | \s) \lor \ub{r}^\star(\s, \a) = 0 && \forall \s \, \forall \a
    \end{alignedat}}\end{equation*}
\end{itemize}
The primal formulation is obtained as
\begin{equation*}\scalemath{
    \begin{optim}
        \min_{\substack{v(\s), q(\s, \a) \\ \smash[b]{\lb{r}(\s, \a), \ub{r}(\s, \a)}}}{\mspace{26mu}\expect[r]_{\;\s_0 \sim \iota(\cdot)}[\big]{(1 - \gamma) v(\s_0)}}
        \st{
            v(\s) &\ge q(\s, \a) &\;& \forall \s \, \forall \a \\
            q(\s, \a) &= \expect[blr]_{\qquad\s' \sim \tau(\cdot | \s, \a)}\bracks[\Big]{r(\s, \a, \s') + \lb{r}(\s, \a) - \sum_{\tilde{\a} \in \A} \lb{r}(\s, \tilde{\a}) \lb{\pi}(\tilde{\a} | \s) \\
            &\mspace{128mu} - \ub{r}(\s, \a) + \smash[b]{\sum_{\tilde{\a} \in \A}} \ub{r}(\s, \tilde{\a}) \ub{\pi}(\tilde{\a} | \s) \\
            &\mspace{312mu}+ \gamma v(\s')} && \forall \s \, \forall \a \\
            \lb{r}(\s, \a) &\ge 0 \land \ub{r}(\s, \a) \ge 0 && \forall \s \, \forall \a
        }
    \end{optim}
}\end{equation*}
From the optimality conditions, it is immediately clear \propref{prop:vl_prob} still holds for this constrained variant and it can be extended as follows.
\begin{proposition}
    The policy $\pi(\a | \s)$ that can be derived from any feasible dual variables $d$ and $p$ is a feasible policy for the considered MDP with action density bounds.
\end{proposition}
\begin{IEEEproof}
    From \propref{prop:vl_prob}, we know that the dual variables correspond to the visitation densities of a policy $\pi$, i.e. $d(\s) = d^\pi(\s)$ and $p(\s, \a) = d^\pi(\s) \pi(\a | \s)$. The additional constraints can thus be rewritten as $d^\pi(\s) \lb{\pi}(\a | \s) \le d^\pi(\s) \pi(\a | \s) \le d^\pi(\s) \ub{\pi}(\a | \s)$, implying $\lb{\pi}(\a | \s) \le \pi(\a | \s) \le \ub{\pi}(\a | \s)$ for all visited states. The dual variables $d$ and $p$ are thus the visitation densities of a policy satisfying those action density bounds.
\end{IEEEproof}
Consequently, \thmref{thm:vl_dual} can also be straightforwardly adapted as follows.
\begin{reptheorem}{thm:adb_dual}[]
    The policy $\pi^\star$ that can be derived from the optimal dual variables $d^\star$ and $p^\star$ is an optimal policy for the considered MDP with action density bounds.
\end{reptheorem}
\begin{IEEEproof}
    Combine the results from the previous Proposition with the proof of \thmref{thm:vl_dual}.
\end{IEEEproof}
\lemref{lem:vl_prim} also holds, albeit for the modified reward function $r_{\mathrm{ADB}}^{}$ \eqref{eq:adb_mod}, with learnable parameters $\lb{r}$ and $\ub{r}$, and corresponding value functions.
\begin{replemma}{lem:adb_prim}[]
    The optimal primal variables $v^\star$ and $q^\star$ satisfy the Bellman optimality equation and are thus equivalent to the optimal adjusted value functions $v_{\mathrm{VDB}}^*$ and $q_{\mathrm{VDB}}^*$ for the considered MDP with optimal modified reward $r_{\mathrm{VDB}}^\star(\s, \a, \s') = r(\s, \a, \s') + \lb{r}^\star(\s, \a) - \sum_{\tilde{\a} \in \A} \lb{r}(\s, \tilde{\a}) \lb{\pi}(\tilde{\a} | \s) - \ub{r}^\star(\s, \a) + \sum_{\tilde{\a} \in \A} \ub{r}(\s, \tilde{\a}) \ub{\pi}(\tilde{\a} | \s)$.
\end{replemma}
\begin{IEEEproof}
    Following the same arguments as in \lemref{lem:vl_prim}, we obtain $v^\star(\s) = \max_{\a} \sum_{\s' \in \S} \tau(\s' | \s, \a) \parens[\big]{r_{\mathrm{ADB}}^\star(\s, \a, \s') + \gamma v^\star(\s')}$.
    Satisfying the Bellman optimality equation, $v^\star$ is thus equivalent to the optimal value function $v_{\mathrm{ADB}}^*$ for the considered MDP with altered reward $r_{\mathrm{ADB}}^\star$.
    From the primal feasibility conditions $q^\star(\s, \a) = \sum_{\s' \in \S} \tau(\s' | \s, \a) \parens[\big]{r_{\mathrm{ADB}}^\star(\s, \a, \s') + \gamma v^\star(\s')}$, we then also obtain the equivalency between $q^\star$ and $q_{\mathrm{ADB}}^*$.
\end{IEEEproof}
Finally, these optimal adjusted value functions and optimal reward modifications satisfy following properties.
\begin{reptheorem}{thm:adb_greedy}[]
    The optimal policy $\pi^\star$ is greedy with respect to the optimal adjusted value functions $v_{\mathrm{ADB}}^*$ and $q_{\mathrm{ADB}}^*$.
\end{reptheorem}
\begin{IEEEproof}
    Following the same arguments as in the proof for \lemref{lem:vl_prim}, we obtain $\A^{\pi^\star}(\s) = \A^*(\s) = \argmax_{\a} \sum_{\s' \in \S} \tau(\s' | \s, \a) \parens[\big]{r_{\mathrm{ADB}}^\star(\s, \a, \s') + \gamma v^\star(\s')}$.
    From the previous Lemma, it then follows that $\A^{\pi^\star}(\s) = \argmax_{\a} q_{\mathrm{ADB}}^*(\s, \a)$, showing the greediness of $\pi^\star$ with respect to the optimal adjusted value functions.
\end{IEEEproof}
\begin{repproposition}{prop:adb_mod}[]
    The optimal modified reward $r_{\mathrm{ADB}}^\star$ is only altered by the tight bounds $\lb{\pi}(\a | \s) = \pi^\star(\a | \s)$ and $\pi^\star(\a | \s) = \ub{\pi}(\a | \s)$.
\end{repproposition}
\begin{IEEEproof}
    This follows from the complementary slackness conditions, as $\lb{r}^\star(\s, \a) = 0$ for the strictly satisfied constraints $d^{\pi^\star}(\s) \lb{\pi}(\a | \s) < p^{\pi^\star}(\s, \a)$, while $\ub{r}^\star(\s, \a) = 0$ for the strictly satisfied constraints $p^{\pi^\star}(\s, \a) < d^{\pi^\star}(\s) \ub{\pi}(\a | \s)$.
\end{IEEEproof}

\subsection{Transition Constraints}\label{app:tc}
\subsubsection{Value Learning}
The altered dual value learning problem with an average transition constraint is denoted as
\begin{equation*}\scalemath{
    \begin{optim}
        \max_{\substack{d(\s) \\ p(\s, \a)}}{\expect[r]_{\subalign{(\s, \a) &\sim p(\cdot, \cdot) \\ \s' &\sim \tau(\cdot | \s, \a)}\mspace{-20mu}}[\big]{r(\s, \a, \s')}}
        \st{
            \sum_{\a \in \A} p(\s, \a) &= d(\s) &\;& \forall \s \\[-6pt]
            d(\s) &= (1 - \gamma) \iota(\s) + \gamma \quad \expect[lr]_{(\s', \a') \sim p(\cdot, \cdot)\;}[\big]{\tau(\s | \s', \a')} && \forall \s \\[-6pt]
            p(\s, \a) &\ge 0 && \forall \s \, \forall \a \\
            0 &\ge p(\s, \a) \expect[lr]_{\qquad\s' \sim \tau(\cdot | \s, \a)}[\big]{c(\s, \a, \s')} && \forall \s \, \forall \a
        }
    \end{optim}
}\end{equation*}
The Lagrangian of this altered dual problem is then obtained as
\begin{equation*}\scalemath{\begin{aligned}
    \MoveEqLeft[2] \mathcal{L}_{ATC}(d(\s), p(\s, \a), \lambda(\s), v(\s), \mu(\s, \a), \rc(\s, \a)) \\
    = &-\sum_{\s \in \S}\sum_{\a \in \A} p(\s, \a) \sum_{\s' \in \S} \tau(\s' | \s, \a) r(\s, \a, \s') \\
    &+ \sum_{\s \in \S} \lambda(\s) \bracks*{\sum_{\a \in \A} p(\s, \a) - d(\s)} \\
    &+ \sum_{\s \in \S} v(\s) \bracks*{d(\s) - (1 - \gamma)\iota(\s) - \gamma \sum_{\s' \in \S} \sum_{\a' \in \A} p(\s', \a') \tau(\s | \s', \a')} \\
    &- \sum_{\s \in \S} \sum_{\a \in \A} \mu(\s, \a) p(\s, \a) \\
    &+ \sum_{\s \in \S} \sum_{\a \in \A} \rc(\s, \a) p(\s, \a) \sum_{\s' \in \S} \tau(\s' | \s, \a) c(\s, \a, \s') \\
    = &-\sum_{\s \in \S} \iota(\s) (1 - \gamma) v(\s) \\
    &+ \sum_{\s \in \S} d(\s) \bracks[\big]{v(\s) - \lambda(\s)} \\
    &+ \sum_{\s \in \S} \sum_{\a \in \A} p(\s, \a) \bracks[\Bigg]{\lambda(\s) - \mu(\s, \a) \\[-12pt]
    &\mspace{182mu} - \sum_{\s' \in \S} \tau(\s' | \s, \a) \parens[\big]{r(\s, \a, \s') + \gamma v(\s') \\[-20pt]
    &\mspace{326mu} - \rc(\s, \a) c(\s, \a, \s')}},
\end{aligned}}\end{equation*}
leading to following optimality (KKT) conditions
\begin{itemize}
    \setdisplayskips{2pt}{2pt}
    \item Stationarity of respectively $d$ and $p$
    \begin{equation*}\scalemath{\begin{alignedat}{2}
        \lambda^\star(\s) &= v^\star(\s) &\;& \forall \s \\
        \mu^\star(\s, \a) &= v^\star(\s) - \smash[b]{\sum_{\s' \in \S}} \tau(\s' | \s, \a) \parens[\big]{r(\s, \a, \s') + \gamma v^\star(\s') \\
        &\mspace{200mu} - \rc^\star(\s, \a) c(\s, \a, \s')} && \forall \s \, \forall \a
    \end{alignedat}}\end{equation*}
    \item Dual feasibility
    \begin{equation*}\scalemath{\mspace{-10mu}\begin{alignedat}{2}
        \sum_{\a \in \A} p^\star(\s, \a) &= d^\star(\s) &\;& \forall \s \\
        d^\star(\s) &= (1 - \gamma) \iota(\s) + \gamma \sum_{\s' \in \S} \sum_{\a' \in \A} p^\star(\s', \a') \tau(\s | \s', \a') && \forall \s \\
        p^\star(\s, \a) &\ge 0 && \forall \s \, \forall \a \\
        0 &\ge p^\star(\s, \a) \sum_{\s' \in \S} \tau(\s' | \s, \a) c(\s, \a, \s') && \forall \s \, \forall \a
    \end{alignedat}}\end{equation*}
    \item Primal feasibility, introducing auxiliary $q$
    \begin{equation*}\scalemath{\begin{alignedat}{2}
        v^\star(\s) &\ge q^\star(\s, \a) &\;& \forall \s \, \forall \a \\
        q^\star(\s, \a) &= \smash[b]{\sum_{\s' \in \S}} \tau(\s' | \s, \a) \parens[\big]{r(\s, \a, \s') - \rc^\star(\s, \a) c(\s, \a, \s') \\
        &\mspace{288mu} + \gamma v^\star(\s')} && \forall \s \, \forall \a \\
        \rc^\star(\s, \a) &\ge 0 && \forall \s \, \forall \a
    \end{alignedat}}\end{equation*}
    \item Complementary slackness
    \begin{equation*}\scalemath{\begin{alignedat}{2}
        v^\star(\s) &= q^\star(\s, \a) 
        \lor p^\star(\s, \a) = 0 &\;& \forall \s \, \forall \a \\
        p^\star(\s, \a) &= 0 \lor \sum_{\s' \in \S} \tau(\s' | \s, \a) c(\s, \a, \s') = 0 \lor \rc^\star(\s, \a) = 0 && \forall \s \, \forall \a
    \end{alignedat}}\end{equation*}
\end{itemize}
The corresponding primal formulation is given by
\begin{equation*}\scalemath{
    \begin{optim}
        \min_{\substack{v(\s), q(\s, \a) \\ \smash[b]{\rc(\s, \a)}}}{\mspace{14mu}\expect[r]_{\;\s_0 \sim \iota(\cdot)}[\big]{(1 - \gamma) v(\s_0)}}
        \st{
            v(\s) &\ge q(\s, \a) &\;& \forall \s \, \forall \a \\
            q(\s, \a) &= \expect[lr]_{\qquad\s' \sim \tau(\cdot | \s, \a)}\bracks[\big]{r(\s, \a, \s') - \rc(\s, \a) c(\s, \a, \s') + \gamma v(\s')} && \forall \s \, \forall \a \\
            \rc(\s, \a) &\ge 0 && \forall \s \, \forall \a
        }
    \end{optim}
}\end{equation*}
From the optimality conditions, it is immediately clear \propref{prop:vl_prob} still holds for this constrained variant and it can be extended as follows.
\begin{proposition}
    The policy $\pi(\a | \s)$ that can be derived from any feasible dual variables $d$ and $p$ is a feasible policy for the considered MDP with average transition constraints.
\end{proposition}
\begin{IEEEproof}
    From \propref{prop:vl_prob}, we know that the dual variables correspond to the visitation densities of a policy $\pi$ with $p(\s, \a) = d^\pi(\s) \pi(\a | \s)$. The additional constraints can thus be rewritten as $d^\pi(\s) \pi(\a | \s) \expect_{\s' \sim \tau(\cdot | \s, \a)}{c(\s, \a, \s')} \le 0$, implying $\bar{c}(\s, \a) \le 0$ for all visited states and actions. The dual variables $d$ and $p$ are thus the visitation densities of a policy satisfying those average transition constraints.
\end{IEEEproof}
Consequently, \thmref{thm:vl_dual} can also be straightforwardly adapted as follows.
\begin{reptheorem}{thm:tc_dual}[]\label{thm:tc_dual}
    The policy $\pi^\star$ that can be derived from the optimal dual variables $d^\star$ and $p^\star$ is an optimal policy for the considered MDP with average transition constraints.
\end{reptheorem}
\begin{IEEEproof}
    Combine the results from the previous Proposition with the proof of \thmref{thm:vl_dual}.
\end{IEEEproof}
\lemref{lem:vl_prim} also holds, albeit for the modified reward function $r_{\mathrm{ATC}}^{}$ \eqref{eq:atc_mod}, with learnable parameters $\rc$, and corresponding value functions.
\begin{replemma}{lem:tc_prim}[]
    The optimal primal variables $v^\star$ and $q^\star$ satisfy the Bellman optimality equation and are thus equivalent to the optimal adjusted value functions $v_{\mathrm{ATC}}^*$ and $q_{\mathrm{ATC}}^*$ for the considered MDP with optimal modified reward $r_{\mathrm{ATC}}^\star(\s, \a, \s') = r(\s, \a, \s') - \rc^\star(\s, \a) c(\s, \a, \s')$.
\end{replemma}
\begin{IEEEproof}
    Following the same arguments as in \lemref{lem:vl_prim}, we obtain $v^\star(\s) = \max_{\a} \sum_{\s' \in \S} \tau(\s' | \s, \a) \parens[\big]{r_{\mathrm{ATC}}^\star(\s, \a, \s') + \gamma v^\star(\s')}$.
    Satisfying the Bellman optimality equation, $v^\star$ is thus equivalent to the optimal value function $v_{\mathrm{ATC}}^*$ for the considered MDP with altered reward $r_{\mathrm{ATC}}^\star$.
    From the primal feasibility conditions $q^\star(\s, \a) = \sum_{\s' \in \S} \tau(\s' | \s, \a) \parens[\big]{r_{\mathrm{ATC}}^\star(\s, \a, \s') + \gamma v^\star(\s')}$, we then also obtain the equivalency between $q^\star$ and $q_{\mathrm{ATC}}^*$.
\end{IEEEproof}
Finally, these optimal adjusted value functions and optimal reward modifications satisfy following properties.
\begin{reptheorem}{thm:tc_greedy}[]\label{thm:tc_greedy}
    The optimal policy $\pi^\star$ is greedy with respect to the optimal adjusted value functions $v_{\mathrm{ATC}}^*$ and $q_{\mathrm{ATC}}^*$.
\end{reptheorem}
\begin{IEEEproof}
    Following the same arguments as in the proof for \lemref{lem:vl_prim}, we obtain $\A^{\pi^\star}(\s) = \A^*(\s) = \argmax_{\a} \sum_{\s' \in \S} \tau(\s' | \s, \a) \parens[\big]{r_{\mathrm{ATC}}^\star(\s, \a, \s') + \gamma v^\star(\s')}$.
    From the previous Lemma, it then follows that $\A^{\pi^\star}(\s) = \argmax_{\a} q_{\mathrm{ATC}}^*(\s, \a)$, showing the greediness of $\pi^\star$ with respect to the optimal adjusted value functions.
\end{IEEEproof}
\begin{repproposition}{prop:tc_mod}[]
    The optimal modified reward $r_{\mathrm{ATC}}^\star$ is only altered by nonvisited state-action pairs or tight bounds $\bar{c}(\s, \a) = 0$.
\end{repproposition}
\begin{IEEEproof}
    This follows from the complementary slackness conditions, as $\rc^\star(\s, \a) = 0$ for the strictly satisfied constraints $\expect_{\s' \sim \tau(\cdot | \s, \a)}{c(\s, \a, \s')} < 0$ and $p^\star(\s, \a) > 0$.
\end{IEEEproof}
The derived properties are especially useful to obtain similar Constrained GPI schemes for solving RL problems with transition constraints, as shown in the next part.

\subsubsection{Policy Iteration}
The altered dual policy evaluation problem with immediate state-action transition constraints can be written as
\begin{subequations}[ITC]
    \label{eq:itc}[!]
    \begin{equation}\scalemath{
        \begin{optim}[t]
            \max_{\substack{d(\s) \\ p(\s, \a)}}{\expect[r]_{\subalign{(\s, \a) &\sim p(\cdot, \cdot) \\ \s' &\sim \tau(\cdot | \s, \a)}\mspace{-20mu}}[\big]{r(\s, \a, \s')}}
            \st{
                d(\s) &= (1 - \gamma) \iota(\s) + \gamma \quad \expect[lr]_{(\s', \a') \sim p(\cdot, \cdot)\;}[\big]{\tau(\s | \s', \a')} \ge 0 &\;& \forall \s \\
                p(\s, \a) &= d(\s) \pi(\a | \s) && \forall \s \, \forall \a \\
                0 &\ge p(\s, \a) \tau(\s' | \s, \a) \pi(\a' | \s') c(\s, \a, \s', \a') && \forall \s \, \forall \a \, \forall \s' \, \forall \a'
            }
        \end{optim}\mspace{-60mu}
        \label{eq:itc_dual}\subtag[\textperiodcentered]{D}
    }\end{equation}
\end{subequations}
and has as Lagrangian
\begin{equation*}\scalemath{\begin{aligned}
    \MoveEqLeft[2] \mathcal{L}_{\mathrm{ITC}}(d(\s), p(\s, \a), v(\s), q(\s, \a), \mu(\s), \rc(\s, \a, \s', \a')) \\
    = &-\sum_{\s \in \S}\sum_{\a \in \A} p(\s, \a) \sum_{\s' \in \S} \tau(\s' | \s, \a) r(\s, \a, \s') \\
    &+ \sum_{\s \in \S} v(\s) \bracks*{d(\s) - (1 - \gamma)\iota(\s) - \gamma \sum_{\s' \in \S} \sum_{\a' \in \A} p(\s', \a') \tau(\s | \s', \a')} \\
    &+ \sum_{\s \in \S} \sum_{\a \in \A} q(\s, \a) \bracks[\big]{p(\s, \a) - d(\s) \pi(\a | \s)} \\
    &- \sum_{\s \in \S} \mu(\s) d(\s) \\
    &+ \sum_{\s \in \S} \sum_{\a \in \A} \sum_{\s' \in \S} \sum_{\a' \in \A} \bracks[\big]{\rc(\s, \a, \s', \a') p(\s, \a) \tau(\s' | \s, \a) \pi(\a' | \s') \\[-12pt]
    &\mspace{390mu} c(\s, \a, \s', \a')} \\
    = &-\sum_{\s \in \S} \iota(\s) (1 - \gamma) v(\s) \\
    &+ \sum_{\s \in \S} d(\s) \bracks*{v(\s) - \mu(\s) - \sum_{\a \in \A} \pi(\a | \s) q(\s, \a)} \\
    &+ \sum_{\s \in \S} \sum_{\a \in \A} p(\s, \a) \bracks[\Bigg]{q(\s, \a) - \sum_{\s' \in \S} \tau(\s' | \s, \a) \parens[\Big]{r(\s, \a, \s') + \gamma v(\s') \\[-12pt]
    &\mspace{150mu} - \sum_{\a' \in \A} \pi(\a' | \s') \rc(\s, \a, \s', \a') c(\s, \a, \s', \a')}},
\end{aligned}}\end{equation*}
The optimality (KKT) conditions can be derived as
\begin{itemize}
    \setdisplayskips{2pt}{2pt}
    \item Stationarity of respectively $d$ and $p$
    \begin{equation*}\scalemath{\mspace{-10mu}\begin{alignedat}{2}
        \mu^\star(\s) &= v^\star(\s) - \sum_{\a \in \A} \pi(\a | \s) q^\star(\s, \a) &\;& \forall \s \\
        q^\star(\s, \a) &=  \smash[b]{\sum_{\s' \in \S}} \tau(\s' | \s, \a) \parens[\Big]{r(\s, \a, \s') + \gamma v^\star(\s') \\[-2pt]
        &\mspace{46mu} - \sum_{\a' \in \A} \pi(\a' | \s') \rc^\star(\s, \a, \s', \a') c(\s, \a, \s', \a')} && \forall \s \, \forall \a
    \end{alignedat}}\end{equation*}
    \item Dual feasibility
    \begin{equation*}\scalemath{\mspace{-10mu}\begin{alignedat}{2}
        d^\star(\s) &= (1 - \gamma) \iota(\s) \\
        &\mspace{40mu}+ \gamma \sum_{\s' \in \S} \sum_{\a' \in \A} p^\star(\s', \a') \tau(\s | \s', \a') \ge 0 &\;& \forall \s \\
        p^\star(\s, \a) &= d^\star(\s) \pi(\a | \s) && \forall \s \, \forall \a \\
        0 &\ge p^\star(\s, \a) \tau(\s' | \s, \a) \pi(\a' | \s') c(\s, \a, \s', \a') && \forall \s \, \forall \a \, \forall \s' \, \forall \a'
    \end{alignedat}}\end{equation*}
    \item Primal feasibility
    \begin{equation*}\scalemath{\mspace{-10mu}\begin{alignedat}{2}
        v^\star(\s) &\ge \sum_{\a \in \A} \pi(\a | \s) q^\star(\s, \a) &\;& \forall \s \\
        0 &\le \rc^\star(\s, \a, \s', \a') && \forall \s \, \forall \a \, \forall \s' \, \forall \a'
    \end{alignedat}}\end{equation*}
    \item Complementary slackness
    \begin{equation*}\scalemath{\mspace{-10mu}\begin{alignedat}{2}
        v^\star(\s) = \sum_{\a \in \A} \pi(\a | \s) q^\star(\s, \a) &\lor d^\star(\s) = 0 &\;& \forall \s \\
        p^\star(\s, \a) \tau(\s' | \s, \a) \pi(\a' | \s') = 0 &\lor c(\s, \a, \s', \a') = 0 \\
        &\lor \rc^\star(\s, \a, \s', \a') = 0 && \forall \s \, \forall \a \, \forall \s' \, \forall \a'
    \end{alignedat}}\end{equation*}
\end{itemize}
The corresponding primal LP is given by
\begin{subequations}[ITC]
    \begin{equation}\scalemath{
        \begin{optim}[t]
            \min_{\substack{v(\s), q(\s, \a) \\ \smash[b]{\rc(\s, \a, \s', \a')}}}{\mspace{26mu}\expect[r]_{\;\s_0 \sim \iota(\cdot)}[\big]{(1 - \gamma) v(\s_0)}}
            \st{
                v(\s) &\ge \expect[lr]_{\mspace{32mu}\a \sim \pi(\cdot | \s)}[\big]{q(\s, \a)} &\;& \forall \s \\
                \mspace{-16mu}q(\s, \a) &= \expect[blr]_{\qquad\subalign{\s' &\sim \tau(\cdot | \s, \a) \\ \a' &\sim \pi(\cdot | \s')}}\bracks[\big]{r(\s, \a, \s') - \rc(\s, \a, \s', \a') c(\s, \a, \s', \a') \\
                &\mspace{282mu}+ \gamma v(\s')} && \forall \s \, \forall \a \\
                0 &\le \rc(\s, \a, \s', \a') && \forall \s \forall \a \forall \s' \, \forall \a'
            }
        \end{optim}\mspace{-90mu}
        \label{eq:itc_prim}\subtag[\textperiodcentered]{P}
    }\end{equation}
\end{subequations}
For feasible policies, satisfying all imposed transition constraints, \lemref{lem:pe_dual} can be proven without major changes. \lemref{lem:pe_prim} also holds, albeit for the modified reward function $r_{\mathrm{ITC}}^{}$, with learnable parameters $\rc$, and corresponding value functions.
\begin{lemma}
    The optimal primal variables $v^\star$ and $q^\star$ satisfy the Bellman equation and are equivalent to the adjusted value functions of the given policy $\pi$ for the considered MDP with optimal modified reward $r_{\mathrm{ITC}}^\star(\s, \a, \s') = r(\s, \a, \s') - \expect_{\a' \sim \pi(\cdot | \s')}{\rc^\star(\s, \a, \s', \a') c(\s, \a, \s', \a')}$.
\end{lemma}
\begin{IEEEproof}
    Following similar arguments as in \lemref{lem:pe_prim}, we obtain $v^\star(\s) = \sum_{\a \in \A} \pi(\a | \s) \sum_{\s' \in \S} \tau(\s' | \s, \a) \parens[\big]{r_{\mathrm{ITC}}^\star(\s, \a, \s') + \gamma v^\star(\s')}$ and $q^\star(\s, \a) = \sum_{\s' \in \S} \tau(\s' | \s, \a) \parens[\big]{r_{\mathrm{ITC}}^\star(\s, \a, \s') + \gamma v^\star(\s')}$.
    From the former equality it is clear that $v^\star$ satisfies the Bellman equation and is thus equivalent to the adjusted value function $v_\mathrm{ITC}^\pi$ of the given policy for the considered MDP with altered reward $r_{\mathrm{ITC}}^\star$. From the latter equality we can then obtain the equivalency between $q^\star$ and $q_{\mathrm{ITC}}^\pi$.
\end{IEEEproof}
Furthermore, these optimal reward modifications satisfy following property.
\begin{lemma}
    The optimal modified reward $r_{\mathrm{ITC}}^\star$ is only altered by nonvisited transitions or transitions with tight bounds $c(\s, \a, \s', \a') = 0$.
\end{lemma}
\begin{IEEEproof}
    This follows from the complementary slackness conditions, as $\rc^\star(\s, \a, \s', \a') = 0$ for the visited transitions with negative cost, i.e. $p^\star(\s, \a) > 0$ and $\tau(\s' | \s, \a) > 0$ and $\pi(\a' | \s') > 0$ and $c(\s, \a, \s', \a') < 0$.
\end{IEEEproof}
On the other hand, if the policy does not satisfy all additional constraints, the dual problem effectively becomes infeasible and the primal becomes unbounded, caused by the $\rc(\s, \a, \s', \a')$ of the infeasible constraints approaching infinity.
As before, this altered Policy Evaluation LP is plugged in the Policy Optimization problem to find optimal constrained policies, using following property.
\begin{lemma}
    The optimal solution $\pi^\star$ of the nested Policy Optimization \eqref{eq:po} and constrained Policy Evaluation \eqref{eq:itc} problems, is the optimal policy $\pi^*$ for the considered MDP with transition constraints.
\end{lemma}
\begin{IEEEproof}
    The partial derivative of the average reward $\rav^\pi$ can be evaluated using Danskin's theorem as $\diffp{\rav^\pi}{\pi(\a | \s)}[\pi^\star] = -\diffp{\mathcal{L}_{\mathrm{ITC}}}{\pi(\a | \s)}[\pi^\star,d^\star,p^\star,q^\star,\rc^\star] = d^{\pi^\star}(\s) q_{\mathrm{ITC}}^{\pi^\star}(\s, \a) - \sum_{\s' \in \S} \sum_{\a' \in \A} p^{\pi^\star}(\s', \a') \tau(\s | \s', \a') \rc^\star(\s', \a', \s, \a) c(\s', \a', \s, \a)$. Notice that the second term evaluates to zero for all actions taken by the policy $\a \in \A^{\pi^\star}(\s)$, because of the complementary slackness conditions.
    From the optimality conditions and \propref{prop:po_cond}, we then have $\lambda^\star(\s) \ge d^{\pi^\star}(\s) q_{\mathrm{ITC}}^{\pi^\star}(\s, \a) - \expect_{(\s', \a') \sim p^{\pi^\star}(\cdot, \cdot)}{\tau(\s | \s', \a') \rc^\star(\s', \a', \s, \a) c(\s', \a', \s, \a)}$ for all $\s$ and $\a$, and the equality holds for all actions taken by the policy $\A^{\pi^\star}(\s) = \argmax_{\a} q_{\mathrm{ITC}}^{\pi^\star}(\s, \a)$.
    Hence, $\pi^\star$ is a greedy policy with respect to its own adjusted value function $q_{\mathrm{ITC}}^{\pi^\star}$. This means $v_{\mathrm{ITC}}^{\pi^\star}$ and $q_{\mathrm{ITC}}^{\pi^\star}$ satisfy the Bellman optimality equation and are equivalent to the optimal adjusted value functions $v_{\mathrm{ITC}}^*$ and $q_{\mathrm{ITC}}^*$. Combining the results from \thmref{thm:tc_dual} and \thmref{thm:tc_greedy} after modifying them for immediate transition constraints, we can then conclude that $\pi^\star$ is an optimal policy for the considered MDP with transition constraints.
\end{IEEEproof}
The resulting nested optimization problem can be seen as a \emph{Constrained} GPI scheme, in which the policy, adjusted value functions and reward modifications are jointly learned.
Note that for an infeasible policy, violating the transition constraints, the inner Policy Evaluation problem diverges. However, the directions in which the learnable reward parameters $\rc$ are updated, remain useful for the combined policy iteration scheme: the reward is effectively decreased for transitions with positive costs. As a result, during subsequent policy optimization steps, there is an increased incentive for the policy to stay away from such transitions.
The reward is thus modified in such a way that constraint-violating policies are no longer optimal or greedy with respect to the adjusted value functions.

\section{Hyperparameters}\label{app:hyper}
\tabref{table:hyper} below shows the used hyperparameters for the \texttt{DualCRL} algorithm on the two used gymnasium environments.
\begin{table}[h]
    \centering
	\caption{Overview of the used hyperparameters for both environments.}
	\label{table:hyper}
	\begin{tabular}{@{\,} >{\raggedright}m{10em} >{$}l<{$} r @{\;} l r @{\;} l @{\,}}
		\toprule
		\multicolumn{2}{@{\,} l}{\textbf{Hyperparameters}} & \multicolumn{2}{c}{\textbf{CliffWalking}} & \multicolumn{2}{c @{\,}}{\textbf{Pendulum}} \\
		\midrule
        Total training timesteps & k_M & \num{50000} & & \num{50000} \\
		Warmup timesteps & \, & \num{640} & & \num{1000} \\
        Replay buffer size & \abs{\mathcal{B}} & \num{10000} & & \num{50000} \\
		Batch size & B & \num{32} & & \num{64} \\
        Discount factor & \gamma & \num{0.99} & & \num{0.99} \\
        Learning rates (strides) & \eta_q & \num{2e-2} & (1) & \num{4e-4} & (1) \\
         & \eta_\pi & \num{4e-3} & (2) & \num{4e-4} & (2) \\
         & \eta_d & \num{4e-3} & (1) & - \\
         & \eta_r & \num{1e-2} & (10) & \num{4e-4} & (10) \\
		Target networks averaging constant (stride) & \tau & \num{1e-3} & (2) & \num{1e-3} & (2) \\
		Network architecture -- hidden dimensions & \, & - & & $50 \times 10$ \\
		Entropy temperature & \alpha & $1 \rightarrow$ & \num[print-unity-mantissa=false]{1e-2} & \num[print-unity-mantissa=false]{1e-1} $\rightarrow$ & \num[print-unity-mantissa=false]{1e-3} \\
		\bottomrule
	\end{tabular}
\end{table}

\end{document}